%% file: main.tex
\documentclass{article} % For LaTeX2e
\usepackage{iclr2027_conference,times}

\usepackage{hyperref}
\usepackage{url}
\usepackage{amssymb}
\usepackage{amsmath,amsfonts,mathrsfs}
\usepackage{amsthm}
\usepackage{algorithm}
\usepackage{algpseudocode}
\usepackage{tikz}
\usepackage{pgfplots}
\usepackage{stmaryrd}
\usepackage{ulem}
\usepackage{hyperref}
\usepackage{iclr2027_conference,times}
\usepackage{subcaption}
\usepackage{titletoc}
\usepackage{multirow}
\usepgfplotslibrary{groupplots}
\newcommand{\bbR}{\mathbb{R}}
\newcommand{\bbN}{\mathbb{N}}

\newcommand{\bbE}{\mathbb{E}}
\newcommand{\lot}{\textsc{lot}}

\newcommand{\frlc}{\textsc{frlc}}
\newcommand{\rcg}{\textsc{rcg}}
\newcommand{\tc}{\textsc{tc}}
\newcommand{\latot}{\textsc{latot}}
\newcommand{\expcp}{\textsc{ecpmot}}
\newcommand{\impcp}{\textsc{cpmot}}
\newcommand{\tucker}{\textsc{tucker}}

\newcommand{\gr}{\textbf}

\newcommand{\tsp}{^\mathsf{T}}
\newcommand{\defeq}{\overset{\text{def}}{=}}
\newcommand{\bs}[1]{\boldsymbol{\mathcal{#1}}}

\newcommand{\ps}[2]{\langle #1, #2\rangle}
\newcommand{\1}{\mathbf{1}}

\newcommand{\diag}{\mathrm{diag}}

\newcommand{\supp}{\mathrm{supp}}
\newcommand{\Var}{\mathrm{Var}}
\newcommand{\KL}{\mathrm{KL}}

\newtheorem{theorem}{Theorem}[section]
\newtheorem{Theorem}[theorem]{Theorem}
\newtheorem{corollary}[theorem]{Corollary}
\newtheorem{property}[theorem]{Property}
\newtheorem{lemma}[theorem]{Lemma}

\title{Low-Rank Tensor Representations for Efficient Multimarginal Optimal Transport}

\iclrfinalcopy
\renewcommand{\headrulewidth}{0pt}
\renewcommand{\footrulewidth}{0pt}
\author{Johann Clément-Cottuz \& Maxime Bérar \\
LITIS UR 4108\\
Université Rouen Normandie\\
Rouen 76000, France\\
\And
Gilles Gasso \\
LITIS UR 4108 \\
INSA Rouen Normandie \\
Rouen 76000, France \\
}

\begin{document}

\author{
Johann Clément-Cottuz$^{1}$ \quad
Gilles Gasso$^{2}$ \quad
Maxime Bérar$^{1}$ \\
\\
$^{1}$ LITIS UR4108, Université Rouen Normandie \quad $^{2}$ LITIS UR4108, INSA Rouen Normandie
}

\maketitle

\begin{abstract}
Multimarginal Optimal Transport (MOT) suffers from exponential computational and memory costs in the number of marginals, as transport plans are represented by high-dimensional tensors. We introduce Low-Rank Multimarginal Optimal Transport (LRMOT), which restricts transport plans to tensors with bounded multilinear rank, providing a compact and structured representation of multimarginal couplings. This structure enables efficient algorithms with a low memory footprint. We establish approximation bounds for the resulting low-rank formulations and demonstrate experimentally that the proposed algorithms substantially reduce computational time while achieving solutions with low transport cost. 
\end{abstract}

\section{Introduction}

\input{section_1_intro}

\input{section_2}

\input{section_3_LRMMOT}

\input{section_4_pairwise}

\input{section_5_algorithms}

\section{Experiments}
\input{section_6_Appendix}

\end{document}

%% file: section_1_intro.tex
\label{seq:intro}

Modelling strongly correlated particles systems in the presence of disorder
\citep{Byczuk2010} and multi-omic data integration
\citep{Hawinkel20} are two instances where sets of distributions must be geometrically averaged.
Multimarginal Optimal Transport (MOT) \citep{gangbo98,pass2015multi} formalizes this problem, aligning a set of $d$ probability distributions with respect to an efficiency cost. In the discrete case, MOT is a linear program between a cost tensor $\bs C$ and a transport tensor $\bs T$ coupling the distributions. As $d$ increases, this problem becomes intractable.
Although entropic regularization enables MOT to be solved via a sequence of scaling operations with the Sinkhorn algorithm \citep{benamou2015}, the coupling $\bs T$ itself remains a dense tensor of size $n^d$. 
This limitation is particularly problematic when MOT is used repeatedly, for example in machine learning problems such as GANs (see \citet{cao2019mwgan}).

 One way to alleviate the curse of dimensionality is to reduce the problem's density by imposing structure on the cost tensor $\bs C$. When $\bs C$ is tree-structured, \citet{Haasler2021tree} exploit this structure to decouple the associated Gibbs kernel $\bs K$. \citet{strossner2023} combine this approach with low-rank approximations of $\bs{K}$, further improving the problem's tractability. \citet{Haasler2021graph} show that MOT is equivalent to a Bayesian marginal inference problem when $\bs C$ has a graphical structure and certain conditions on the marginal are met. \citet{Ba2022fft} further exploit tree- or a circle- structured costs using FFT-based methods.

A second way to overcome this bottleneck is to impose structure on the transport tensor itself, as in the classic literature on low-rank (bipartite) optimal transport \citep{scetbon_low-rank_2021}. While this line of work has been extended to yield linear-space algorithms for Wasserstein, Gromov-Wasserstein, and fused Gromov-Wasserstein objectives \citep{forrow19fc,lin_making_2021,scetbon_linear-time_2022,halmos_low-rank_2024,jawanpuria_riemannian_2026}, low-rank structure has not yet been imposed directly on the transport tensor in the multimarginal setting. Imposing rank constraints on the transport plan is also motivated by the desire to obtain a more structured representation of the transport and improved statistical convergence guarantees. To the best of our knowledge, the closest work is \citet{tran2021fcmmot}, where the authors penalize the full tensor $\bs T$ to be closer to a rank-$(1,\ldots,1)$ coupling.

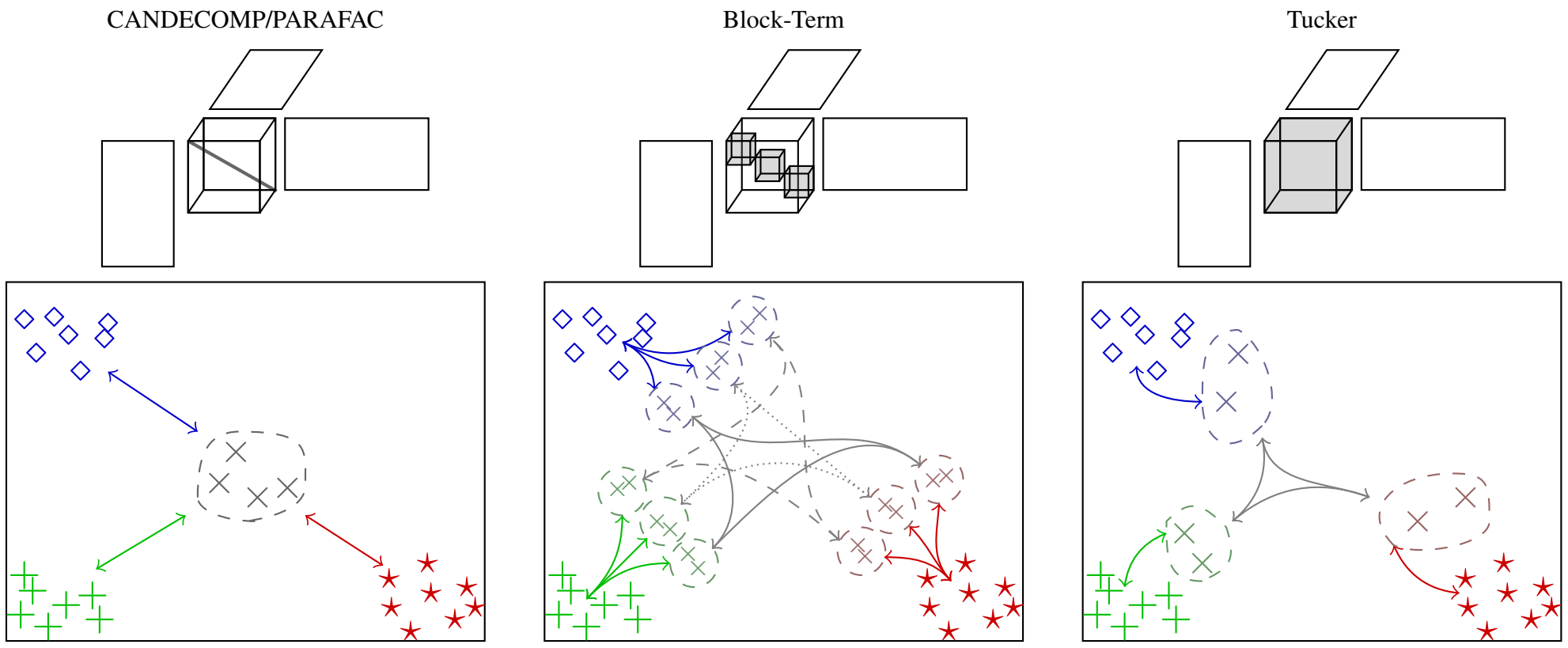
\begin{figure}[H]
    \centering
    \input{figure_intro} 
   \caption{ \textbf{Low-rank MOT through different tensor decompositions}.
   CP decomposition (left) aligns all distributions to one latent measure. Tucker decomposition (right) reduces the MOT problem through $d$ latent measures. Block-Term decomposition provides an intermediary setting (center).}
    \label{fig:3_decomp}
\end{figure}

Low-rank tensor structure encompasses a wide range of settings, most of which can be directly linked to tensor decompositions \citep{Kolda09}. As shown in Figure \ref{fig:3_decomp}, the barycenter view of MOT can be expressed directly through Candecomp/Parafac (CP) \citep{Carroll1970,Hars1970} decomposition (left), while Tucker decomposition \citep{Tucker1966} offers a more general setting in which each distribution is synthesised into a smaller support (right), enabling a smaller MOT to be solved instead. Block-Term decomposition \citep{lathauwer08bt} provides an intermediary structure (center), where mass is transported through selected hubs. Other decompositions such as Tensor Trains \citep{Oseledets11TT} or Tensor Ring \citep{zhao2016tensorringn} connect different tensors through their various unfoldings and lead to other factorization of the transport tensor.

In this work, we define Low-Rank Multimarginal Optimal Transport (LRMOT) by restricting the admissible transport tensors to tensors having an upper bounded multilinear rank. We show that LRMOT is well defined and state bounds on the approximation error with respect to MOT. We instantiate different algorithms related to different decompositions and derive an algorithm taking advantage of a class of structured costs. Their performance is demonstrated through several experiments. The contributions of this paper can be summarized as follows:

\begin{itemize}
\item We define Low-Rank Multimarginal Optimal Transport (Section~\ref{seq:lrmmot}).
\item We incorporate cost structure by formulating an "explicit" CPMOT, provide a bound relating on the optima of the original CPMOT and the explicit CPMOT (Section~\ref{seq:cost_struct}).
\item We present three solvers for the instantiated low-rank problems CPMOT and TuckerMOT, based on tensor decompositions (Section~\ref{seq:algorithms}).
\item We show that our algorithms perform well on multi-marginal problems (Section~\ref{seq:exp}).
\end{itemize}

%% file: figure_intro.tex
    \begin{tikzpicture}
        \draw (0,0) rectangle (4,3);

       % mu_1
    \begin{scope}[color=green!75!black]
        \node at (0.50,0.30) {$+$};
        \node at (0.22,0.42) {$+$};
        \node at (0.72,0.38) {$+$};
        \node at (0.35,0.12) {$+$};
        \node at (0.12,0.22) {$+$};
        \node at (0.15,0.55) {$+$};
        \node at (0.78,0.18) {$+$};
        \draw[<->] (0.75,0.6) -- (1.5,1.05);
        %\node at (1.02,1) {\tiny $Q^{(1)}$};
    \end{scope}

    % mu_2
    \begin{scope}[color=red!80!black]
        \node at (3.55,0.40) {$\star$};
        \node at (3.22,0.28) {$\star$};
        \node at (3.75,0.18) {$\star$};
        \node at (3.38,0.08) {$\star$};
        \node at (3.85,0.45) {$\star$};
        \node at (3.20,0.52) {$\star$};
        \node at (3.52,0.65) {$\star$};
        \node at (3.92,0.28) {$\star$};
        \draw[<->] (3.15,0.63) -- (2.5,1.05);
        %\node at (3,1.05) {\tiny $Q^{(2)}$};
    \end{scope}

    % mu_3
    \begin{scope}[color=blue!80!black]
        \node at (0.25,2.40) {$\diamond$};
        \node at (0.82,2.52) {$\diamond$};
        \node at (0.52,2.55) {$\diamond$};
        \node at (0.62,2.25) {$\diamond$};
        \node at (0.15,2.68) {$\diamond$};
        \node at (0.40,2.7) {$\diamond$};
        \node at (0.85,2.65) {$\diamond$};
        \draw[<->] (0.85,2.25) -- (1.6,1.75);
        %\node at (1.5,2.18) {\tiny $Q^{(3)}$};
    \end{scope}

       % mu_4
    %\begin{scope}[color=yellow!70!black]
        %\node at (3.50,2.50) {$\triangle$};
        %\node at (3.22,2.62) {$\triangle$};
        %\node at (3.72,2.58) {$\triangle$};
        %\node at (3.35,2.32) {$\triangle$};
        %\node at (3.62,2.28) {$\triangle$};
        %\node at (3.12,2.42) {$\triangle$};
       % \node at (3.33,2.75) {$\triangle$};
      %  \node at (3.78,2.38) {$\triangle$};
     %   \draw[<->] (3.15,2.25) -- (2.4,1.75);
    %\end{scope}

    % g
    \begin{scope}[color=gray!80!black]
        \node at (2.10,1.20) {$\times$};
        \node at (1.78,1.32) {$\times$};
        \node at (2.35,1.28) {$\times$};
        \node at (1.92,1.58) {$\times$};
        %\node at (2.4,1.9) {\tiny $g$};
        \draw[dashed]
        (1.6,1.2) .. controls +(0,0.3) and +(-0.3,0) ..
        (1.95,1.75) .. controls +(0.6,0) and +(0,0.3) .. 
        (2.5,1.3) .. controls +(0,-0.3) and +(-0.3,0) .. 
        (2.1,1)  .. controls +(-0.3,0) and +(0,-0.1) .. cycle;
    \end{scope}
    \begin{scope}[yshift = 3cm, xshift = 1cm]
    \def\L{0.15}
    \def\vx{0.13}
    \def\vy{0.19}
        % P1
        \draw (-0.2,0.13) rectangle (-0.2+4*\L,0.13+7*\L);
        % P3
        \draw (0.65,1.37)++ (0.4*0.13,0.4*0.19) -- (0.65+3*0.13,1.37+3*0.19) --(0.65+4*\L+3*0.13,1.37+3*0.19)-- (0.65+4*\L+0.4*0.13,1.37+0.4*0.19)  -- cycle;
        % P2
        \draw (1.33,0.77) rectangle (1.33+8*\L,0.77+4*\L);
        
        \draw (0.65,1.37) rectangle ++(4*\L,-4*\L);
        \draw[color=gray!80!black, thick] (0.52,1.18) -- (1.26,0.76) ;
        \draw (0.52,1.18) rectangle ++(4*\L,-4*\L);
        \draw (0.65,1.37) rectangle ++(4*\L,-4*\L);
        \draw (0.52,1.18) rectangle ++(4*\L,-4*\L);
        \draw (0.52,1.18) -- ++ (\vx,\vy);
        \draw (0.52+4*\L,1.18) -- ++ (\vx,\vy);
        \draw (0.52+4*\L,1.18-4*\L) -- ++ (\vx,\vy);
        \draw (0.52,1.18-4*\L) -- ++ (\vx,\vy);
        %\node[color=green!75!black] at (-0.2+2*\L,0.13+3.5*\L) {\tiny $P^{(1)}$};
        %\node[color=red!80!black] at (1.33+4*\L,0.77+2*\L) {\tiny $P^{(2)}$};
        %\node[color=blue!80!black] at (0.65+2*\L+1.4*0.13,1.37+1.7*0.19) {\tiny $P^{(3)}$};

        \node at (1,2.2){\tiny CANDECOMP/PARAFAC};
       % \node at (1,-0.7){\tiny $\bs T =\sum_{s=1}^r g_s p^{(1)}_s \circ \cdots \circ p^{(d)}_s$};
    \end{scope}

    \begin{scope}[xshift = 4.5cm]
          \draw (0,0) rectangle (4,3);

       % mu_1
    \begin{scope}[color=green!75!black]
        \node at (0.50,0.30) {$+$};
        \node at (0.22,0.42) {$+$};
        \node at (0.72,0.38) {$+$};
        \node at (0.35,0.12) {$+$};
        \node at (0.12,0.22) {$+$};
        \node at (0.15,0.55) {$+$};
        \node at (0.78,0.18) {$+$};
        \draw[<->] (0.35,0.35) to[in=180,out=45] (1.05,0.65);
        \draw[<->] (0.35,0.35) to[in=-135,out=45] (0.86,0.86);
        \draw[<->] (0.35,0.35) to[in=-90,out=45] (0.65,1.05);
    \end{scope}

    % mu_2
    \begin{scope}[color=red!80!black]
        \node at (3.55,0.40) {$\star$};
        \node at (3.22,0.28) {$\star$};
        \node at (3.75,0.18) {$\star$};
        \node at (3.38,0.08) {$\star$};
        \node at (3.85,0.45) {$\star$};
        \node at (3.20,0.52) {$\star$};
        \node at (3.52,0.65) {$\star$};
        \node at (3.92,0.28) {$\star$};
        \draw[<->] (3.4,0.5) to[in=-95,out=132] (3.3,1.15);
        \draw[<->] (3.4,0.5) to[in=-45,out=132] (3.05,0.96);
        \draw[<->] (3.4,0.5) to[in=0,out=132] (2.84,0.7);
    \end{scope}

    % mu_3
    \begin{scope}[color=blue!80!black]
        \node at (0.25,2.40) {$\diamond$};
        \node at (0.82,2.52) {$\diamond$};
        \node at (0.52,2.55) {$\diamond$};
        \node at (0.62,2.25) {$\diamond$};
        \node at (0.15,2.68) {$\diamond$};
        \node at (0.40,2.7) {$\diamond$};
        \node at (0.85,2.65) {$\diamond$};
        \draw[<->] (0.65,2.5) to[bend right] (1.56,2.59);
        \draw[<->] (0.65,2.5) to[in=180,out=-28] (1.25,2.3);
        \draw[<->] (0.65,2.5) to[bend left] (0.92,2.1);
    \end{scope}

       % mu_4
    %\begin{scope}[color=yellow!70!black]
        %\node at (3.50,2.50) {$\triangle$};
        %\node at (3.22,2.62) {$\triangle$};
        %\node at (3.72,2.58) {$\triangle$};
        %\node at (3.35,2.32) {$\triangle$};
        %\node at (3.62,2.28) {$\triangle$};
        %\node at (3.12,2.42) {$\triangle$};
       % \node at (3.33,2.75) {$\triangle$};
      %  \node at (3.78,2.38) {$\triangle$};
     %   \draw[<->] (3.15,2.25) -- (2.4,1.75);
    %\end{scope}

    \begin{scope}[color=gray!80!green]
        \draw[dashed] (1,1) circle (0.2);
        \draw[dashed] (1.25,0.65) circle (0.2);
        \draw[dashed] (0.65,1.25) circle (0.2);
        \node at (1.05,0.93) {\tiny $\times$};
        \node at (0.94,1) {\tiny $\times$};
        \node at (1.2,0.73) {\tiny $\times$};
        \node at (1.285,0.61) {\tiny $\times$};
        \node at (0.61,1.27) {\tiny $\times$};
        \node at (0.71,1.32) {\tiny $\times$};

    \end{scope}
    \begin{scope}[color=gray!80!red]
        \draw[dashed] (2.9,1.1) circle (0.2);
        \draw[dashed] (3.3,1.35) circle (0.2);
        \draw[dashed] (2.65,0.75) circle (0.2);
        \node at (2.94,1.08) {\tiny $\times$};
        \node at (2.85,1.14) {\tiny $\times$};
        \node at (3.25,1.34) {\tiny $\times$};
        \node at (3.36,1.38) {\tiny $\times$};
        \node at (2.68,0.71) {\tiny $\times$};
        \node at (2.62,0.80) {\tiny $\times$};
    \end{scope}
    \begin{scope}[color=gray!80!blue]
        \draw[dashed] (1.45,2.3) circle (0.2);
        \draw[dashed] (1.05,1.95) circle (0.2);
        \draw[dashed] (1.74,2.68) circle (0.2);
        \node at (1.41,2.24) {\tiny $\times$};
        \node at (1.48,2.37) {\tiny $\times$};
        \node at (1,1.99) {\tiny $\times$};
        \node at (1.08,1.9) {\tiny $\times$};
        \node at (1.7,2.62) {\tiny $\times$};
        \node at (1.8,2.74) {\tiny $\times$};
    \end{scope}

    \begin{scope}[color=gray]
        \draw[<->] (3.15,1.465) to[in=45, out=140] (1.4,0.77);
        \draw[<->] (3.15,1.465) to[in=-40, out=140] (1.23,1.88);
        \draw[<->] (1.4,0.77) to[in=-40, out=45] (1.23,1.88);

        \draw[<->,densely dotted] (2.72,1.2)  to[in=45, out=140]  (1.14,1.14);
        \draw[<->,densely dotted] (2.72,1.2)  to[in=-40, out=140]  (1.58,2.15);
        \draw[<->,densely dotted] (1.14,1.14)  to[in=-40, out=45]  (1.58,2.15);

        \draw[<->, dashed] (2.47,0.84) to[in=30, out=140] (0.82,1.35);
        \draw[<->,dashed] (2.47,0.84) to[in=-40, out=140](1.88,2.54);
        \draw[<->,dashed] (0.82,1.35) to[in=-40, out=30] (1.88,2.54);
    \end{scope}
    
    \begin{scope}[yshift = 3cm, xshift = 1cm]
    \def\L{0.15}
    \def\vx{0.13}
    \def\vy{0.19}
        % P1
        \draw (-0.2,0.13) rectangle (-0.2+4*\L,0.13+7*\L);
        % P3
        \draw (0.65,1.37)++ (0.4*0.13,0.4*0.19) -- (0.65+3*0.13,1.37+3*0.19) --(0.65+4*\L+3*0.13,1.37+3*0.19)-- (0.65+4*\L+0.4*0.13,1.37+0.4*0.19)  -- cycle;
        % P2
        \draw (1.33,0.77) rectangle (1.33+8*\L,0.77+4*\L);
        % G
        \draw (0.65,1.37) rectangle ++(4*\L,-4*\L);
        \draw (0.52,1.18) rectangle ++(4*\L,-4*\L);
        \draw (0.52,1.18) -- ++ (\vx,\vy);
        \draw (0.52+4*\L,1.18) -- ++ (\vx,\vy);
        \draw (0.52+4*\L,1.18-4*\L) -- ++ (\vx,\vy);
        \draw (0.52,1.18-4*\L) -- ++ (\vx,\vy);
        % cubes
        \fill[gray, fill opacity=0.3]
            (0.52 ,1.18) -- (0.52+4/3*\L ,1.18) --
            (0.52+4/3*\L ,1.18-4/3*\L) -- (0.52 ,1.18-4/3*\L) -- cycle;

        \fill[gray, fill opacity=0.3]
            (0.52+4/3*\L ,1.18) -- (0.52+4/3*\L +\vx/3, 1.18+\vy/3)  -- (0.52+4/3*\L +\vx/3 ,1.18-4/3*\L +\vy/3) -- (0.52+4/3*\L ,1.18-4/3*\L) -- cycle;

        \fill[gray, fill opacity=0.3]
            (0.52+4/3*\L ,1.18) -- (0.52+4/3*\L +\vx/3, 1.18+\vy/3)  -- (0.52 +\vx/3 ,1.18 +\vy/3) -- (0.52 ,1.18) -- cycle;

        \draw (0.52+\vx / 3,1.18 + \vy /3) rectangle ++(4/3*\L,-4/3*\L);
        \draw (0.52 ,1.18) rectangle ++(4/3*\L,-4/3*\L);
        \draw (0.52+4/3*\L ,1.18) -- ++(\vx/3,\vy/3);
        \draw (0.52+4/3*\L ,1.18-4/3*\L) -- ++(\vx/3,\vy/3);
        \draw (0.52,1.18-4/3*\L) -- ++(\vx/3,\vy/3);
        % cube 2
        \fill[gray, fill opacity=0.3]
            (0.52+4/3*\L+\vx/3 ,1.18-4/3*\L+\vy/3) -- (0.52+8/3*\L+\vx/3,1.18-4/3*\L+\vy/3) --
            (0.52+8/3*\L+\vx/3,1.18-8/3*\L+\vy/3) -- (0.52+4/3*\L+\vx/3 ,1.18-8/3*\L+\vy/3) -- cycle;

        \fill[gray, fill opacity=0.3]
            (0.52+4/3*\L+\vx/3, 1.18-4/3*\L + \vy/3) -- (0.52+4/3*\L+2*\vx/3, 1.18-4/3*\L + 2*\vy/3) -- (0.52+8/3*\L+2*\vx/3, 1.18-4/3*\L + 2*\vy/3)-- (0.52+8/3*\L+\vx/3, 1.18-4/3*\L + \vy/3) -- cycle;

        \fill[gray, fill opacity=0.3]
            (0.52+8/3*\L+\vx/3, 1.18-4/3*\L + \vy/3) -- (0.52+8/3*\L+2*\vx/3, 1.18-4/3*\L + 2*\vy/3) -- (0.52+8/3*\L+2*\vx/3, 1.18-8/3*\L + 2*\vy/3)  -- (0.52+8/3*\L+\vx/3, 1.18-8/3*\L + \vy/3) -- cycle;

        \draw (0.52+4/3*\L+\vx/3, 1.18-4/3*\L + \vy/3) rectangle ++(4/3*\L,-4/3*\L);
        \draw (0.52+4/3*\L+2*\vx/3, 1.18-4/3*\L + 2*\vy/3) rectangle ++(4/3*\L,-4/3*\L);
        
        \draw (0.52+4/3*\L+\vx/3+4/3*\L, 1.18-4/3*\L + \vy/3) -- ++(\vx/3,\vy/3);
        \draw (0.52+4/3*\L+\vx/3+4/3*\L, 1.18-4/3*\L + \vy/3-4/3*\L) -- ++(\vx/3,\vy/3);
        \draw (0.52+4/3*\L+\vx/3, 1.18-4/3*\L + \vy/3-4/3*\L) -- ++(\vx/3,\vy/3);
        \draw (0.52+4/3*\L+\vx/3, 1.18-4/3*\L + \vy/3) -- ++(\vx/3,\vy/3);

        % cube 3
        \fill[gray, fill opacity=0.3]
            (0.52+8/3*\L+2*\vx/3, 1.18-8/3*\L+2*\vy/3) -- (0.52+12/3*\L+2*\vx/3, 1.18-8/3*\L+2*\vy/3) -- (0.52+12/3*\L+2*\vx/3, 1.18-12/3*\L+2*\vy/3) -- (0.52+8/3*\L+2*\vx/3, 1.18-12/3*\L+2*\vy/3) -- cycle;

        \fill[gray, fill opacity=0.3]
            (0.52+8/3*\L+2*\vx/3, 1.18-8/3*\L+2*\vy/3) -- (0.52+8/3*\L+3*\vx/3, 1.18-8/3*\L+3*\vy/3) -- (0.52+12/3*\L+3*\vx/3, 1.18-8/3*\L+3*\vy/3) -- (0.52+12/3*\L+2*\vx/3, 1.18-8/3*\L+2*\vy/3)-- cycle;

        \fill[gray, fill opacity=0.3]
             (0.52+12/3*\L+2*\vx/3, 1.18-8/3*\L+2*\vy/3) --  (0.52+12/3*\L+3*\vx/3, 1.18-8/3*\L+3*\vy/3) -- (0.52+12/3*\L+3*\vx/3, 1.18-12/3*\L+3*\vy/3) -- (0.52+12/3*\L+2*\vx/3, 1.18-12/3*\L+2*\vy/3) -- cycle;

        \draw (0.52+8/3*\L+2*\vx/3, 1.18-8/3*\L+2*\vy/3) rectangle ++(4/3*\L,-4/3*\L);
        \draw (0.52+8/3*\L+3*\vx/3, 1.18-8/3*\L+3*\vy/3) rectangle ++(4/3*\L,-4/3*\L);
        
        \draw (0.52+8/3*\L+2*\vx/3, 1.18-8/3*\L+2*\vy/3) -- ++(\vx/3,\vy/3);
        \draw (0.52+12/3*\L+2*\vx/3, 1.18-8/3*\L+2*\vy/3) -- ++(\vx/3,\vy/3);
        \draw (0.52+12/3*\L+2*\vx/3, 1.18-12/3*\L+2*\vy/3) -- ++(\vx/3,\vy/3);
        \draw (0.52+8/3*\L+2*\vx/3, 1.18-12/3*\L+2*\vy/3) -- ++(\vx/3,\vy/3);
       
        %\node[color=green!75!black] at (-0.2+2*\L,0.13+3.5*\L) {\tiny $P^{(1)}$};
        %\node[color=red!80!black] at (1.33+4*\L,0.77+2*\L) {\tiny $P^{(2)}$};
        %\node[color=blue!80!black] at (0.65+2*\L+1.4*0.13,1.37+1.7*0.19) {\tiny $P^{(3)}$};

        \node at (1,2.2){\tiny Block-Term};
        %\node at (1,-0.7){\tiny $\bs T  = \sum_{s=1}^R \bs G_s\times_1 P_s^{(1)} \cdots \times _d P_s^{(d)}$};
    \end{scope}
    \end{scope}

    \begin{scope}[xshift = 9cm]
          \draw (0,0) rectangle (4,3);

       % mu_1
    \begin{scope}[color=green!75!black]
        \node at (0.50,0.30) {$+$};
        \node at (0.22,0.42) {$+$};
        \node at (0.72,0.38) {$+$};
        \node at (0.35,0.12) {$+$};
        \node at (0.12,0.22) {$+$};
        \node at (0.15,0.55) {$+$};
        \node at (0.78,0.18) {$+$};
        \draw[<->] (0.7,0.9) to[bend right] (0.35,0.45);
        %\node at (0.4,1) {\tiny $Q^{(1)}$};
    \end{scope}

    % mu_2
    \begin{scope}[color=red!80!black]
        \node at (3.55,0.40) {$\star$};
        \node at (3.22,0.28) {$\star$};
        \node at (3.75,0.18) {$\star$};
        \node at (3.38,0.08) {$\star$};
        \node at (3.85,0.45) {$\star$};
        \node at (3.20,0.52) {$\star$};
        \node at (3.52,0.65) {$\star$};
        \node at (3.92,0.28) {$\star$};
        \draw[<->] (3.15,0.4) to[bend left] (2.6,0.8);
        %\node at (2.7,0.28) {\tiny $Q^{(2)}$};
    \end{scope}

    % mu_3
    \begin{scope}[color=blue!80!black]
        \node at (0.25,2.40) {$\diamond$};
        \node at (0.82,2.52) {$\diamond$};
        \node at (0.52,2.55) {$\diamond$};
        \node at (0.62,2.25) {$\diamond$};
        \node at (0.15,2.68) {$\diamond$};
        \node at (0.40,2.7) {$\diamond$};
        \node at (0.85,2.65) {$\diamond$};
        \draw[<->] (0.45,2.3) to[in=180,out=-90]  (1,2);
       % \node at (0.45,1.85) {\tiny $Q^{(3)}$};
    \end{scope}
        
    % g1
    \begin{scope}[color=gray!80!green]
        \draw[dashed]
        (1,0.5) .. controls +(-0.3,0) and +(0,-0.3) .. 
        (0.7,1)  .. controls +(0.2,0.2) and +(-0.2,0.2) ..
        (1.2,0.95) .. controls +(0.1,-0.3) and +(0.2,0.1) .. cycle;
        \node at (1,0.65) {$\times$};
        \node at (0.85,0.9) {$\times$};
        %\node at (1.32,0.4) {\tiny $g^{(1)}$};
    \end{scope}
    \begin{scope}[color=gray!80!red]
        \draw[dashed]
        (2.6,0.8) .. controls +(-0.3,0.3) and +(-0.5,0) .. 
        (3.1,1.4)  .. controls +(0.3,0) and +(0,0.3) ..
        (3.4,1) .. controls +(0,-0.2) and +(0.2,-0.1) .. cycle;
        \node at (2.8,1) {$\times$};
        \node at (3.2,1.2) {$\times$};
        %\node at (3.4,1.6) {\tiny $g^{(2)}$};
    \end{scope}
    \begin{scope}[color=gray!80!blue]
        \draw[dashed]
        (1.5,1.8) .. controls +(-0.3,-0.3) and +(0,-0.3) .. 
        (1,2)  .. controls +(0,0.3) and +(-0.3,0) ..
        (1.3,2.6) .. controls +(0.2,-0.1) and +(0.2,0.2) .. cycle;
        \node at (1.2,2) {$\times$};
        \node at (1.3,2.4) {$\times$};
        %\node at (1.65,2.73) {\tiny $g^{(3)}$};
    \end{scope}
    \draw[<->,gray] (1.5,1.7) to[bend left] (1.25,1);
    \draw[<->,gray] (2.4,1.2) to[bend right] (1.25,1);
    \draw[<->,gray] (2.4,1.2) to[in=-80,out=160] (1.5,1.7);
    %\node[gray] at (1.95,1.55) {\tiny $\bs G$};
    \begin{scope}[yshift = 3cm, xshift = 1cm]
    \def\L{0.15}
    \def\vx{0.13}
    \def\vy{0.19}
        % P1
        \draw (-0.2,0.13) rectangle (-0.2+4*\L,0.13+7*\L);
        % P3
        \draw (0.65,1.37)++ (0.4*0.13,0.4*0.19) -- (0.65+3*0.13,1.37+3*0.19) --(0.65+4*\L+3*0.13,1.37+3*0.19)-- (0.65+4*\L+0.4*0.13,1.37+0.4*0.19)  -- cycle;
        % P2
        \draw (1.33,0.77) rectangle (1.33+8*\L,0.77+4*\L);
        
        \fill[gray, fill opacity=0.3]
            (1.13,1.18) --
            (1.26,1.37) --
            (1.26,0.76) --
            (1.13,0.57) -- cycle;
        
        \fill[gray, fill opacity=0.3]
            (0.52,1.18) --
            (1.13,1.18) --
            (1.13,0.57) --
            (0.52,0.57) -- cycle;
        
        \fill[gray, fill opacity=0.3]
            (0.65,1.37) --
            (1.26,1.37) --
            (1.13,1.18) --
            (0.52,1.18) -- cycle;
        
        \draw (0.65,1.37) rectangle ++(4*\L,-4*\L);
        \draw (0.52,1.18) rectangle ++(4*\L,-4*\L);
        
        \draw (0.65,1.37) rectangle ++(4*\L,-4*\L);
        \draw (0.52,1.18) rectangle ++(4*\L,-4*\L);
        \draw (0.52,1.18) -- ++ (\vx,\vy);
        \draw (0.52+4*\L,1.18) -- ++ (\vx,\vy);
        \draw (0.52+4*\L,1.18-4*\L) -- ++ (\vx,\vy);
        \draw (0.52,1.18-4*\L) -- ++ (\vx,\vy);

        %\node[color=green!75!black] at (-0.2+2*\L,0.13+3.5*\L) {\tiny $P^{(1)}$};
        %\node[color=red!80!black] at (1.33+4*\L,0.77+2*\L) {\tiny $P^{(2)}$};
        %\node[color=blue!80!black] at (0.65+2*\L+1.4*0.13,1.37+1.7*0.19) {\tiny $P^{(3)}$};
        \node at (1,2.2){\tiny Tucker};
        %\node at (1,-0.7){\tiny $\bs T = \bs G \times_1 P^{(1)} \cdots \times_d P^{(d)}$};
    \end{scope}
    \end{scope}

\end{tikzpicture}
 

%% file: section_2.tex
\section{Discrete Multimarginal Optimal Transport}
\label{seq:mmot}

\subsection{Tensor operations and notations}
For each integer $d \geq 1$, we denote by $\llbracket d \rrbracket$ the set $\{1, \ldots, d\}$ .
Given an integer $d \geq 1$ and any positive integers $n_1, \ldots, n_d$, we call $\bs{T} \in  \bbR^{n_1 \times \cdots \times n_d}$
a $d-$mode tensor. For any two tensors $\bs T \in \mathbb{R}^{n_1 \times \dots \times n_d}$ and $\bs S \in \mathbb{R}^{m_1 \times \dots \times m_e}$, we define the algebraic operations \textit{Frobenius Inner Product} (when $d=e$) and \textit{Outer Product} $\circ$ by $$\textstyle\ps{\bs{S}}{\bs{T}} \defeq \sum_{i_1=1}^{n_1} \dots \sum_{i_d=1}^{n_d} \bs{S}_{i_1 \dots i_d} \bs{T}_{i_1 \dots i_d}, \quad (\bs S \circ \bs T)_{i_1\ldots i_d j_1\ldots j_e} \defeq \bs S_{i_1\ldots i_d}\bs T_{j_1\ldots j_e}.$$ 

We denote by $\otimes$ the Kronecker product between two matrices. Given a subset of indices $\mathcal{I} \subseteq \llbracket d \rrbracket$  and its complement $\bar{\mathcal{I}}$, any tensor can be unfolded into a matrix $\mathbf{T}^{(\mathcal{I})}$ with $n_{\mathcal{I}}=\prod_{i \in \mathcal{I}} n_i$ rows and $n_{\bar{\mathcal{I}}}=\prod_{j \in \bar{\mathcal{I}}} n_j$ columns whose elements are ordered according to a chosen lexicographical order of the modes. In particular, the matricization of $\bs{T}$ with respect to mode $k$ is a matrix denoted $\mathbf{T}^{(k)}$ with $n_k $ rows and $\bar{n}_k = \prod_{\ell \in \llbracket d \rrbracket \setminus \{k\}} n_\ell $ columns. From these unfoldings, numerous tensor-matrix products can be defined. For instance, the product $\bs{T} \times_{\mathcal{I}} \mathbf{M}$ refolds the result of the matrix product $\mathbf{M} \mathbf{T}^{(\mathcal{I})}$. In particular, the mode-$k$ product, denoted by $\times_k$, corresponds to the matrix multiplication $\mathbf{M} \mathbf{T}^{(k)}$. The multilinear rank of a tensor $\bs T$ is $(R_1,\ldots, R_d)$ where $R_k$ is the (column) rank of $\mathbf T^{(k)}$.

A tensor is a probability tensor if its entries are nonnegative and the sum of all entries is $1$. In that case, we denote $\bs T\in\Delta_{n_1\times\cdots\times n_d}$. The Shannon entropy $\mathcal{H}(\bs{T})$ is defined by $\mathcal{H}(\bs{T}) =  -\ps{\bs{T}}{\log(\bs{T}) - \mathbf{1} }$, where the logarithm operator is element-wise, with the convention that $0 \log 0 = 0$.
The marginal of $\bs{T}$ according to the mode $k$ denoted $\mathrm{Marg}_k(\bs T)$ is given by the vector of row sums of ${\bf{T}}^{(k)}$. 
More generally, the marginal of $\bs{T}$ according to the subset of modes $\mathcal{I}$ denoted $\mathrm{Marg}_{\mathcal{I}}(\bs T)$ is given by the vector of row sums of ${\bf{T}}^{(\mathcal{I})}$. 
%\newpage

\subsection{Discrete Multimarginal Optimal Transport}
Given a collection $\mathbf{a}$ of $d$ discrete probability measures $a^{(1)},\ldots, a^{(d)}\in\mathcal P(\mathcal X)$ with $n_1,\ldots, n_d$ atoms respectively, a transport tensor $\bs T$ associated to $\mathbf{a}$ is a probability tensor whose mode-$k$ marginals satisfy  $\mathrm{Marg}_k(\bs T) = a^{(k)}$ for all $k \in \llbracket d \rrbracket$. The set of all such transport tensors is defined as: $$ \Pi (\mathbf{a})\defeq \{\bs T \in\bbR^{n_1\times\cdots\times n_d}_+\ |\ \forall k \in[\![ d]\!],\  \mathrm{Marg}_k(\bs T) = a^{(k)} \}.$$ Consequently,  each mode-$k$ matricization of $\bs T$ implied by these marginal constraints can be interpreted as a coupling between $a^{(k)}$ and a joint measure $b^{(k)}$ supported on the product space $\prod_{\ell \in \llbracket d \rrbracket \setminus \{k\}}\supp(a^{(\ell)})$.

Given a cost tensor $\bs{C}\in \bbR_+^{n_1\times\cdots\times n_d}$, the discrete Multimarginal Optimal Transport problem  \citep{gangbo98, pass2015multi} is the multimarginal generalization of discrete optimal transport, and minimizes the total transport cost $ \ps{\bs C}{\bs T}$ over the set of valid transport tensors associated to $\mathbf{a}$:
\begin{equation}
    \mathrm{MOT}(\mathbf{a},\bs C)=\min_{\bs T \in \Pi(\mathbf{a})} \ps{\bs C}{\bs T}. \label{eq:mmot}
\end{equation} 

Intuitively, MOT can be viewed as solving simultaneously $d$ coupled, relaxed optimal transport problems. For each $k \in \llbracket d \rrbracket$, the $k$-th problem uses the mode-$k$ matricization $\mathbf{C}^{(k)}$ as its cost matrix, with $a^{(k)}$ acting as the row (left) marginal. However, these $d$ sub-problems are fundamentally interdependent: their respective transport plans are not free to vary independently, as they must form valid, structurally consistent matricizations of the same underlying tensor $\bs{T}$. Finally, when $d=2$, MOT exactly recovers the standard two-marginal (bipartite) optimal transport problem.

Since problem~\eqref{eq:mmot} consists of minimizing a continuous linear objective function over a compact convex polytope $\Pi(\mathbf{a})$, the existence of a global minimum is guaranteed. Despite this guarantee, the optimal solution is generally not unique due to the linear nature of the objective function over the polytope. To overcome this lack of uniqueness and mitigate computational challenges, a standard approach is to introduce entropic regularization \citep{benamou2015}.
The entropically regularized multimarginal optimal transport problem with regularization parameter $\varepsilon>0$ is formulated as:
\begin{equation} 
\mathrm{MOT}_{\varepsilon}(\mathbf{a},\bs C)=\min_{\bs{T} \in \Pi(\mathbf{a})} \langle \bs{C}, \bs{T} \rangle - \varepsilon \mathcal{H}(\bs{T}). \label{eq:mmot_entropic} 
\end{equation}
Regularized MOT leads to algorithms based on Bregman projections, the Sinkhorn MOT algorithm \citep{benamou2015} (see Algorithm~\ref{alg:mmsk}) and its greedy and accelerated versions \citep{lin2022accsk}. The first-order Karush-Kuhn-Tucker optimality conditions reveal that the unique solution $\bs T^\varepsilon$ of Eq.~\eqref{eq:mmot_entropic} exhibits a factorized Gibbs-like structure. It can be expressed as a scaling of the kernel tensor $\bs K = \exp(-\bs C / \varepsilon)$ \citep{peyre_optimal_2026}:
$$\textstyle\bs{T}^\varepsilon_{i_1 \dots i_d} = \bs{K}_{i_1 \dots i_d}\big( \prod_{k=1}^d u^{(k)}_{i_k} \big) ,$$ 
where $u^{(k)} \in \mathbb{R}^{n_k}_+$ for $k \in \llbracket d \rrbracket$ are scaling vectors (or scaling factors), which are unique up to a positive multiplicative constant.  The Sinkhorn algorithm needs the full tensor $\bs K$, which cannot be stored as the size of the problem grows. This calls for the use of structured costs that reduce the memory footprint, as proposed in \cite{Haasler2021tree}, \cite{Haasler2021graph}, \cite{Ba2022fft} and \cite{strossner2023}.

%% file: section_3_LRMMOT.tex
\section{Low-Rank Multimarginal Optimal Transport}
\label{seq:lrmmot}
Proofs of the stated properties are deferred to Appendix~\ref{app:proofs}.

\subsection{Low-Rank Transport Tensors}

Contrary to the aforementioned works relying on the structure of $\bs C$, the low-rank formulations proposed by the literature in the standard two-marginal case enforce low-rank constraints on the transport plan and change the space of the problem, redefining the feasible set. In the multi-marginal case, there exist numerous ways to enforce such a low-rank constraint as the rank of a tensor itself has several definitions \citep{Kolda09}. Indeed, as the number of measures $d$ increases, the number of available tensor unfoldings grows exponentially. Each of these unfolded tensors can then be constrained via a low-rank assumption.

To this end, take $\bs I = (\mathcal I_m)_{m=1}^M$ a collection of $M$ subsets of $\llbracket d \rrbracket$ and associated nonnegative ranks $\mathbf r = (r_m)_{m=1}^M$ \citep{cohen93nnrank}. The nonnegative rank satisfies $\mathrm{rank}(A)\le \mathrm{rank}_+(A)$ for any nonnegative matrix $A$. Define $N_m = \prod_{j\in\mathcal I_m} n_j$. We define the set $\Pi_{\mathbf r,\bs I}(\mathbf a) \subseteq \Pi(\mathbf a)$ of admissible couplings $\bs T$ satisfying low-rank assumptions according to $\bs I$ and $\mathbf r$ as $$\Pi_{\mathbf r,\bs I}(\mathbf a) = \Pi(\mathbf a) \cap \{\bs T \in \bbR_+^{n_1\times \cdots \times n_d} \ |\ \mathrm{rank}_+(\mathbf T^{(\mathcal I_m)}) \le r_m \text{ for } 1\le m \le M\}.$$ 
\begin{property}
\label{ppt:setmot_nonempty}
    The set $\Pi_{\mathbf r,\bs I}(\mathbf a)$ is not empty.
\end{property}

When $\bs I$ is a general collections of indices, the rank constraints and the marginals constraints may not be aligned. To mitigate this issue, we impose that the elements of $\bs I$ are \textit{pairwise disjoint}. Without loss of generality, we can in fact impose that $\bs I$ is a \textit{partition} of $\llbracket d\rrbracket$ by adding every mode $k$ not represented in $\bs I$ as a singleton $\{k\}$ and setting the corresponding rank $r_m$ to $n_k$, so that no rank restriction is imposed on these modes.

The following property shows that solving the MOT problem on a collection of product measures $\alpha^{(m)}$ induced by the $a^{(k)}$'s gives a valid transport tensor for $\Pi_{\mathbf r,\bs I}(\mathbf a)$.
\begin{property}
    \label{ppt:rearrange_T}
    Let $\bs I$ be a partition of $\llbracket d \rrbracket$. Let $\alpha^{(m)} = \otimes_{j\in\mathcal I_m} a^{(j)}$ be the product measure of the $a^{(j)}$'s associated to $\mathcal I_m$. If $\bs T \in \bbR_+^{N_1\times\cdots\times N_M}$ belongs to $\Pi(\bs \alpha)$ and satisfies $\mathrm{rank}_+(\mathbf T^{(m)})\le r_m$ for any $1\le m\le M$, then we can reshape $\bs T$ into a tensor $\widetilde{\bs T} \in\bbR_+^{n_1\times\cdots\times n_d}$ belonging to $\Pi_{\mathbf r,\bs I}(\mathbf{a})$.
\end{property}

Given a partition $\bs I$, we can thus consider a new problem on the set of singletons $\bs J=\{\{m\}\}_{m=1}^M$ by taking the product measures as marginals. In the following, we describe the case $\bs I = \{\{k\}\}_{k=1}^d$, and write $\Pi_{\mathbf r,\bs I}(\mathbf a) = \Pi_{\mathbf r}(\mathbf a)$ accordingly.  In order to derive numerical methods solving low-rank problems, we enforce tensor decompositions structure to the transport tensor in the following subsections. 

\subsection{Tucker decomposition}
% I = [[d]] does not imply tucker cf ex 3.12 alexandrov2022nntc

In this section, we restrict the set of admissible couplings $\Pi_{\mathbf r}(\mathbf a)$ to its subset of tensors that admit a nonnegative Tucker decomposition (see \cite{alexandrov2022nntc} for characterizations). The tensor $\bs T$ admits a nonnegative Tucker decomposition with multilinear ranks $\mathbf r$ if there exist $\bs G \in\bbR_+^{r_1\times \cdots \times r_d}$ and $P^{(k)} \in\bbR_+^{n_k \times r_k}$ such that 
\begin{equation}
\label{eq:tucker_decomp}
\bs{T} = \bs{G} \times_{1} P^{(1)}\times_2 \cdots \times_{d} P^{(d)}. 
\end{equation} 
The tensor $\bs T$ can then be stored using $\prod_{k=1}^d r_k +\sum_{k=1}^d n_kr_k$ entries. Such representation can always be made probabilistic by normalizing each $P^{(k)}$ to be a column-stochastic matrix and absorbing the normalization constants into $\bs G$. We thus consider the following subset of $\Pi_{\mathbf r}(\mathbf a)$, with $\mathcal S_{\text{col}}^{n_k\times r_k}$ the set of column-stochastic matrices of size $n_k\times r_k$: $$\Pi_{\mathbf r}^{\rm T}(\mathbf a) = \{\bs T\in\Pi(\mathbf a)\ |\ \exists \bs G\in\Delta_{r_1\times\cdots\times r_d}, \exists P^{(k)}\in\mathcal S_{\text{col}}^{n_k\times r_k}, \bs T = \bs G \times_{1}P^{(1)}\cdots \times_{d} P^{(d)} \}.$$

A particular case of this general decomposition structure appears in \cite{tran2021fcmmot}, where the authors consider a partition $\bs I$ of $\llbracket d\rrbracket$ and introduce the tensor $\widetilde{\bs T} = \mathrm{Marg}_{\mathcal I_1}(\bs T)\circ\cdots\circ\mathrm{Marg}_{\mathcal I_M}(\bs T)$. This outer product has multilinear rank $(1,\ldots,1)$ with respect to the partition $\bs I$, since each block factor $\mathrm{Marg}_{\mathcal I_m}(\bs T)$ spans a one-dimensional subspace.\footnote{Property~\ref{ppt:feasibleTucker} ensures that the infimum in the definition of F-MMOT in \cite{tran2021fcmmot} is attained.} 

\begin{property}
\label{ppt:feasibleTucker}
The set $\Pi_{\mathbf r}^{\rm T}(\mathbf a)$ is compact and not empty.
\end{property}

We can thus define the Tucker MOT problem as \begin{equation}
    \mathrm{TuckerMOT}_{\mathbf r}(\mathbf a,\bs C) = \min_{\bs T\in \Pi_{\mathbf r}^{\rm T}(\mathbf a)} \ps{\bs C}{\bs T}. \label{eq:part_mmot}
\end{equation}
We recover the original MOT problem by setting $r_k = n_k$ for every $k$ so that $\Pi(\mathbf a) \subseteq \Pi_{\mathbf r}^{\mathrm T}(\mathbf a)$. The following properties state bounds on the approximation error of TuckerMOT.

\begin{property}[Approximation error]
    \label{ppt:approx_error}
    Let $R$ be the minimum number of nonnegative rank one factors\footnote{$R = \mathrm{rank}_{\mathrm{CP+}}(\bs T^\star)$ where $\mathrm{rank}_{\mathrm{CP+}}$ is defined in \cite{Lim2009cpplus}} necessary to construct the optimal transport tensor $\bs T^*$ of MOT. For $2\le r\le R$, and for any choice of $\mathbf r = (r_k)_{k=1}^d$ such that $r_{\min} = r$, we have $$0 \le \mathrm{TuckerMOT}_{\mathbf r}(\mathbf{a},\bs C) - \mathrm{MOT}(\mathbf{a},\bs C) \le \|\bs C\|_\infty \ln\Big(\frac{R}{r-1}\Big).$$
\end{property}

In order to state the following bound, let us define the $k$-th entropy number of a set $\mathcal W \subset \mathcal X$ by $\mathcal N_k(\mathcal W) = \inf\{\varepsilon\ |\ \exists z_1,\ldots, z_{2^k}\in \mathcal X, \mathcal W\subset \bigcup_{i=1}^{2^k} \mathrm{B}(z_i,\varepsilon)\}$ where $\mathrm{B}(z_i,\varepsilon)$ is the ball of radius $\varepsilon$ centered at $z_i$ for the metric on $\mathcal X$.

\begin{property}[Approximation error with Lipschitz cost]
    \label{ppt:approx_error2}
    Assume that $\bs C : \mathcal X^d \to \bbR_+$ is $L$-Lipschitz with respect to each variable. Then, for any $r\ge 1$, taking $q=\lfloor r^{1/d}\rfloor$, we have $$0 \le \mathrm{TuckerMOT}_{\mathbf r}(\mathbf{a},\bs C) - \mathrm{MOT}(\mathbf{a},\bs C) \le 2dL\max_{k}\mathcal N_{\lfloor \log_2 q \rfloor}(\supp(a^{(k)})).$$
\end{property}

For the pairwise squared euclidean cost used in Section~\ref{seq:cost_struct}, this second bound has rate $\mathcal O(r^{-1/dm})$ where $m$ is the dimension of $\mathcal X$. To prove Properties~\ref{ppt:approx_error} and \ref{ppt:approx_error2}, we in fact establish the bound in the CP case (detailed below), which is more restrictive.

\paragraph{Introducing bipartite OT for mode $k$.} %Figure~\ref{fig:3_decomp} (right panel) provides a schematic illustration of the Tucker decomposition. 
Under the Tucker decomposition \eqref{eq:tucker_decomp}, 
each measure $a^{(k)}$ can be seen as being transported on a latent measure $g^{(k)}= \mathrm{Marg}_k\big(\bs G\big)$. Indeed, any matricization $\mathbf{T}^{(k)}$ can be rewritten as a product of $P^{(k)}$ (coupling the measure  $a^{(k)}$ with $g^{(k)}$), of the matricization $\mathbf{G}^{(k)}$ of the probability core tensor $\bs G$, and of the Kronecker product of the coupling matrices associated to the other modes \citep{Kolda09}. The column stochasticity of the matrices $P^{(k)}$ implies that if $\bs T\in\Pi(\mathbf{a})$, $$a^{(k)} = \mathrm{Marg}_k\big(\bs T \big)=P^{(k)}\mathbf{G}^{(k)}\big(\textstyle \bigotimes_{\ell\neq k} P^{(\ell)}\big)\tsp \1_{\bar{n}_k} =P^{(k)}  \mathbf{G}^{(k)}\1_{\bar{r}_k} = P^{(k)} \mathrm{Marg}_k\big(\bs G\big) = P^{(k)} g^{(k)} .$$

As $P^{(k)}$ is not a transport plan, we define $Q^{(k)}=P^{(k)}\diag(g^{(k)})$. Remark that $Q^{(k)}\1_{r_k}=a^{(k)}$ and that $(Q^{(k)})\tsp\1_{\bar{n}_k}=\diag(g^{(k)}) (P^{(k)})\tsp \1_{\bar{n}_k} =g^{(k)}$ as $P^{(k)}$ is column-stochastic. Thus, $Q^{(k)}$ can be interpreted as a transport plan between the measure $a^{(k)}$ and the latent measure $g^{(k)}$. 

\begin{figure}[h]
    \centering
    \input{figure_tucker}
    \caption{Transport from $a^{(1)}$ to $b^{(1)}$ through intermediary latent measures.}
    \label{fig:draw_transport_tucker}
\end{figure}
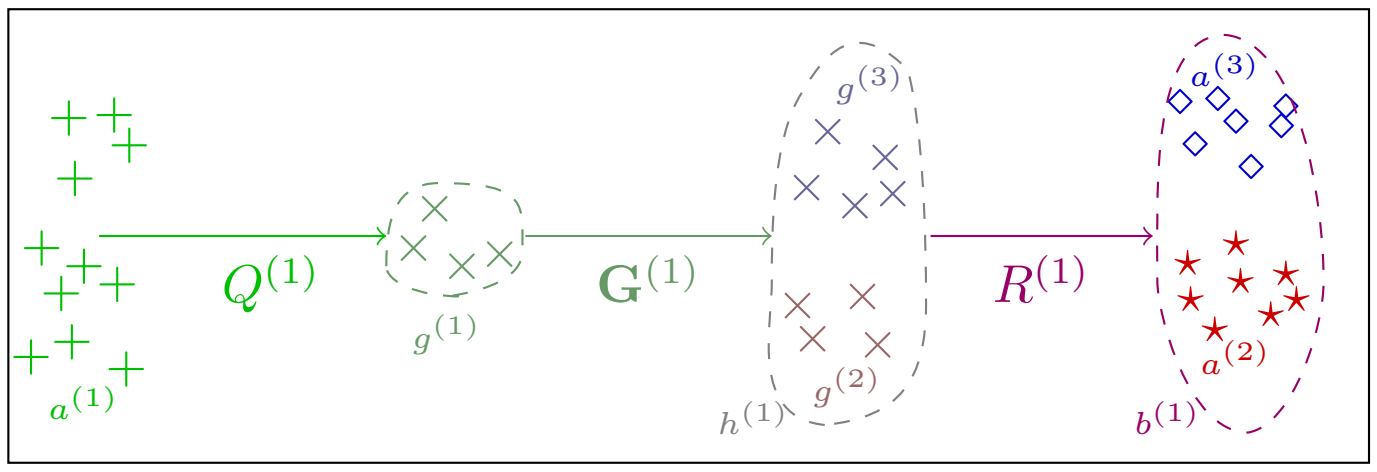

Conversely, the joint measure $b^{(k)}$ is coupled to $\bs G$ through the matrix
$(\bigotimes_{\ell\ne k} P^{(\ell)})$. Denoting  $h^{(k)} = (\mathbf{G}^{(k)})\tsp \1_{\bar{r}_k}$ the latent joint measure, we can define $R^{(k)} =  (\bigotimes_{\ell\ne k} P^{(\ell)})\diag(h^{(k)})$.
The matrix $R^{(k)}$ is a transport plan from $h^{(k)}$ to $b^{(k)}$. This yields the hierarchical sequence of Figure~\ref{fig:draw_transport_tucker}.

\paragraph{Soft clustering.} Low-rank multimarginal optimal transport can be interpreted as a soft clustering assignment. Indeed, each entry of the latent measure $g^{(k)}$ can be associated with a cluster. For each mode $k$, the $i_k$-th row of the matrix $Q^{(k)}$ can be normalized to encode the probabilities that $x_{i_k}^{(k)}$ belongs to each cluster $s$. In Figure~\ref{fig:tucker_barycenters} (Section~\ref{seq:exp}), one can see how latent measures are related to clusters.

We now present another classical tensor decomposition of $\bs T$ more restrictive than the Tucker decomposition.

\subsection{CP decomposition}

\paragraph{CP decomposition.} 
CP decomposition imposes that the core tensor $\bs G$ in the Tucker decomposition is hyperdiagonal  (see Figure~\ref{fig:3_decomp}, left panel). As a consequence, $r_1=\cdots=r_d=r$, and the core tensor can be represented by a probability vector $g\in\Delta_r$. The tensor $\bs T$ can then be expressed as a sum of rank-1 tensors formed by columns of $P^{(k)}$, with $p^{(k)}_s$ denoting the $s$-th column of $P^{(k)}$: $$\textstyle\bs T = \sum_{s=1}^r g_s p^{(1)}_s \circ\cdots \circ p^{(d)}_s.$$
The tensor $\bs T$ can then be stored using $r+r\sum_{k=1}^d n_k$ entries.

We define the CP formulation of the MOT problem as the minimization of the transport cost subject to the constraint that the nonnegative CP rank of the transport tensor $\bs T$ is upper bounded. Define the feasible set of CP-rank-$r$ admissible transport tensors as $$\Pi_r^{\text{CP}}(\mathbf{a})\defeq\Pi(\mathbf{a}) \cap \{\bs T \in\bbR_+^{n_1\times\cdots\times n_d} \ |\ \mathrm{rank}_{\text{CP}+}(\bs T) \le r\}.$$ 
\begin{property}
    \label{ppt:cp_close}
    The set $\Pi_r^{\textup{CP}}(\mathbf{a})$ is compact and non empty. 
\end{property}
 The CPMOT problem then reads \begin{equation}
    \mathrm{CPMOT}_r(\mathbf{a},\bs C) = \min_{\bs T \in \Pi_r^{\text{CP}}(\mathbf{a})}\ps{\bs C}{\bs T}.\label{eq:cplrmmot}
\end{equation} 

\paragraph{Introducing bipartite OT for mode $k$.}
As in the Tucker formulation, each measure $a^{(k)}$ is mapped to the latent measure $g^{(k)}$ through a transport plan $Q^{(k)}=P^{(k)}\diag(g^{(k)})$. However as $\bs G$  is hyperdiagonal, each marginal $\mathrm{Marg}_k(\bs G)$ is represented by the same vector $g\in\Delta_r$. Thus, there is only one common latent measure $g$. As a consequence, the transport plan associated to mode $k$ is given by $Q^{(k)}=P^{(k)}\diag(g)$. In a similar way, the latent joint measure $h^{(k)}$ of the CP core over all latent modes except $k$ reduces to a measure $\bar{g}^{(k)}$ whose only non-null entries are associated to the $r$ elements of the hyperdiagonal of $\bs G$. Consequently, $R^{(k)}=(\bigotimes_{\ell\neq k}P^{(\ell)})\diag(\bar{g}^{(k)})$. When computing $R^{(k)}$ only $r$ rows of the Kronecker product are necessary. Finally, we can represent the transport from $g$ to $b^{(k)}$ by a single $r\times \bar n_k$ matrix. 

\paragraph{Regularization.} We consider a regularized version of problem~\eqref{eq:cplrmmot}. Let $h(\mathbf{Q}) = \sum_k \mathcal H(Q^{(k)})$. Taking the infimum over $g\in\Delta_r^*$ and $Q^{(k)}\in\Pi(a^{(k)},g)$, the regularized problem reads \begin{equation}   \mathrm{CPMOT}_{r,\varepsilon}(\mathbf{a},\bs C)= \inf_{\mathbf Q, g} \ps{\bs C}{\textstyle \sum_{s=1}^r g_s p^{(1)}_s \circ\cdots \circ p^{(d)}_s} -\varepsilon h(\mathbf{Q}), \label{eq:lrmmotQ_e}\end{equation}

%% file: figure_tucker.tex
\begin{tikzpicture}
        \draw (0,0) rectangle (9,3);

        \begin{scope}[color=green!75!black]
        \node at (0.50,1.30) {$+$};
        \node at (0.22,1.42) {$+$};
        \node at (0.72,1.18) {$+$};
        \node at (0.35,1.12) {$+$};
        \node at (0.42,0.8) {$+$};
        \node at (0.15,0.7) {$+$};
        \node at (0.78,0.62) {$+$};
        \node at (0.8,2.1) {$+$};
        \node at (0.7,2.3) {$+$};
        \node at (0.4,2.28) {$+$};
        \node at (0.44,1.88) {$+$};
        \node at (0.5,0.4) {\tiny $a^{(1)}$};
        \draw[->] (0.6,1.5) -- (2.5,1.5)node[pos=0.6, anchor=north]{$Q^{(1)}$};
        \end{scope}

        \begin{scope}[color=gray!80!green, xshift = 0.9cm, yshift=0.1cm]
        \node at (2.10,1.20) {$\times$};
        \node at (1.78,1.32) {$\times$};
        \node at (2.35,1.28) {$\times$};
        \node at (1.92,1.58) {$\times$};
        \node at (2,0.75) {\tiny $g^{(1)}$};
        \draw[dashed]
        (1.6,1.2) .. controls +(0,0.3) and +(-0.3,0) ..
        (1.95,1.75) .. controls +(0.6,0) and +(0,0.3) .. 
        (2.5,1.3) .. controls +(0,-0.3) and +(-0.3,0) .. 
        (2.1,1)  .. controls +(-0.3,0) and +(0,-0.1) .. cycle;
        \draw[->] (2.52,1.4) -- (4.15,1.4) node[midway, anchor=north]{$\mathbf G^{(1)}$};
        \end{scope}
        
        \begin{scope}[color=gray!80!blue, yshift=0.5cm, xshift=3.5cm]
        \node at (2.10,1.20) {$\times$};
        \node at (1.78,1.32) {$\times$};
        \node at (2.35,1.28) {$\times$};
        \node at (2.30,1.52) {$\times$};
        \node at (1.92,1.69) {$\times$};
        \node at (2.2,2) {\tiny $g^{(3)}$};
        \end{scope}

         \begin{scope}[color=gray!80!red, yshift=-0.5cm, xshift=3.5cm]
        \node at (2.15,1.6) {$\times$};
        \node at (1.82,1.32) {$\times$};
        \node at (2.25,1.28) {$\times$};
        \node at (1.72,1.54) {$\times$};
        \node at (2.05,1) {\tiny $g^{(2)}$};
        \draw[dashed,gray]
        (1.9,0.75) .. controls +(-0.5,0.05) and +(0.0,-0.3) ..
        (1.55,1.7) .. controls +(0,0.3) and +(-0.2,-0.3) .. 
        (1.75,2.95) .. controls +(0.2,0.35) and +(-0.3,0.3) .. 
        (2.45,3.1)  .. controls +(0.1,-0.4) and +(0.0,0.5) .. 
        (2.57,1.55)  .. controls +(0.0,0-0.5) and +(0.5,0.05) .. cycle;
        \node[gray] at (1.43,0.8) {\tiny $h^{(1)}$};
        \draw[purple!80!blue, ->] (2.6,2) -- (4.07,2) node[midway, anchor=north]{$R^{(1)}$};
        \end{scope}

         % mu_2
    \begin{scope}[color=red!80!black, xshift = 4.6cm, yshift=0.8cm]
        \node at (3.55,0.40) {$\star$};
        \node at (3.22,0.28) {$\star$};
        \node at (3.75,0.18) {$\star$};
        \node at (3.38,0.08) {$\star$};
        \node at (3.85,0.45) {$\star$};
        \node at (3.20,0.52) {$\star$};
        \node at (3.52,0.65) {$\star$};
        \node at (3.92,0.28) {$\star$};
        \node at (3.52,-0.1) {\tiny $a^{(2)}$};
        
    \end{scope}

    % mu_3
    \begin{scope}[color=blue!80!black, xshift = 7.6cm, yshift=-0.3cm]
        \node at (0.25,2.40) {$\diamond$};
        \node at (0.82,2.52) {$\diamond$};
        \node at (0.52,2.55) {$\diamond$};
        \node at (0.62,2.25) {$\diamond$};
        \node at (0.15,2.68) {$\diamond$};
        \node at (0.40,2.7) {$\diamond$};
        \node at (0.85,2.65) {$\diamond$};
        \node at (0.45,2.9) {\tiny $a^{(3)}$};
        \draw[dashed,purple!80!blue]
        (0.45,3.13) .. controls +(0.5,-0.05) and +(0.0,0.5) ..
        (1.1,1.5) .. controls +(0,-0.3) and +(0.3,0.05) ..
        (0.6,0.5) .. controls +(-0.4,0) and +(0,-0.2) ..
        (0,1.5).. controls +(0,0.6) and +(-0.5,0) ..
        cycle;
        \node[purple!80!blue] at (0.07,0.6) {\tiny $b^{(1)}$};
    \end{scope}
    
    \end{tikzpicture}

%% file: section_4_pairwise.tex
\section{Incorporating cost structure}
\label{seq:cost_struct}
In this Section, we describe a class of cost tensors having a complete graph structure that allows us to define an \textit{explicit} CPMOT (eCPMOT), where the support of the latent measure $g$ is instantiated. This formulation yields an upper bound for CPMOT.

\subsection{Explicit CPMOT}

\paragraph{Conditionally of negative type costs.} Following \cite{houry2026}, we define costs that are conditionally of negative type (CNT). Such a cost $c:\mathcal X \times \mathcal X \to \bbR_+$ is CNT if it is continuous, symmetric, and verifies $c(x,x) = 0$ for all $x\in\mathcal X$, and for any $n\in\bbN$, any $x_1,\ldots, x_n\in \mathcal X$, any $\lambda_1,\ldots,\lambda_n$ with $\sum_{i=1}^n \lambda_i =0$ we have $\sum_{i=1}^n \sum_{j=1}^n \lambda_i\lambda_j c(x_i,x_j)\le 0.$ Equivalently, a cost $c$ is CNT if there exists a real Hilbert space $\mathcal H$ and a continuous mapping $\varphi : \mathcal X \to \mathcal H$ verifying $c(x,x') = \|\varphi(x)-\varphi(x')\|_{\mathcal H}^2$ for all $x,x'\in\mathcal X$. 

\paragraph{Pairwise costs.} Let $a^{(1)},\ldots,a^{(d)}$ be $d$ measures on $\mathcal P(\mathcal X)$ with atoms $(x^{(k)}_{i_k})_{i_k=1}^{n_k}$ respectively. Let $c : \mathcal X \times \mathcal X\to \bbR_+$ be a CNT cost, with its associated Hilbert space $\mathcal H$ and mapping $\varphi$. We consider pairwise costs of the form $$\bs C(x_{i_1}^{(1)},\ldots,x_{i_d}^{(d)})=\sum_{k<l}\| \varphi(x_{i_k}^{(k)}) -\varphi(x_{i_l}^{(l)})\|_{\mathcal H}^2.$$

Let us denote by $z=(z_s)_{s=1}^r$ the support of the latent measure $g$. Denote by $B^{(k)}(\varphi(z)) = [\|\varphi(x_{i_k}^{(k)})-\varphi(z_s)\|^2_{\mathcal H}]_{i_k,s} \in\bbR_+^{n_k\times s}$ the cost matrix between $a^{(k)}$ and $g$. Introducing $z$ in each cost pair yields an upper bound for CPMOT.

\begin{Theorem}
\label{thm:expl_upperbound}
    The following problem is an upper bound of CPMOT. 
\begin{equation}
   \mathrm{eCPMOT}_{r}(\mathbf a, \bs C) = d\inf_{\mathbf{Q},g,\varphi(z)}\sum_{k=1}^d \ps{B^{(k)}}{Q^{(k)}},\label{eq:multi_expl_pb}
\end{equation}
where the minimization is taken over $g\in\Delta_r^\star$, $\varphi(z)\in\mathcal H^r$, and each $Q^{(k)}\in\Pi(a^{(k)},g)$.
\end{Theorem}

In Theorem~\ref{thm:expl_upperbound}, the vector $g$ is required to have non zero entries. This does not pose a problem for numerical methods, since $\mathrm{CPMOT}_r(\mathbf{a},\bs C) \le \mathrm{CPMOT}_{r-1}(\mathbf{a},\bs C)$.

Adding regularization yields the following upper bound of  $\mathrm{CPMOT}_{r,\varepsilon}$ (Eq.~\eqref{eq:lrmmotQ_e}): \begin{equation}
 \mathrm{eCPMOT}_{r,\varepsilon}(\mathbf a, \bs C) = d \inf_{\mathbf{Q},g,\varphi(z)}\sum_{k=1}^d \ps{B^{(k)}}{Q^{(k)}}-\frac{\varepsilon}{d}\mathcal H(Q^{(k)}).\label{eq:multi_expl_pb_e}
\end{equation}

\subsection{Optimality gap induced by eCPMOT} For a low-rank transport plan decomposition $(\mathbf{Q},g)$, we denote by $F(\bs T)$ the objective from eCPMOT (Eq.~\eqref{eq:multi_expl_pb}). Let $A^{(k)}_s$ be independent random variables of respective law $A^{(k)}_s\sim \sum_{i_k=1}^{n_k} p^{(k)}_{i_ks}\delta_{\varphi(x_{i_k}^{(k)})}$, where $p^{(k)}_{i_ks}=Q^{(k)}_{i_ks}/g_s$. 
\begin{Theorem}
\label{thm:bound_mmot}
    Denote by $\mu^{(k)}_s=\bbE[A_s^{(k)}]$. Let us denote $$V(\bs T) = \sum_{s=1}^r g_s \sum_{k=1}^d \Var(A^{(k)}_s),\qquad
          B(\bs T)=\frac 12 \sum_{s=1}^r g_s  \sum_{k\ne l} \|\mu_s^{(k)}-\mu_s^{(l)}\|_{\mathcal H}^2.$$ Then, the different objectives of Eq.~\eqref{eq:cplrmmot} and Eq.~\eqref{eq:multi_expl_pb} can be expressed as $$\ps{\bs C}{\bs T} = (d-1)V(\bs T) + B(\bs T),\qquad F(\bs T) = dV(\bs T)+B(\bs T).$$
\end{Theorem}

\paragraph{Empirical interpretation.} The transport from $a^{(k)}$ to $g$ induces $r$ measures $(a^{(k)}_s)_{s=1}^r$ sharing the support of $a^{(k)}$ such that $a^{(k)} = \sum_{s=1}^r g_sa^{(k)}_s$. The quantity $V(\bs T)$ measures the within-component dispersion of each measure $a^{(k)}_s$ around its mean $\mu^{(k)}_s$. We refer to Figure~\ref{fig:unit_roots} in Appendix~\ref{app:exp_d2} for an example showing positions of the $\mu^{(k)}$'s.

\begin{corollary}
     If $\bs T^\star$ is an optimal transport plan for CPMOT (Eq.~\eqref{eq:cplrmmot}) and $\widehat{\bs{T}}$ is an optimal transport plan for eCPMOT (Eq.~\eqref{eq:multi_expl_pb}), then $$0\le \ps{\bs C}{\widehat{\bs{T}}}- \ps{\bs C}{\bs T^\star} \le  V(\bs T^\star) - V(\widehat{\bs T}).$$
\end{corollary}

Hence, the reduction in within-component dispersion resulting from the explicit barycentric optimization provides an upper bound on the excess transport cost of $\widehat{\bs{T}}$. In particular, if $V(\bs T^\star) = V(\widehat{\bs T})$, then $\widehat{\bs T}$ is also optimal for the original low-rank problem.

\paragraph{Statistical convergence.} In the bipartite case, low-rank constraints on the transport plan were originally introduced by \cite{forrow19fc} to overcome the curse of dimensionality in optimal transport. Indeed, given two measures $a$ and $b$, and an empirical measure $\hat a_n = \frac 1n \sum_{i=1}^n \delta_{X_i}$ based on an i.i.d. sample $X_1,\ldots,X_n\sim a\in\mathcal P(\bbR^m)$, we have \citep{peyre_optimal_2026}: $$\bbE |\mathrm{W}_2(\hat a_n,b) - \mathrm{W}_2(a,b)| \le \bbE[\mathrm{W}_2(\hat a_n,a)] = \mathcal O (n^{-\frac 1m}).$$ \cite{bayraktar2024} (Theorem 5.3) show that curse of dimensionality applies in the multimarginal case. \cite{forrow19fc} (Theorem 4) establish a $\mathcal O (n^{-\frac12})$ rate for their proxy problem. In Theorem~\ref{thm:statconv}, we formulate an analogue result. 

\begin{Theorem}
\label{thm:statconv}
    For every $1\le k \le d$, let $a^{(k)}$ be a measure supported on the unit ball where the space containing the supports has dimension $m<\infty$. Let $\hat a_{n_k}^{(k)}$ be an empirical distribution comprising $n_k$ i.i.d. samples from $a^{(k)}$. Then there exists $C > 0$ such that with probability at least $1-\delta$, $$|\mathrm{eCPMOT}_{r}(\hat{\mathbf a},\bs C)-\mathrm{eCPMOT}_{r}(\mathbf a,\bs C)| \le Cd\sum_{k=1}^d\sqrt{\frac{r^3m\log (r)+\log (d/\delta)}{n_k}}.$$
\end{Theorem}

Up to rescaling, we also get a $\mathcal O (n^{-\frac 12})$ rate for all compactly supported measures.

%% file: section_5_algorithms.tex
\section{Algorithms}
\label{seq:algorithms}

In this section, we derive two algorithms for solving Eq.~\eqref{eq:cplrmmot} and one algorithm for solving Eq.~\eqref{eq:part_mmot}. The first two algorithms presented below are designed for a general cost tensor $\bs C$, and operations involving this tensor should be adapted to take advantage of any structure in $\bs C$. In our experiments, we use the class of structured costs introduced in Section~\ref{seq:cost_struct}. When $d=2$, we recover the methods proposed in \cite{halmos_low-rank_2024}, \cite{scetbon_low-rank_2021} and \cite{forrow19fc}; see Appendix~\ref{app:exp_d2} for details. When using the low-rank solvers together with the pairwise euclidean cost structure, the per-iteration cost of the Sinkhorn MOT algorithm, $\mathcal O(dn ^d)$, is reduced to $\mathcal O(drn)$ in the CP case and to $\mathcal O(drn+dr^d)$ in the Tucker case. When the cost is structured according to a graphical model \citep{strossner2023} of maximum degree $\alpha$, \impcp\ has a per-iteration cost of $\mathcal O(drn^\alpha)$.

\subsection{Mirror Descent scheme optimizing the Tucker formulation}
\label{seq:tuck_md}
    
For each mode $k$, the matrix $Q^{(k)}$ is a transport plan from the marginal $a^{(k)}$ to the latent marginal $g^{(k)}$. At the latent level, the core tensor $\bs G$ couples the latent marginals $g^{(1)},\ldots,g^{(d)}$. Hence, the Tucker representation decomposes the MOT problem into a first level of $d$ semi-relaxed transports $a^{(k)}\to g^{(k)}$ followed by a latent MOT problem. Costs associated to these sub-problems are derived from the following equalities $$\ps{\bs C}{\bs G \times_{k=1}^d P^{(k)}} = \ps{\bs C \times_{k=1}^d (P^{(k)})\tsp}{\bs G} = \ps{[\bs C\times_{\ell\ne k} (P^{(\ell)})\tsp\times_{\ell\ne k} \bs G] \diag(g^{(k)})^{-1}}{Q^{(k)}}.$$ Thus, constructing the mode-$k$ subcost directly from this expression requires access to the full cost tensor $\bs C$. Denote by $\bs C_{\bs G}$ and $C_{Q^{(k)}}$ the cost tensors associated with the latent problem and the mode-$k$ semi-relaxed transport problem, respectively. Since $\bs C_{\bs G}=C_{Q^{(k)}}\diag(g^{(k)})\times_k (P^{(k)})\tsp,$ we can approximate $C_{Q^{(k)}}$ without explicitly forming $\bs C$ by approximately inverting $(P^{(k)})\tsp$. In particular, using the Moore-Penrose pseudoinverse $A^+$ of a matrix $A$ \citep{Barata2011pinverse}, we can consider  $C_{Q^{(k)}}\approx[\bs C_{\bs G}\times_k((P^{(k)})\tsp)^+]\diag(g^{(k)})^{-1}$.

\cite{scetbon_low-rank_2021} explain how the classical regularized OT problem can be seen as a mirror descent step on the unregularized problem, justifying mirror descent schemes in low-rank approaches. After adding a regularization term $\mathcal H(\bs G)$ to TuckerMOT (Eq.~\eqref{eq:part_mmot}), we follow these steps an derive an algorithm (\tucker) using a mirror descent scheme in Appendix~\ref{app:tucker_md}.

\subsection{Mirror Descent scheme optimizing the CP formulation}
\label{seq:mmot_md}
We also perform a mirror descent in the CP formulation. In this case, we add entropic regularization on both $\bf Q$ and $g$. The algorithm \impcp\ for Eq.~\eqref{eq:cplrmmot} is derived in Appendix~\ref{app:md_impcp}. The algorithm can take advantage of graphically structured costs to improve its time and space complexity.

\subsection{Alternating minimization of the explicit CP formulation}
\label{seq:mmot_expl}

To solve Eq.~\eqref{eq:multi_expl_pb_e}, we use an alternating minimization scheme, successively minimizing over $(\mathbf{Q},g)$ and over $\varphi(z)$.

\begin{property}
\label{ppt:min_z}
    When the $Q^{(k)}$'s and $g$ are fixed, the minimizing $\varphi(z)$ is defined by $$\varphi(z^\star_s) = \frac{1}{dg_s}\sum_k\sum_{i_k}Q^{(k)}_{i_ks}\varphi(x^{(k)}_{i_k}).$$
\end{property}

 Thus, the arithmetic mean of the latent hubs minimizes $\varphi(z)$.

\paragraph{Minimization over $(\mathbf{Q},g)$.} When the support $ \varphi(z)$ of the intermediate measure is fixed, \cite{benamou2015} provide the \textsc{SinkhornBary} Algorithm \ref{alg:sk_mmbary} to find the minimizing weights of $g$ and couplings $Q^{(k)}$.

This alternating minimization scheme results in Algorithm~\ref{alg:blrmmot} (\expcp). In Appendix~\ref{seq:ummot_expl}, we present an extension of this method to the unbalanced setting \citep{Beier2022ummot}.

\begin{algorithm}[h]
    \caption{Explicit CPMOT}
    \label{alg:blrmmot}
    \begin{algorithmic}[1] 
        \Function{ECPMOT}{$r$, $(x^{(k)})_{k=1}^d$, $\mathbf{a}$, $\varepsilon$} 
             \State $\varphi(z)\gets$ \textsc{d-kmeans-z}($(\varphi(x^{(k)}))_k,r$)
            \State $[C^{(k)}]_{k=1}^d \gets \big[ [\|\varphi(x^{(k)}_{i_k})-\varphi(z_s)\|^2_{\mathcal H}]_{i_ks} \big]_{k=1}^d$
    \For{$t\ge 0$} \Comment{until convergence}
    
    \State $\mathbf{Q},g\gets$ \textsc{SinkhornBary}($(C^{(k)})_{k=1}^d,\mathbf{a},\varepsilon/d$)
    \State $\varphi(z) \gets [\frac{1}{dg_s}\sum_{k=1}^d Q_{i_ks}^{(k)} \varphi(x_{i_ks}^{(k)})]_{s=1}^r$
    \State $[C^{(k)}]_{k=1}^d \gets \big[ [\|\varphi(x^{(k)}_{i_k})-\varphi(z_s)\|^2_{\mathcal H}]_{i_ks} \big]_{k=1}^d$
    \EndFor
    \State \textbf{return} $\mathbf{Q},g$
    \EndFunction
    \end{algorithmic}
\end{algorithm}

%% file: section_6_Appendix.tex
\label{seq:exp}

\paragraph{Time gain and optimality.} In Table~\ref{tab:mmot_time_opt}, we compare the methods derived in Section~\ref{seq:algorithms} with the standards and greedy Sinkhorn MOT algorithms in terms of transport cost and running time. We consider that convergence has been reached when the relative change in the objective is below $10^{-6}$. We sample $(n_1,n_2,n_3,n_4) = (25,50,25,50)$ points in $\bbR^2$ from four Gaussian mixtures (detailed in Appendix~\ref{app:exp_dd}). We set $\varepsilon = 0.005$ for the CP formulation, and $\varepsilon = 0.04$ for the Tucker formulation. For \tucker, we report results using both Sinkhorn \eqref{alg:mmsk} (\textsc{msk}) and Greedy Sinkhorn \eqref{alg:greedy_mmsk} (\textsc{g-msk}) for the core subproblem. The cost tensors are normalized.

\begin{table}[h]
\caption{Comparison of the transport costs and running times of the methods and Sinkhorn MOT. 2D Gaussian mixtures with $(n_1,n_2,n_3,n_4) = (25,50,25,50)$. Values are averaged over 30 runs}
\label{tab:mmot_time_opt}
\begin{center}
\begin{tabular}{c l c c c c}
\hline
$r$ & Method & Transport cost & Marg. err & Iterations & Time (s)\\
\hline
& \textsc{msk} & $0.186 \pm 0.022$ & $10^{-8}$ & $246 \pm 62$ & $20.3 \pm 6.3$ \\
& \textsc{g-msk} & $0.186 \pm 0.022$ & $10^{-7}$ & $580 \pm 122$ & $8.2 \pm 1.9$ \\
\hline
\multirow{4}{*}{$2$}
& \expcp & $\gr{0.229} \pm \gr{0.022}$ & $10^{-8}$ & $7 \pm 3$ & $\gr{0.2} \pm \gr{0.1}$ \\
& \impcp & $0.258 \pm 0.028$ & $10^{-8}$ & $2007 \pm 690$ & $2.8 \pm 1.9$ \\
& \tucker\ \textsc{g-msk}  & $0.232 \pm 0.022$ & $10^{-17}$ & $532 \pm 415$ & $60.2 \pm 0.1$ \\
& \tucker\ \textsc{msk} & $0.233 \pm 0.022$ & $10^{-17}$ & $492 \pm 377$ & $60.2 \pm 0.3$ \\ 
\hline
\multirow{4}{*}{$4$}
&\expcp & $ \gr{0.202} \pm \gr{0.022}$ & $10^{-8}$ & $20 \pm 9$ & $\gr{0.4} \pm \gr{0.2}$ \\
&\impcp & $0.231 \pm 0.028$ & $10^{-8}$ & $2379 \pm 568$ & $4.4 \pm 2.2$ \\
&\tucker\ \textsc{g-msk} &  $0.231 \pm 0.022$ & $10^{-17}$ & $1434 \pm 786$ & $59.6 \pm 1.8$ \\ 
&\tucker\ \textsc{msk} & $0.231 \pm 0.022$ & $10^{-17}$ & $1677 \pm 970$ & $56.9 \pm 6.6$ \\ \hline
\multirow{4}{*}{$10$}
&\expcp &$\gr{0.191} \pm \gr{0.022}$ & $10^{-8}$ & $56 \pm 36$ & $\gr{0.7} \pm \gr{0.4}$ \\
&\impcp & $0.219 \pm 0.023$ & $10^{-8}$ & $2511 \pm 432$ & $5.1 \pm 1.7$ \\
&\tucker\ \textsc{g-msk}  &  $0.235 \pm 0.022$ & $10^{-17}$ & $759 \pm 219$ & $60.3 \pm 0.4$ \\
&\tucker\ \textsc{msk} &  $0.234 \pm 0.022$ & $10^{-17}$ & $875 \pm 295$ & $60.1 \pm 0.1$ \\
\end{tabular}
\end{center}
\end{table}

We observe that the running time is reduced with the CP formulation, particularly for the explicit method. Moreover, the low-rank methods achieve transport costs close to those obtained with Sinkhorn while maintaining low marginal errors, even for a very small rank such as $r=4$. The cost decreases as $r$ increases, except for \tucker, for which there was insufficient time to perform as many iterations in the $r=10$ case. Overall, Greedy Sinkhorn performs better than Sinkhorn. However, for a certain range of small $r^d$ values in the Tucker core subproblem, the standard Sinkhorn routine performs better, allowing Tucker to perform more iterations within the allotted time.

\begin{figure}[h]
    \centering
    \begin{tikzpicture}
\begin{axis}[
    width=8cm,
    height=4cm,
    xtick=\empty,
    ytick=\empty,
    %axis equal image,
    legend style={
        at={(1.02,1)},
        anchor=north west,
        font=\small,legend columns=2,
},]

\addplot[
    only marks,
    mark=*,
    mark size=1.8pt,
    opacity=0.15,orange
]
table [x=x, y=y] {tucker_barycenters/X_1.txt};
\addlegendentry{$x^{(1)}$}

\addplot[
    only marks,
    mark=triangle,
    mark size=4pt,orange
]
table [x=x, y=y] {tucker_barycenters/Z_1.txt};
\addlegendentry{$\mu^{(1)}$}
% Mode 2
\addplot[
    only marks,
    mark=*,
    mark size=1.8pt,
    opacity=0.1,green!70!black
]
table [x=x, y=y] {tucker_barycenters/X_2.txt};
\addlegendentry{$x^{(2)}$}

\addplot[
    only marks,
    mark=pentagon,
    mark size=4pt,green!70!black
]
table [x=x, y=y] {tucker_barycenters/Z_2.txt};
\addlegendentry{$\mu^{(2)}$}
% Mode 3
\addplot[
    only marks,
    mark=*,
    mark size=1.8pt,
    opacity=0.1,blue
]
table [x=x, y=y] {tucker_barycenters/X_3.txt};
\addlegendentry{$x^{(3)}$}

\addplot[
    only marks,
    mark=star,
    mark size=4pt,blue
]
table [x=x, y=y] {tucker_barycenters/Z_3.txt};
\addlegendentry{$\mu^{(3)}$}
\end{axis}
\end{tikzpicture}
    \caption{Supports of the $a^{(k)}$'s and $g^{(k)}$'s. $(n_1,n_2,n_3) = (30,50,60) $, $(r_1,r_2,r_3)= (4,5,6)$}
    \label{fig:tucker_barycenters}
\end{figure}
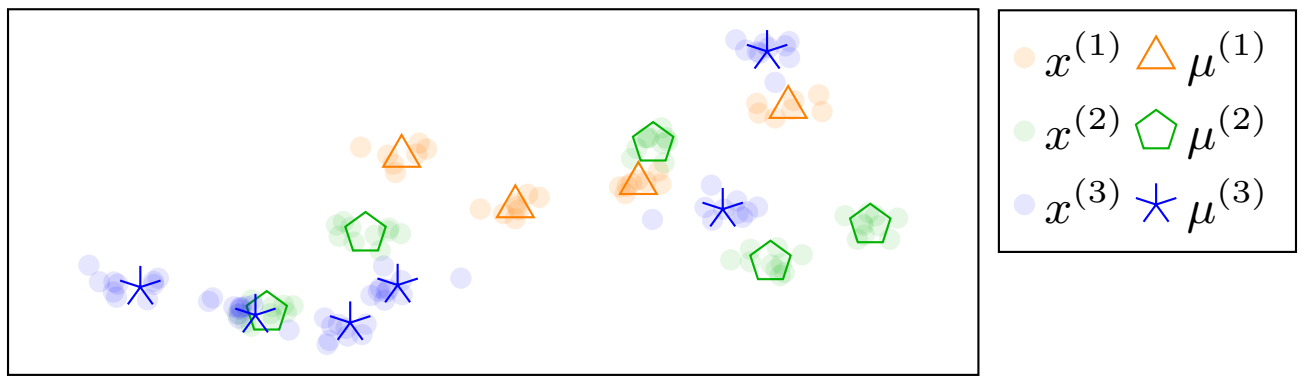

\paragraph{Latent barycenters.} In Figure~\ref{fig:tucker_barycenters}, we solve TuckerMOT for a problem involving $d=3$ measures, where the $k$-th measure $a^{(k)}$ is supported on $r_k$ blobs. In the Tucker formulation, each measure $a^{(k)}$ is associated with a latent measure $g^{(k)}$, whose support points can be recovered as $\mu^{(k)}_s = \sum_{i_k=1}^{n_k} Q^{(k)}_{i_ks}x^{(k)}_{i_k}/g^{(k)}_s$. We observe that the resulting latent measures accurately represent the $a^{(k)}$'s.

\paragraph{Estimating ensemble flows on a hidden Markov chain.} In Appendix~\ref{app:exp}, we use \impcp\ in the experiment from Section 6 in \cite{Haasler2021tree} (uncoupled case). We study a population of $N$ agents walking on a $n\times n$ grid according to a lazy random walk on a the associated graph during $\tau$ time steps. To generate observations, we use binary sensors distributed over the grid. The objective is to reconstruct the random walk from the observation of the sensors. We observe that \impcp\ correctly tracks the center of mass of the population, and that the estimation gets better when the rank $r$ increases.

\section{Conclusion}
We defined a low-rank multimarginal optimal transport problem and showed that it encompasses several formulations found in the literature. We instantiated this framework with several algorithms. In particular, the \expcp\ algorithm finds transport tensors with a low transport cost two orders of magnitude faster than the usual Sinkhorn algorithm.

\newpage
\paragraph{Acknowledgments}

This work was supported by the French Agence Nationale de
la Recherche (ANR), under grant ANR-23-CE23-0004 (project ODD).

\paragraph{Reproducibility} Source code for reproducing LRMOT experiments is provided at \texttt{https://github.com/clemejo/lrmot}.

\paragraph{AI use statement} 
In this work, we used generative AI tools to check language and grammar, and to give wording suggestions.

\bibliography{LOT}
\bibliographystyle{iclr2027_conference}
\newpage
\appendix

\section*{Appendices}
\addcontentsline{toc}{section}{Appendices}

\startcontents[appendices]
\printcontents[appendices]{}{1}{}

\section{Notations}

\begin{itemize}
    \item $d$ the number of measures in a MOT problem,
    \item $\bs T \in\bbR_+^{n_1\times\cdots\times n_d}$ a transport tensor,
    \item $a^{(k)}\in\mathcal P (\mathcal X)$ the $k$-th measure in a MOT problem,
    \item $n_k$ the size of the support of $a^{(k)}$,
    \item $b^{(k)}$ the product measure $(\mathbf{T}^{(k)})\tsp\1_{n_k}$,
    \item $\bar n_k = \prod_{\ell\ne k} n_\ell$ the size of the support of $b^{(k)}$,
    \item $(x^{(k)})_{i_k= 1}^{n_k}\in \mathcal X^{n_k}$ the support of $a^{(k)}$, 
    \item $r_k$ the rank constraint associated with the $k$-th measure,
    \item $\Pi(\mathbf a)$ the set of admissible transport tensors with marginals $\mathbf a = (a^{(k)})_{k=1}^d$,
    \item $\bs C\in\bbR_+^{n_1\times\cdots\times n_d}$ a cost tensor,
    \item $\mathcal H$ the entropy function $\mathcal H(X) = -\ps{X}{\log(X)-\1}$,
    \item $\varepsilon>0$ a regularization parameter,
    \item $\bs I = (\mathcal I_m)_{m=1}^M$ a collection of subsets of $\llbracket d\rrbracket$,
    \item $N_m = \prod_{j\in\mathcal I_m} n_j$,
    \item $\alpha^{(m)} = \otimes_{j\in\mathcal I_m} a^{(j)}$,
    \item  $\bs G \in\bbR_+^{r_1,\ldots, r_d}$ a Tucker core,
    \item $\Delta_{n_1,\ldots,n_d}$ the probability tensors of size $n_1\times\cdots\times n_d$,
    \item $g^{(k)} = \mathrm{Marg}_k(\bs G)\in\bbR_+^{r_k}$ a latent measure,
    \item $P^{(k)}\in\mathcal S_{\text{col}}^{n_k\times r_k}$ a column-stochastic matrix,
    \item $Q^{(k)} = P^{(k)}\diag(g^{(k)})\in\bbR_+^{n_k\times r_k}$ a transport plan from $a^{(k)}$ to $g^{(k)}$, $\mathbf Q = (Q^{(k)})_{k=1}^d$,
    \item $\varphi : \mathcal X \to \mathcal H$ a mapping for the CNT cost $c$,
    \item $\mu^{(k)}$ hubs associated with $x^{(k)}$,
    \item $z\in\mathcal X$ the latent hubs positions.
\end{itemize}

\input{appendix_d2_exp}

\section{Proof details}
\label{app:proofs}
\subsection{Proof of Property \ref{ppt:setmot_nonempty}}
\begin{proof}
    Consider $\bs T = a^{(1)} \circ\cdots \circ a^{(d)}$ so that an entry of $\bs T$ is $$\bs T_{i_1\dots i_d} = \prod_{k=1}^d a_{i_k}^{(k)}.$$ Let $\mathcal I_m$ be a set of modes. In order to show that $\bs T \in \Pi_{\mathbf r,\bs I}(\mathbf a)$, we need to show that $\mathrm{rank}_+(\mathbf T^{(\mathcal I_m)}) \le r_m$. If $k$ correspond to the indices $(i_j)_{j\in\mathcal I_m}$ and $l$ to the indices $(i_j)_{j\in\bar{\mathcal I}_m}$, then $$\mathbf{T}^{(\mathcal I_m)}_{kl} = \prod_{j=1}^d a_{i_j}^{(j)} = \prod_{j\in\mathcal I_m} a_{i_j}^{(j)} \times \prod_{j\in\bar{\mathcal I}_m} a_{i_j}^{(j)}.$$ So, defining $u = \otimes_{j\in\mathcal I_m} a^{(j)}$ and $v = \otimes_{j\in\bar{\mathcal I}_m} a^{(j)}$, we have $\mathbf{T}^{(\mathcal I_m)} = uv\tsp$. 
    
    Thus, $\mathrm{rank}_+(\mathbf{T}^{(\mathcal I_m)}) = 1 \le r_m$.
\end{proof}

\subsection{Proof of Property \ref{ppt:rearrange_T}} \begin{proof}
    For each block $\mathcal I_m = \{j_1 < \cdots < j_J\}$ define a bijection $$\phi_m : \prod_{j\in\mathcal I_m} \llbracket n_j\rrbracket\to \llbracket N_m\rrbracket, $$ for example $\phi_m(i_1,\ldots,i_J) = 1 + \sum_{\ell = 1}^J (i_{j_\ell}-1) \prod_{q=\ell+1}^{J}n_{j_q}$. We define the ungrouped tensor $\widetilde{\bs T}\in\bbR_+^{n_1\times\cdots\times n_d}$ by $$\widetilde{\bs T}_{i_1,\ldots,i_d} = \bs T_{\phi_1((i_j)_{j\in\mathcal I_1}),\ldots,\phi_M((i_j)_{j\in\mathcal I_M})}.$$ Let $j\in\mathcal I_m$, then as $\bs I$ is a partition of $\llbracket d \rrbracket$, taking the $j$-th marginal of the ungrouped tensor is equivalent to first taking the $m$-th marginal of the grouped tensor $\bs T$, and then the marginal of the result on the mode corresponding to $j$ (that we call $j$ with an abuse of notation): $$\mathrm{Marg}_j(\widetilde{\bs T}) = \mathrm{Marg}_{j}(\mathrm{Marg}_m(\bs T)) = \mathrm{Marg}_{j}(\otimes_{\ell\in\mathcal I_m}a^{(\ell)}) = a^{(j)},$$ so $\bs T \in\Pi(\mathbf a)$. Now, the two matrices $\widetilde{\bf T}^{(\mathcal I_m)}$ and $\mathbf T^{(m)}$ differ only by a permutation of the rows and columns. Nonnegative rank is invariant under such permutations, and thus $$\mathrm{rank}_+(\widetilde{\bf T}^{(\mathcal I_m)}) = \mathrm{rank}_+(\mathbf T^{(m)})\le r_m. $$ Finally, $\widetilde{T} \in \Pi_{\mathbf r,\bs I}(\mathbf a)$.    
\end{proof}

\subsection{Proof of Property \ref{ppt:feasibleTucker}} \begin{proof}
$\bullet\ $ \textbf{Not empty} : The independent coupling $\bs T=a^{(1)}\circ\cdots\circ a^{(d)}$ belongs to $\Pi_{\mathbf r}^{\rm T}(\mathbf a)$. Indeed, setting $P^{(k)}=\begin{bmatrix}p^{(k)} & \cdots & p^{(k)}\end{bmatrix}\in\mathbb R_+^{n_k\times r_m},$ each $P^{(k)}$ is column-stochastic. For any $\bs G\in\Delta_{r_1\times\cdots\times r_d}$, $$\bs T=\bs G\times_1P^{(1)}\cdots\times_d P^{(d)}.$$
Consequently, $\Pi_{\mathbf r}^{\rm T}(\mathbf a)$ is nonempty.

$\bullet\ $ \textbf{Compact} : The parameter space $$\Delta_{r_1\times\cdots\times r_d} \times \prod_{k=1}^d \mathcal S_{\text{col}}^{n_k\times r_m}$$ is compact and the map $$(\bs G, P^{(1)},\ldots, P^{(d)}) \mapsto \bs G \times_1 P^{(1)}\cdots \times_d P^{(d)}$$ is continuous. Therefore, its image is compact and intersecting it with the compact set $\Pi(\mathbf a)$ gives that $\Pi_{\mathbf r}^{\rm T}(\mathbf a)$ is compact.
\end{proof}

\subsection{Proof of Property \ref{ppt:approx_error}} 
\label{app:proof_bound1}\begin{proof}
We follow ideas from the proof of Proposition 1 in \cite{Scetbon2022stats}. Let $\bs T^\star$ be an optimal solution of $\mathrm{MOT}(\mathbf a,\bs C)$. Since $\bs T^\star\ge0$, it admits a nonnegative rank-one decomposition. Since $\bs T^\star$ has total mass one, this decomposition can be written as $$\bs T^\star=\sum_{s=1}^R\lambda_sr_s^{(1)}\circ\cdots\circ r_s^{(d)}$$ with $\lambda\in\Delta_R^*$ and $r_s^{(k)}\in\Delta_{n_k}$. Without loss of generality, assume $\lambda_1\ge\lambda_2\ge\cdots\ge\lambda_R.$ Since $\lambda\in\Delta_R$, we have $k\lambda_k\leq\sum_{s=1}^k\lambda_s\leq1,$ and hence $\lambda_k\leq\frac1k.$ Define the mass $$\Lambda_r=\sum_{s=r}^R\lambda_s$$ and, for every $k\in\llbracket d\rrbracket$, the normalized tail marginal $$\bar r^{(k)}=\frac1{\Lambda_r}\sum_{s=r}^R\lambda_s r_s^{(k)}.$$

Since every $r_s^{(k)}$ belongs to $\Delta_{n_k}$, we have $\bar r^{(k)}\in\Delta_{n_k}$. Now, we construct a transport tensor $\widetilde{\bs T}_r\in\Pi_r(\mathbf{a})$: $$\widetilde{\bs T}_r=\sum_{s=1}^{r-1}\lambda_sr_s^{(1)}\circ\cdots\circ r_s^{(d)}+\Lambda_r\bar r^{(1)}\circ\cdots\circ\bar r^{(d)}.$$

Indeed, the $k$-th marginal of $\widetilde{\bs T}_r$ is $$\mathrm{Marg}_k(\widetilde{\bs T}_r)=\sum_{s=1}^{r-1}\lambda_s r_s^{(k)}+\Lambda_r\bar r^{(k)}=a^{(k)}.$$

It remains to verify that $\widetilde{\bs T}_r$ belongs to $\Pi_{\mathbf r}^{\rm T}(\mathbf a)$. The tensor $\widetilde{\bs T}_r$ has a CP decomposition with $r$ nonnegative components. Therefore, for every mode $k$, the unfolding $\widetilde{\mathbf T}_r^{(k)}$ admits a nonnegative rank-$r$ decomposition, and hence $\mathrm{rank}_+(\widetilde{\mathbf T}_r^{(k)})\le r.$

Write $$\widetilde{\bs T}_r=\sum_{s=1}^{r}\mu_s u_s^{(1)}\circ\cdots\circ u_s^{(d)},$$

where the last component corresponds to the tail marginal. For each mode $k$, define

$$\textstyle p_s^{(k)}= u_s^{(k)}\in\Delta_{n_k}.$$

Complete these $r$ columns with arbitrary probability vectors $p_{r+1}^{(k)},\ldots,p_{r_k}^{(k)}$ and set $P^{(k)}=\big[p_1^{(k)}\cdots p_{r_k}^{(k)}\big].$ Thus $ P^{(k)}\in\mathcal S_{\mathrm{col}}^{n_k\times r_m}.$ Then, $$\widetilde{\bs T}_r=\bs G\times_1P^{(1)}\cdots\times_d P^{(d)},$$ where $\bs G_{i_1\dots i_d} = \mu_s$ if $i_1=\cdots=i_d = s$ and $s\le r$, and $0$ otherwise. Therefore $\widetilde{\bs T}_r\in\Pi_{\mathbf r}^{\mathrm T}(\mathbf a).$ Since $\mathrm{TuckerMOT}_{\mathbf r}$ minimizes over a subset of $\Pi(\mathbf a)$, $$0\le\mathrm{TuckerMOT}_{\mathbf r}(\mathbf a,\bs C)-\mathrm{MOT}(\mathbf a,\bs C).$$

Furthermore, using $\widetilde{\bs T}_r$ as a feasible candidate, \begin{align*}\mathrm{TuckerMOT}_{\mathbf r}(\mathbf a,\bs C)-\mathrm{MOT}(\mathbf a,\bs C) &\le\ps{\bs C}{\widetilde{\bs T}_r}-\ps{\bs C}{\bs T^\star}\\
&\le \ps{\bs C}{\Lambda_r\bar r^{(1)}\circ \cdots\circ \bar r^{(d)}}-\ps{\bs C}{\sum_{s=r}^R \lambda_s r_s^{(1)}\circ \cdots \circ r_s^{(d)}}\\
&\le\|\bs C\|_\infty\Lambda_r.\end{align*}

Using $\lambda_k \le 1/k$ and $\sum_{s=r}^R \tfrac{1}{s}\le \int_{r-1}^R \tfrac{1}{x}\,\mathrm{d}x = \ln(R/(r-1))$, we get $$\mathrm{TuckerMOT}_{\mathbf r,\bs I}(\mathbf a,\bs C)-\mathrm{MOT}(\mathbf a,\bs C) \le \|\bs C\|_\infty\ln\Big(\frac{R}{r-1}\Big),$$ which proves the result.
\end{proof}

\subsection{Proof of Property \ref{ppt:approx_error2}} \begin{proof}
Again, we follow ideas from the proof of Proposition 2 in \cite{Scetbon2022stats}. Let $\bs T^\star$ be an optimal solution of $\mathrm{MOT}(\mathbf a,\bs C)$. Define $m = \lfloor \log_2 q\rfloor$ and $\varepsilon_k = \mathcal N_m(\supp(a^{(k)})$. By definition of $\varepsilon_k$, there exist\footnote{Here we should in fact take $\varepsilon_k + \eta$ and then let $\eta \to 0$, but we keep $\varepsilon_k$ for simplicity.} at most $2^m \le q$ points $z_{k,1},\ldots, z_{k,q}\in\mathcal X$ such that $$\supp (a^{(k)})\subset \bigcup_{i=1}^q \mathrm{B}(z_{k,i},\varepsilon_k).$$ Thus, we have a partition (with potentially empty sets) of $\supp(a^{(k)})$ as $\bigcup_{i=1}^q \mathcal X_{k,i}$ with $\mathrm{diam}(\mathcal X_{k,i}) \le 2\varepsilon_k$. 

For every $J=(j_1,\ldots,j_d)\in \llbracket q\rrbracket^d$, define the weight $w_J = \sum_{\substack{x_k\in\mathcal X_{k,j_k}\\ 1\le k\le d}} \bs T^\star (x_1,\ldots, x_d)$. 

If $w_J >0$, we define the distribution $$p^{(k)}_J(x_k)=\frac{1}{w_J}\sum_{\substack{x_\ell \in\mathcal X_{\ell, j_\ell} \\ \ell\ne k}} \bs T^\star(x_1,\ldots,x_d)$$ where $x_k\in\mathcal X_{k,j_k}$, and $p^{(k)}_J(x_k)=0$ elsewhere. 

Now, we can construct a transport tensor $\tilde {\bs T}_r\in\Pi_r(\mathbf a)$: $$\tilde {\bs T}_r = \sum_{\substack{J\in \llbracket q\rrbracket^d \\ w_J > 0}} w_J p_J^{(1)}\circ \cdots \circ p_J^{(d)}.$$ Indeed, $q^d\le r$, so $\mathrm{rank}_{\mathrm{CP}+}(\tilde {\bs T}_r) \le r$, and for any $x_k\in\supp(a^{(k)})$, $$\mathrm{Marg}_k(\tilde {\bs T}_r)(x_k)=\sum_J w_Jp_J^{(k)}(x_k) = \sum_{\substack{J\\ x_k\in\mathcal X_{k,j_k}}} \sum_{\substack{x_\ell \in\mathcal X_{\ell,j_\ell}\\ \ell\ne k}} \bs T^\star(x_1,\ldots,x_d) = a^{(k)}(x_k).$$ 

By the Lipschitz assumption, we have $$\sup_{\mathbf x,\mathbf y\in \mathcal{X}_{1,j_1}\times\cdots\times \mathcal X_{d,j_d}} |\bs C(\mathbf x) - \bs C(\mathbf y)| \le L\sum_{k=1}^d \mathrm{diam}(\mathcal X_{k,i_k})\le 2dL\max_k \varepsilon_k.$$

For every $J$ such that $w_J > 0$, the restriction of $\bs T^\star$ to $\mathcal{X}_{1,j_1}\times\cdots\times \mathcal X_{d,j_d}$ and $w_Jp_J^{(1)}\circ\cdots \circ p_J^{(d)}$ both have the same mass $w_J$ and the same support. Thus, \begin{align*}
    \ps{\mathcal C}{\tilde {\bs T}_r}-\ps{\bs C}{\bs T^\star} \le \sum_J (2dL\max_k \varepsilon_k) w_J = 2dL\max_k \varepsilon_k.
\end{align*}

We can complete the $p_J^{(k)}$'s with zeros to cover the Tucker case as in Appendix~\ref{app:proof_bound1}.

We believe a factor 2 is missing from the Proposition 2 in \cite{Scetbon2022stats}, as the claim "$d_{\mathcal X}(x,y)\le \varepsilon_{\mathcal X}$" should be "$d_{\mathcal X}(x,y)\le 2\varepsilon_{\mathcal X}$".
\end{proof}

\subsection{Proof of Property \ref{ppt:cp_close}} \begin{proof}
$\bullet$ \gr{Non empty} : The independent coupling $\bs T = a^{(1)}\circ \cdots \circ a^{(d)}$ is in $\Pi_r^{\text{CP}}(\mathbf{a})$.

$\bullet$ \gr{Compact} : 
Proposition 6.2 from \cite{Lim2009cpplus} gives us that $ \{\bs T \in\bbR_+^{n_1\times\cdots\times n_d} \ |\ \mathrm{rank}_{\text{CP}+}(\bs T) \le r\}$ is closed. This closure comes from the choice of the \textit{nonnegative} rank and the \textit{inequality} constraint. Therefore, intersecting it with the compact set $\Pi(\mathbf a)$ gives that $\Pi_r^{\text{CP}}(\mathbf{a})$ is compact.
\end{proof}

\subsection{Proof of Theorem \ref{thm:expl_upperbound}} \label{app:barymmot} For any $s$, let $(X_s^{(k)})_{k=1}^d$ be independent random variables with respective laws $X_s^{(k)} \sim \sum_{i_k=1}^{n_k} p^{(k)}_{i_ks}\delta_{\varphi(x_{i_k}^{(k)})}$ which are denoted by $A^{(k)}_s$ in the main article. Therefore, using the joint law of the $(X^{(k)}_s)_{k=1}^d$, the total cost in Eq.~\eqref{eq:multi_expl_pb} rewrites 
\begin{align}
     \sum_{i_1,\ldots,i_d}\mathcal C_{i_1,\ldots,i_d} \sum_{s=1}^r \frac{Q_{i_1s}^{(1)}\times\cdots\times Q^{(d)}_{i_ds}}{g_s^{d-1}} &= \sum_{i_1,\ldots,i_d}\sum_{s=1}^r g_s \sum_{k<l}\|\varphi(x^{(k)}_{i_k})-\varphi(x_{i_l}^{(l)})\|_{\mathcal H}^2  p^{(1)}_{i_1s}\times\cdots\times p^{(d)}_{i_ds} \nonumber\\
     &= \sum_{s=1}^rg_s \sum_{k<l}\sum_{i_1,\ldots,i_d} f_{kl}(x_{i_1}^{(1)},\ldots,x_{i_d}^{(d)})P((X_s^{(1)},\ldots,X_s^{(d)})=(x_{i_1}^{(1)},\ldots,x_{i_d}^{(d)}))\nonumber\\
     &=\sum_{s=1}^rg_s \sum_{k<l}\bbE\big[f_{kl}(X_s^{(1)},\ldots,X_s^{(d)})\big]\nonumber\\
     &= \sum_{s=1}^rg_s \bbE\big[\sum_{k<l} \| X_s^{(k)} - X_s^{(l)}\|^2_{\mathcal H} \big]\label{eq:multi_esp}
\end{align}
where $f_{kl}(x_{i_1}^{(1)},\ldots,x_{i_d}^{(d)})=\|\varphi(x^{(k)}_{i_k})-\varphi(x_{i_l}^{(l)})\|_{\mathcal H}^2$.

 In order to introduce the hubs positions, we denote $\bar X=\tfrac 1d \sum_k X^{(k)}_s$ and state the following lemma:

 \begin{lemma}\label{lem:mmot} For all $z\in\mathcal X$, the following equalities holds:
      $$\sum_{k<l} \| X_s^{(k)} - X_s^{(l)}\|^2_{\mathcal H}  = d \sum_{k=1}^d \|X_s^{(k)} -\varphi(z)\|^2_{\mathcal H} -d^2\|\bar X -\varphi(z)\|^2 _{\mathcal H},$$ $$\bbE\big[ \|X_s^{(k)}-\varphi (z)\|_{\mathcal H}^2\big] = \sum_{i_k=1}^{n_k} \|X_s^{(k)}-\varphi (z)\|_{\mathcal H}^2 p^{(k)}_{i_ks}$$
 \end{lemma}

In particular, $\sum_{k<l} \| X_s^{(k)} - X_s^{(l)}\|^2_{\mathcal H}  \le d\sum_{k=1}^d \|X_s^{(k)} -\varphi(z)\|^2_{\mathcal H}$. Plugging this inequality in Eq.~\eqref{eq:multi_esp} and using the second equality yields \begin{align}
    \eqref{eq:multi_esp}&\le d\sum_{s=1}^r g_s\bbE\big[\sum_{k=1}^d\| X_s^{(k)}-\varphi(z_s)\|_{\mathcal H}^2\big]   =d\sum_{s=1}^r\sum_{k=1}^d \sum_{i_k=1}^{n_k} Q_{i_ks}^{(k)}
\|\varphi(x_{i_k}^{(k)})-\varphi(z_s)\|^2_{\mathcal H},\label{eq:multi_expl}\end{align} so that the explicit CPMOT is $$
    \inf_{\mathbf{Q},g}d\sum_{s=1}^r\sum_{k=1}^d \sum_{i_k=1}^{n_k} Q_{i_ks}^{(k)}
\|\varphi(x_{i_k}^{(k)})-\varphi(z_s)\|^2_{\mathcal H}. $$

\textit{Proof of Lemma~\ref{lem:mmot}}. First, \begin{align*}
    \sum_{k<l} \|X_s^{(k)}-X_s^{(l)}\|_{\mathcal H}^2 &= (d-1)\sum_{k=1}^d \|X_s^{(k)}\|_{\mathcal H}^2 -2\sum_{k<l} \ps{X_s^{(k)}}{X_s^{(l)}}_{\mathcal H}\\
    &= d\sum_{k=1}^d \|X_s^{(k)}\|_{\mathcal H}^2 -d^2\|\bar X\|_{\mathcal H}^2.
\end{align*}

The last equality comes from the fact that $\|\sum_kX_s^{(k)}\|_{\mathcal H}^2=\sum_k \|X_s^{(k)}\|_{\mathcal H}^2+2\sum_{k<l} \ps{X_s^{(k)}}{X_s^{(l)}}_{\mathcal H}$.

Then, \begin{align}
    \sum_{k=1}^d \| X_s^{(k)}-\bar X \|_{\mathcal H}^2&=\sum_{k=1}^d \| X_s^{(k)}\|_{\mathcal H}-2\ps{\sum_{k=1}^dX_s^{(k)}}{\bar X}_{\mathcal H}+d\|\bar X \|_{\mathcal H}^2\nonumber\\ & =\sum_{k=1}^d \|X_s^{(k)}\|_{\mathcal H}^2 -d\|\bar X\|_{\mathcal H}^2.\label{eq:appendix1}
\end{align}

So that \begin{equation}
     \sum_{k<l} \|X_s^{(k)}-X_s^{(l)}\|_{\mathcal H}^2 = d\sum_{k=1}^d \| X_s^{(k)}-\bar X \|_{\mathcal H}^2.
\end{equation}

For any $z\in\mathcal X$, \begin{align*}
    \sum_{k=1}^d \| X_s^{(k)}-\bar X \|_{\mathcal H}^2 &= \sum_{k=1}^d \| X_s^{(k)}-\varphi(z)-(\bar X-\varphi(z)) \|_{\mathcal H}^2\\
    &=\sum_{k=1}^d \| X_s^{(k)}-\varphi(z)\|_{\mathcal H}^2-d\|\bar X-\varphi(z)\|_{\mathcal H}^2\quad \text{as in Eq.~\eqref{eq:appendix1}}.
\end{align*}

Finally, $$\forall z\in \mathcal X,\quad \sum_{k<l} \|X_s^{(k)}-X_s^{(l)}\|_{\mathcal H}^2 = d\sum_{k=1}^d \| X_s^{(k)}-\varphi(z)\|_{\mathcal H}^2-d^2\|\bar X-\varphi(z) \|_{\mathcal H}^2.$$

Since $\|X_s^{(k)}-\varphi(z_s)\|_{\mathcal H}^2$ is a function of $X_s^{(k)}$ only, its expectation depends only on the marginal distribution of $X_s^{(k)}$. Hence, the second equality:   \begin{equation*}
     \bbE\big[\|X_s^{(k)}-\varphi(z_s)\|_{\mathcal H}^2\big] =\sum_{i_k=1}^{n_k}  \|\varphi(x_{i_k}^{(k)})-\varphi(z_s)\|_{\mathcal H}^2p_{i_ks}^{(k)}.\qedhere
\end{equation*}

\subsection{Proof of Theorem \ref{thm:bound_mmot}} \label{app:mmotbound}
We denote by $X^{(k)}_s$ the quantity $A^{(k)}_s$ from the main article. From $\varphi(z_s^\star) = \frac 1d \sum_{k=1}^d \bbE [X^{(k)}_s]$, we have for any $k$, $$\bbE [X^{(k)}_s]-\varphi(z_s^\star) = \frac 1d \sum_{l=1}^d\bbE [X^{(k)}_s]-\bbE [X^{(l)}_s].$$ Thus, denoting $\mu_s^{(k)} = \bbE[X_s^{(k)}]$, the expectation from Eq.~\eqref{eq:multi_expl}, it rewrites \begin{align*}
     \bbE \big[\|X_s^{(k)}-\varphi (z_s^\star) \|_{\mathcal H}^2\big] &= \|\mu_s^{(k)}-\varphi(z_s^\star)\|_{\mathcal H}^2+\Var(X^{(k)}_s)\\
     &=\frac{1}{d^2}\|\sum_{l\ne k}\mu_s^{(k)}-\mu_s^{(l)} \|_{\mathcal H}^2+\Var(X^{(k)}_s).
\end{align*}

So that the objective of explicit CPMOT \eqref{eq:multi_expl} is $$F(\bs T)= \sum_s g_s \big(d\sum_k\Var(X^{(k)}_s) + \frac 1d\sum_k\| \sum_{l\ne k} \mu_s^{(k)}-\mu_s^{(l)}\|_{\mathcal H}^2\big).$$

On the other hand, expectation from Eq.~\eqref{eq:multi_esp} is \begin{align*}
    \bbE\big[\|X_s^{(k)}-X_s^{(l)}\|_{\mathcal H}^2\big] &= \Var (X^{(k)}_s) + \Var(X_s^{(l)}) + \|\mu_s^{(k)}-\mu_s^{(l)}\|_{\mathcal H}^2.
\end{align*} and the total transport cost \eqref{eq:multi_esp} is $$\ps{\bs C}{\bs T} = \sum_s g_s \big( (d-1)\sum_{k=1}^d\Var (X^{(k)}_s)+\frac 12 \sum_{k\ne l} \|\mu_s^{(k)}-\mu_s^{(l)}\|_{\mathcal H}^2\big).$$

\begin{lemma}
\label{lem:lathubs}
    The following equality holds, $$\frac 1d\sum_k\| \sum_{l\ne k} \mu_s^{(k)}-\mu_s^{(l)}]\|_{\mathcal H}^2 = \frac 12 \sum_{k\ne l} \|\mu_s^{(k)}-\mu_s^{(l)}\|_{\mathcal H}^2.$$
\end{lemma}

\begin{proof}
    Let $\bar \mu_s = \frac 1d\sum_{k=1}^d \mu_s^{(k)}$ For fixed $s$ and $k$, we have $$\sum_{l\ne k} (\mu_s^{(k)}-\mu_s^{(l)}) = d(\mu_s^{(k)}-\bar\mu_s).$$ Therefore, $$\frac 1d \sum_{k=1}^d \| \sum_{l\ne k} \mu_s^{(k)}-\mu_s^{(l)}]\|_{\mathcal H}^2 = d \sum_{k=1}^d\|\mu^{(k)}_s-\bar\mu_s \|_{\mathcal H}^2.$$ Using that $\sum_{k=1}^d\|\mu^{(k)}_s-\bar\mu_s \|_{\mathcal H}^2 = \frac 1d \sum_{k<l} \|\mu_s^{(k)}-\mu_s^{(l)}]\|_{\mathcal H}^2 $ (Eq.~\eqref{eq:appendix1}), we get $$\frac 1d \sum_{k=1}^d \| \sum_{l\ne k} \mu_s^{(k)}-\mu_s^{(l)}]\|_{\mathcal H}^2=\sum_{k<l} \|\mu_s^{(k)}-\mu_s^{(l)}]\|_{\mathcal H}^2 = \frac 12 \sum_{k\ne l}\|\mu_s^{(k)}-\mu_s^{(l)}]\|_{\mathcal H}^2.$$
\end{proof}

Combining the expressions for $F(\bs T)$ and $\ps{\bs C}{\bs T}$, and the Lemma~\ref{lem:lathubs}, we get Theorem~\ref{thm:bound_mmot}.

\subsection{Proof of Theorem \ref{thm:statconv}} \begin{proof}
Let us reformulate Eq.~\eqref{eq:multi_expl_pb} as a Wasserstein barycenter problem in the space $\mathcal H$. Define $A^{(k)} = \varphi_\#a^{(k)} = \sum_{i_k=1}^{n_k} a_{i_k}^{(k)}\delta_{\varphi(x_{i_k}^{(k)})}$, so that we have $$\inf_{Q^{(k)}\in\Pi(a^{(k)}, g)} \ps{B^{(k)}}{Q^{(k)}} = \mathrm{W}_{2,\mathcal H}^2(A^{(k)},g).$$ Thus we have $$\mathrm{eCPMOT}_{r}(\mathbf a, \bs C) = d\inf_{g\in\mathcal D_r(\mathcal H)} \sum_{k=1}^d \mathrm{W}_{2,\mathcal H}^2(A^{(k)},g) $$ where $\mathcal D_r(\mathcal H)$ is the set of probability measures supported on at most $r$ points of $\mathcal H$ (here $g$ is treated as a measure rather than a probability vector of $\Delta_r$).

    Now, \begin{align*}
        |\mathrm{eCPMOT}_r(\hat{\mathbf a},\bs C) - \mathrm{eCPMOT}_r(\mathbf a,\bs C)| & = d|\inf_g  \mathrm{W}_{2,\mathcal H}^2(\hat A_{n_k}^{(k)},g)-  \inf_g\mathrm{W}_{2,\mathcal H}^2(A^{(k)},g)|\\
        &\le d \sup_g\big| \sum_{k=1}^d  \mathrm{W}_{2,\mathcal H}^2(\hat A_{n_k}^{(k)},g)- \mathrm{W}_{2,\mathcal H}^2(A^{(k)},g) \big|\\
        &\le d \sum_{k=1}^d  \sup_g |\mathrm{W}_{2,\mathcal H}^2(\hat A^{(k)}_{n_k},g)-\mathrm{W}_{2,\mathcal H}^2(A^{(k)},g)|.
    \end{align*}

    By Theorem 4 from \citep{forrow19fc}, for every $1\le k\le d$, there exists $C_k>0$ such that with probability at least $1-\delta / d$, $$ \sup_g |\mathrm{W}_{2,\mathcal H}^2(\hat A^{(k)}_{n_k},g)-\mathrm{W}_{2,\mathcal H}^2(A^{(k)},g)| \le C_k\sqrt{\frac{r^3m\log r+\log(d/\delta)}{n_k}}$$

    Denoting $E_k$ the event "$\sup_g |\mathrm{W}_{2,\mathcal H}^2(\hat A^{(k)}_{n_k},g)-\mathrm{W}_{2,\mathcal H}^2(A^{(k)},g)| \le C_k\sqrt{\frac{r^3m\log r+\log(d/\delta)}{n_k}}$", we have $$\textstyle\mathbf P(\bigcap_{k=1}^d E_k) = 1 - \mathbf P(\bigcup_{k=1}^d \overline{E_k}) \ge 1-\sum_{k=1}^d \mathbf P(\overline{E_k})\ge 1-d\frac{\delta}{d} = 1-\delta.$$

    Defining $C = \max_k C_k$, we conclude that with probability at least $1-\delta$, $$ |\mathrm{eCPMOT}_r(\hat{\mathbf a},\bs C) - \mathrm{eCPMOT}_r(\mathbf a,\bs C)| \le Cd \sum_{k=1}^d\sqrt{\frac{r^3m\log r+\log(d/\delta)}{n_k}}.$$
\end{proof}

\subsection{Proof of Property \ref{ppt:min_z}} \begin{proof}
    For a fixed index $s$, the terms depending on $\varphi(z_s)=u$ are $$F_s(u) = \sum_{k=1}^d \sum_{i_k=1}^{n_k} Q_{i_ks}^{(s)} \|\varphi(x_{i_k}^{(k)})-u\|_{\mathcal H}^2.$$

Expanding the squared norm gives

$$F_s(u)=
\sum_{k,i_k}Q^{(k)}_{i_ks}
\|\varphi(x^{(k)}_{i_k})\|_{\mathcal H}^2-2\ps{\sum_{k,i_k}Q^{(k)}_{i_ks}\varphi(x^{(k)}_{i_k})}{u}_{\mathcal H}+\big(\sum_{k,i_k}Q^{(k)}_{i_ks}\big)\|u\|_{\mathcal H}^2.$$
We have $\sum_{k,i_k}Q^{(k)}_{i_ks}=dg_s$, so $$F_s(u)=\mathrm{const}-2\ps{\sum_k g_s\bbE[X^{(k)}_s]}{u}_{\mathcal H}+dg_s\|u\|_{\mathcal H}^2.$$

Completing the square,
$$F_s(\varphi(z_s))=\mathrm{const}+dg_s\Big\|u-\frac{\sum_k g_s\bbE[X^{(k)}_s]}{dg_s}\Big\|_{\mathcal H}^2-\mathrm{const}_2.$$

Therefore, $$\varphi(z_s^\star)=\frac{1}{d}\sum_{k=1}^d\bbE[X^{(k)}_s],$$ which is the result.
\end{proof}

\newpage
\section{Sinkhorn MOT}
\label{app:mmotsk}
\subsection{Classical Sinkhorn MOT}
\cite{benamou2015} introduce an entropic regularization of Eq.~\eqref{eq:mmot}, yielding the entropic MOT problem:\begin{equation}
     \mathrm{MOT}_\varepsilon(\mathbf{a},\bs C)=\min_{\bs T \in \Pi(\mathbf{a})}\ps{\mathcal C}{\mathcal T}-\varepsilon \mathcal H(\bs{T}).\label{eq:emmot}
\end{equation} Authors show that Eq.~\eqref{eq:emmot} has a unique minimizer $\bs T^\star_\varepsilon$ that converges to a solution of Eq.~\eqref{eq:mmot} as $\varepsilon\to 0$. They also derive the multimarginal Sinkhorn algorithm, which we present in the log domain in Algorithm~\ref{alg:mmsk}, using the broadcasting function $\mathrm{broadcast}_i : \bbR^{n_i}\to \bbR^{n_1\times\cdots\times n_d}$ defined by $[\mathrm{broadcast}_i(\mathrm{f}^{(i)})]_{j_1,\ldots,j_d} = \mathrm{f}^{(i)}_{j_i}$. This work was further extended in \cite{lin2022accsk}, introducing a greedy step and acceleration.

\begin{algorithm}[H]
    \caption{Multimarginal Sinkhorn}
    \label{alg:mmsk}
    \begin{algorithmic}[1] 
        \Function{msk}{$\bs C$, $\mathbf{a}$, $\varepsilon$} 
		\For{$1\le k\le d$}
            \State $\mathrm{f}^{(k)}_0\gets \mathbf{0}\in\bbR^{n_k}$
        \EndFor
        \State $\log\bs K \gets -\bs C/\varepsilon$
        \State $\log \bs T \gets \log\bs K$
        \For{$t\ge 0$} \Comment{until convergence}
            \For{$1\le i \le d$}
            \State $\mathrm{f}^{(i)}_{t+1}\gets \mathrm{f}^{(i)}_{t}+ \log(a^{(i)})-\log(\mathrm{Marg}_i(\bs T))$ 
            \State $\log \bs T \gets \log\bs K +\sum_{i=1}^d \mathrm{broadcast}_i(\mathrm{f}^{(i)}_t)$
            \EndFor
        \EndFor
    \State \textbf{return} $\bs T = \exp(\log \bs T)$
    \EndFunction
    \end{algorithmic}
\end{algorithm}

\subsection{Greedy and accelerated Sinkhorn MOT}
In the greedy version (Algorithm~\ref{alg:greedy_mmsk}), we choose an index $K$ on which we perform the Bregman projection with
\begin{equation}
    K \gets \arg\max_{i\in\llbracket d\rrbracket} \1_{n_i}\tsp (\mathrm{Marg}_i(\bs T)-a^{(i)}) + \KL(a^{(i)} || \mathrm{Marg}_i(\bs T)). 
    \label{eq:greedy}
\end{equation}

The accelerated version (Algorithm~\ref{alg:acc_mmsk}) consider a dual entropic regularized problem: $$\min_{(\mathrm f^{(k)})_k}\varphi(\mathrm f^{(1)},\ldots, \mathrm f^{(d)})$$ where $\varphi(\mathrm f^{(1)},\ldots, \mathrm f^{(d)}) = \log(\|\bs T(\mathrm f)\|_1)-\sum_{k=1}^d (\mathrm f^{(k)})\tsp a^{(k)}$. 

\begin{algorithm}[H]
    \caption{Greedy Multimarginal Sinkhorn}
    \label{alg:greedy_mmsk}
    \begin{algorithmic}[1] 
        \Function{g-msk}{$\bs C$, $\mathbf{a}$, $\varepsilon$} 
		\For{$1\le k\le d$}
            \State $\mathrm{f}^{(k)}_0\gets \mathbf{0}\in\bbR^{n_k}$
        \EndFor
        \State $\log\bs K \gets -\bs C/\varepsilon$
        \State $\log \bs T \gets \log\bs K +\sum_{k=1}^d \mathrm{broadcast}_k(\mathrm{f}^{(k)}_0)$
        \For{$t\ge 0$} \Comment{until convergence}
            \State Choose $K$ according to \eqref{eq:greedy}
            \State $\mathrm{f}^{(K)}_{t+1}\gets \mathrm{f}^{(K)}_{t}+ \log(a^{(K)})-\log(\mathrm{Marg}_K(\bs T))$ 
            \State $\log \bs T \gets \log\bs K +\sum_{i=1}^d \mathrm{broadcast}_i(\mathrm{f}^{(i)}_t)$
        \EndFor
    \State \textbf{return} $\bs T = \exp(\log \bs T)$
    \EndFunction
    \end{algorithmic}
\end{algorithm}

\begin{algorithm}[H]
    \caption{Accelerated Multimarginal Sinkhorn}
    \label{alg:acc_mmsk}
    \begin{algorithmic}[1] 
        \Function{a-msk}{$\bs C$, $\mathbf{a}$, $\varepsilon$} 
		\For{$1\le k\le d$}
            \State $\check{\mathrm{f}}^{(k)}_0\gets \mathbf{0}\in\bbR^{n_k}$, $\tilde{\mathrm{f}}^{(k)}_0\gets \mathbf{0}\in\bbR^{n_k}$
        \EndFor
        \State $\theta_0\gets 1$, $K\gets 1$
        \State $\log\bs K \gets -\bs C/\varepsilon$\Comment{used to compute $\bs T(\mathrm{f})$}
        
        \For{$t\ge 0$} \Comment{until convergence}
            \State $\bar{\mathrm{f}}_t \gets (1-\theta_t)\check{\mathrm{f}}_t + \theta_t \tilde{\mathrm{f}}_t$
            \For{$1\le k \le d$}
             \State $\tilde{\mathrm{f}}^{(k)}_{t+1} \gets \tilde{\mathrm{f}}^{(k)}_{t} - \tfrac{1}{d\theta_t}(\mathrm{Marg}_k(\bs T(\bar{\mathrm{f}}))/\|(\bs T(\bar{\mathrm{f}})\|_1) - a^{(k)})$
             \EndFor
             \State $\grave{\mathrm{f}}_t \gets \bar{\mathrm{f}}_t + \theta_t (\tilde{ \mathrm{f}}_{t+1}-\tilde{\mathrm{f}}_t)$
             \For{$1\le k \le d$}
             \If{$k = K$}
             \State $\hat{\mathrm{f}}_t^{(k)} \gets \grave{\mathrm{f}}_t^{(k)} +\log (a^{(k)})-\log(\mathrm{Marg_k(\bs T(\grave{\mathrm{f}}_t))})$
             \Else 
              \State $\hat{\mathrm{f}}_t^{(k)} \gets \grave{\mathrm{f}}_t^{(k)}$
             \EndIf
             \EndFor
             \State $\mathrm{f}_t \gets\arg\min\{\varphi(\mathrm{f})\ |\ \mathrm{f}\in\{\check{\mathrm{f}}_t, \hat{\mathrm{f}}_t\}\}$
             \State Choose $K$ according to \eqref{eq:greedy}
             \For{$1\le k \le d$}
             \If{$k = K$}
             \State $\check{\mathrm{f}}_{t+1}^{(k)} \gets \mathrm{f}_t^{(k)} +\log (a^{(k)})-\log(\mathrm{Marg_k(\bs T(\mathrm{f}_t))})$
             \Else 
              \State $\hat{\mathrm{f}}_{t+1}^{(k)} \gets \mathrm{f}_t^{(k)}$
             \EndIf
             \EndFor
             \State $\theta_{t+1} \gets \theta_t (\sqrt{\theta_t^2+4}-\theta_t)/2$
        \EndFor
    \State \textbf{return} $\bs T(\mathrm{f}_t)$
    \EndFunction
    \end{algorithmic}
\end{algorithm}

\section{Mirror Descent for TuckerMOT}
\label{app:tucker_md}

We consider the regularized problem \begin{equation}
    \min_{\bs G,\mathbf{Q}} \ps{\bs C}{\bs G\times_1 P^{(1)}\cdots \times_d P^{(d)}}-\varepsilon \mathcal H(\bs G) \label{eq:tucker_md}
\end{equation}
with the constraints $Q^{(k)}\in\Pi(a^{(k)},g^{(k)})$ and $g^{(k)} = \mathrm{Marg}_k(\bs G)$ where $g^{(k)}$ is an independent variable and $Q^{(k)} = P^{(k)}\diag(g^{(k)})$. We perform alternating KL mirror-descent/proximal updates on the factors and an entropic Sinkhorn update on the core.

\paragraph{Step on $\mathbf Q$.} Fix an index $k$, an keep $\bs G$ and the other factors $P^{(\ell)}$ fixed. The entropy term $-\varepsilon \mathcal H(\bs G)$ is then constant. Define $$A^{(k)} = \bs C\times_{\ell\ne k}(P^{(\ell)})\tsp\times_{\ell\ne k}\bs G.$$ The cost $\ps{\bs C}{\bs T}$ is then equal to $$\ps{A^{(k)}}{P^{(k)}}=\ps{\underbrace{A^{(k)}\diag(g^{(k)})^{-1}}_{C_{Q^{(k)}}}}{Q^{(k)}}.$$ We want to minimize $$F(Q^{(k)},g^{(k)})=\ps{C_{Q^{(k)}}}{Q^{(k)}}$$ under the constraints $$g^{(k)}-\mathrm{Marg}_k(\bs G)=0,\qquad(Q^{(k)})\tsp\1_{n_k}-g^{(k)}=0.$$

Denoting $\alpha^{(k)}$ and $\beta^{(k)}$ the Lagrange multipliers associated with these constraints, the Lagrangian reads $$\mathcal L=F(Q^{(k)},g^{(k)})+\ps{\alpha^{(k)}}{g^{(k)}-\mathrm{Marg}_k(\bs G)}+\ps{\beta^{(k)}}{(Q^{(k)})\tsp\mathbf1_{n_k}-g^{(k)}}.$$ Consequently, $$\nabla_{Q^{(k)}}\mathcal L=\nabla_{Q^{(k)}}F+\mathbf1_{n_k}(\beta^{(k)})\tsp.$$

We now compute the dependence of $F$ on $g^{(k)}$. Since $C_{Q^{(k)}} = A^{(k)} / g^{(k)}$, we have $\nabla_{g^{(k)}} C_{Q^{(k)}} = - A^{(k)}/(g^{(k)})^2$. We define $$
\bar A_s^{(k)}=\sum_i\frac{Q^{(k)}_{is}}{g_s^{(k)}}A^{(k)}_{is}$$ so that $$\nabla_{g^{(k)}} F = -\frac{\bar A^{(k)}}{g^{(k)}}.$$ The first order condition with respect to $g^{(k)}$ gives $$\nabla_{g^{(k)}}\mathcal L=-\frac{\bar A^{(k)}}{g^{(k)}}+\alpha^{(k)}-\beta^{(k)}=0.$$ In particular, $\beta^{(k)}=\alpha^{(k)}-\frac{\bar A^{(k)}}{g^{(k)}}.$ 

The gradient with respect to $Q^{(k)}$ becomes
$$\nabla_{Q^{(k)}}\mathcal L=C_{Q^{(k)}}-\1_{n_k}(\bar C^{(k)})\tsp+\1_{n_k}(\alpha^{(k)})\tsp$$ with $\bar C^{(k)}=\frac{\bar A^{(k)}}{g^{(k)}}$.

We then perform a KL mirror-descent step based on this Lagrangian gradient. Given the current factor $Q^{(k)}_{t}$, define $$K^{(k)}=Q^{(k)}_{t}\odot\exp(-\eta\nabla_{Q^{(k)}}\mathcal L).$$

The marginal constraint $Q^{(k)}\mathbf1_{r_k}=a^{(k)}$ is imposed exactly through a KL projection, while the latent marginal is only semi-relaxed during this factor step. Writing $Q^{(k)}_{t+1}=\diag(u^{(k)})K^{(k)}\diag(v^{(k)})$, the row scaling is given by $$u^{(k)}=\frac{a^{(k)}}{K^{(k)}v^{(k)}}.$$ The latent scaling is updated according to the semi-relaxed KL projection with parameter $\tau$ $$v^{(k)}=\Big(\frac{g^{(k)}_{t}}{(K^{(k)})\tsp u^{(k)}}\Big)^{\frac{\tau}{\tau+\eta^{-1}}}.$$ The resulting latent marginal is then allowed to change: $$g^{(k)}_{t+1}=(Q^{(k)}_{t+1})\tsp\mathbf1_{n_k}.$$ The factors are updated for all modes using the same current feasible pair $(\mathbf Q,\bs G)$. After all factor updates have been performed, the core is recomputed using the resulting latent marginals.

\paragraph{Step on $\bs G$.} Given the updated factors $\bf P$ and latent marginals $\bf g$, we update the core by solving the entropic MOT problem with cost $$\bs C_{\bs G}=\bs C\times_{k=1}^d P^{(k)\sf T}$$ and prescribed marginals $\bf g$. This is performed using the Sinkhorn MOT algorithm.

\begin{algorithm}[H]
    \caption{Tucker MOT - Mirror Descent}
    \label{alg:tucker_md}
    \begin{algorithmic}[1]
        \Function{Tucker}{$\bs C,\mathbf a,\varepsilon,\mathbf r,\tau$}
            \State Initialize $\mathbf Q,\mathbf P,\bs G$
            \For{$t\ge 0$} \Comment{until convergence}
                \For{$1\le k\le d$}
                    \State $D^{(k)}
                    \gets C_{Q^{(k)}}-\mathbf1(\bar C^{(k)})\tsp
                    +\mathbf1(\alpha^{(k)})\tsp$
                    \State $Q^{(k)}\gets$
                    \textsc{SemiRelaxed}({$Q^{(k)}$, $D^{(k)}$, $a^{(k)}$, $\tau$})
                    \State $g^{(k)}\gets(Q^{(k)})\tsp\1_{n_k}$
                \EndFor
                \State Update $\bf P$
                \State $\bs C_{\bs G}\gets
                \mathcal C\times_{k=1}^d(P^{(k)})\tsp$
                \State $\bs G\gets$
                \textsc{Sinkhorn-MOT}($\bs C_{\bs G}$, $\mathbf g$, $\varepsilon$) \Comment{or Greedy Sinkhorn MOT}
            \EndFor
            \State \textbf{return} $\mathbf Q,\mathbf P,\bs G$
        \EndFunction
    \end{algorithmic}
\end{algorithm}

\section{Mirror Descent for CPMOT}
\label{app:md_impcp}

In this Section, we write $\xi = (\mathbf Q,g)$ and $f(\xi)$ the objective from Eq.~\eqref{eq:cplrmmot}. Following \cite{scetbon_low-rank_2021}, we define the three sets $$\mathcal C_1 (\mathbf{a},r,\alpha) = \{(\mathbf{Q},g)\in \bbR_+^{n_1r}\times\cdots\times\bbR_+^{n_dr}\times\bbR_+^{r} \ |\ \forall k,\ Q^{(k)}\1_r = a^{(k)}, g\ge \alpha\},$$ and $$\mathcal C_2 (r) = \{(\mathbf{Q},g)\in \bbR_+^{n_1r}\times\cdots\times\bbR_+^{n_dr}\times\bbR_+^{r} \ |\ \forall k,\ Q^{(k)\mathsf{T}}\1_{n_k} = g\},$$ and $\Xi = \mathcal C_1 (\mathbf{a},r,\alpha)\cap\mathcal C_2 (r)$ so that adding a regularization term on $g$ in Eq.~\eqref{eq:lrmmotQ_e} is equivalent to solving $$\inf_{\xi\in\Xi} f(\xi) -\varepsilon h(\xi),$$ for $\alpha$ sufficiently small. We perform a KL mirror descent with step size $\eta$:
\begin{align*}
     \xi_{t+1} &= \mathrm{Proj}_\Xi^{\KL}(\xi_t \odot \exp(-\eta [ \nabla f(\xi_t) + \varepsilon\log (\xi_t)]))\\
     &= \mathrm{Proj}_\Xi^{\KL}(\xi_t^{1-\eta\varepsilon}\odot \exp(-\eta \nabla f(\xi_t))
\end{align*}

In order to perform the KL projection onto $\Xi$, we use the Dykstra algorithm with projection onto $\mathcal{C}_1(\mathbf{a},r,\alpha)$ and $\mathcal C_2 (r)$.

\paragraph{Gradient of $f$.}

We write $f(\xi) = \sum_{i_1,\ldots,i_d}\bs C_{i_1,\ldots,i_d} \sum_{s=1}^r \frac{\prod_{k} Q^{(k)}_{i_ks}}{g_s^{d-1}}$.

For any $k$, $j$, $s$, we have $$\frac{\partial f}{\partial Q^{(k)}_{js}} = \sum_{\substack{i_1,\ldots,i_d\\i_k=j}} \bs C_{i_1,\ldots,i_d} \frac{\prod_{\ell \ne k} Q^{(\ell)}_{i_\ell s}}{g_s^{d-1}} = \sum_{\substack{i_1,\ldots,i_d\\i_k=j}} \bs C_{i_1,\ldots,i_d} \prod_{\ell \ne k} p^{(\ell)}_{i_{\ell}s}.$$

Using $\frac{\partial f}{\partial Q^{(k)}_{js}}=u^{(k)}_{js}$ with $u^{(k)}_{.s} = \bs C \times_{\ell\ne k} p^{(\ell)}_{s}\in\bbR^{n_k}$, we have $$\nabla_{Q^{(k)}}f = \left(u_{.1}^{(k)},\ldots,u_{.r}^{(k)}\right) \in\bbR^{n_kr}.$$

Considering $g$, $$\frac{\partial f}{\partial g_{s}} = -(d-1)  \sum_{i_1,\ldots,i_d} \bs C_{i_1,\ldots,i_d}\big(\prod_{k} Q^{(k)}_{i_ks}\big)/g_s^{d} = -(d-1)(\bs C \times_1 p^{(1)}_s\times_2\cdots \times_dp_s^{(d)}) $$

Denoting $\mathbf{c} = \begin{pmatrix}
    \bs C \times_1 p^{(1)}_1\cdots \times_dp_1^{(d)}\\ \vdots \\ \bs C \times_1 p^{(1)}_r\cdots \times_dp_r^{(d)}
\end{pmatrix}$, we have $\nabla_g f=-(d-1)\mathbf{c}\in\bbR^r$.

\paragraph{Projection onto $\mathcal C_1$.}

Each constraint in $\mathcal C_1$ is separable between the $Q^{(k)}$'s and $g$. For each $k$, we seek $$\min_{Q\in\mathcal C_1^{(k)}} \KL(Q \| Q^{(k)}) = \min_{Q\in\mathcal C_1^{(k)}} \sum_{is} Q_{is}(\log Q_{is}/ Q_{is}^{(k)}-1). $$ The Lagrangian reads $$\mathcal L (Q,\lambda)=\sum_{is} Q_{is}(\log Q_{is}/ Q_{is}^{(k)}-1) + \sum_i \lambda_{i}(\sum_{s}Q_{is}-a^{(k)}_i).$$ First order optimality conditions gives $$\nabla_Q \mathcal L = \log Q/ Q^{(k)} + \lambda\1_r\tsp = \mathbf{0}_{n_kr}\implies Q^\star =  \diag( e^{-\lambda}) Q^{(k)}$$ The $Q^\star \1_r = a^{(k)}$ condition enforces $e^{-\lambda} = a^{(k)}/Q^{(k)}\1_r$. Finally, $$\mathrm{Proj}_{\mathcal C_1}^{\KL}(Q^{(k)}) = \diag(a^{(k)}/Q^{(k)}\1_r)Q^{(k)}$$

We also have $\mathrm{Proj}_{\mathcal C_1}^{\KL}(g) = \max(g,\alpha)$, as in \cite{scetbon_low-rank_2021}.

\paragraph{Projection onto $\mathcal C_2$.} Denoting $\lambda_k\in\bbR^r$ associated with the constraint $Q^{(k)\mathsf{T}}\1_{n_k} = g$ in the Lagrangian, we obtain $$\begin{cases}
    Q^{(k)\star} = Q^{(k)}\diag(e^{-\lambda_k}),\\
    g^\star = g \odot e^{\sum_k \lambda_k}.
\end{cases}$$ 

As $Q^{(k)\star\mathsf T}\1_{n_k} = g^\star$, we have $e^{-\lambda_k} = g^\star / Q^{(k)\mathsf T}\1_{n_k}$. Thus, $(g^{\star})^{(d+1)} = g\bigodot_{k}  Q^{(k)\mathsf T}\1_{n_k}$. Finally, $$\mathrm{Proj}_{\mathcal C_2}^{\KL}(g) = \big( g\bigodot_{k=1}^d  Q^{(k)\mathsf T}\1_{n_k} \big)^{\frac{1}{d+1}},\qquad \mathrm{Proj}_{\mathcal C_2}^{\KL}(Q^{(k)}) = Q^{(k)}\diag(\mathrm{Proj}_{\mathcal C_2}^{\KL}(g) / Q^{(k)\mathsf T}\1_{n_k}).$$

We use the Dykstra algorithm to project onto $\Xi = \mathcal C_1 \cap \mathcal C_2$, as stated in Algorithm~\ref{alg:lrmmotmd}.

\begin{algorithm}[H]
    \caption{CPMOT - Mirror Descent}
    \label{alg:lrmmotmd}
    \begin{algorithmic}[1] 
        \Function{CPMOT}{$\mathcal C$, $\mathbf{a}$,$\varepsilon$,$r$,$\alpha$, $\eta_t$,$\delta$} 
             \State Initialize $(Q_0^{(k)})_k,g_0$ 
    \For{$t\ge 0$} \Comment{until convergence}
    \For{$1\le k\le d$}
    \State $K_t^{(k)} \gets (Q_t^{(k)})^{1-\eta\varepsilon}\odot \exp(-\eta \nabla_{Q^{(k)}_t}f(\xi_t))$ \Comment{$\xi_t = (Q^{(1)}_t,\ldots,Q^{(d)}_t,g_t)$}
    \EndFor
    \State $K_t^g \gets  g_t^{1-\eta\varepsilon} \odot\exp(-\eta \nabla_g(\xi_t))$
    \State $(Q_{t+1}^{(k)})_k , g_{t+1} \gets$ \textsc{Dykstra}($(K_t^{(k)},K_t^g),\alpha,\delta$) \Comment{using projections onto $\mathcal C_1$ and $\mathcal C_2$}
    \EndFor
    \State \textbf{return} $(Q_{t+1}^{(k)})_k$, $g_{t+1}$
    \EndFunction
    \end{algorithmic}
\end{algorithm}

\paragraph{Computational cost.} If no assumption is made on the cost $\bs C$, computing all $u_{.s}^{(k)}$'s has cost $\mathcal O (dr\prod_{k=1}^d n_k)$. However, as noted by \citep{Haasler2021graph, strossner2023}, the cost tensor $\bs C \in \bbR_+^{n_1\times\cdots\times n_d}$ usually carries additional structure in applications, so that we can assume that $\bs C$ has a \textit{graphical structure}. More precisely, the entries of $\bs C$ can be written $$\bs C_ {i_1,\ldots,i_d} = \sum_{\alpha\in F}\bs C^\alpha_{I_\alpha},$$ for every index tuple $I = (i_1,\ldots, i_d)$, where $$F\subseteq \bigcup_{K=1}^d\{(\alpha_1,\ldots,\alpha_K)\in\bbN^K\ |\ 1\le \alpha_1<\cdots <\alpha_K\le d\}$$ is a fixed set of index tuples, $I_\alpha \defeq (i_{\alpha_1},\ldots,i_{\alpha_K})$ and $C^\alpha\in\bbR_+^{n_{\alpha_1}\times\cdots\times n_{\alpha_K}}$. The case of pairwise CNT costs used in Section \ref{seq:cost_struct} $$\bs C_ {i_1,\ldots,i_d}=\sum_{k<l}\| \varphi(x_{i_k}^{(k)}) -\varphi(x_{i_l}^{(l)})\|_{\mathcal H}^2,$$ has a graphical structure. Indeed, let $F = \{(k,l)\in\bbN^2\ |\ 1\le k < l \le d\}$. Then for each $\alpha = (k,l)\in F$, the associated tensor $\bs C^\alpha$ is defined by $\bs C^{(k,l)}_{i_k,i_l} = \| \varphi(x_{i_k}^{(k)}) -\varphi(x_{i_l}^{(l)})\|_{\mathcal H}^2$.

Using the structure of the cost allows us to reduce the computational cost of computing all $u_{.s}^{(k)}$'s. Indeed, we can rewrite $\frac{\partial f}{\partial Q^{(k)}_{js}}$ as \begin{align*}
    \frac{\partial f}{\partial Q^{(k)}_{js}}&= \sum_{\substack{i_j, j\ne k}} \bs C_{i_1,\ldots,j,\ldots,i_d} \prod_{\ell\ne k} p^{(\ell)}_{i_\ell s}\\
    &= \sum_{\substack{\alpha\in F\\ k\in \alpha}} \big[\underbrace{\sum_{I_{\alpha\setminus\{k\}}} \bs C^{\alpha}_{I_{\alpha\setminus\{k\}}, j}\prod_{\ell \in\alpha\setminus\{k\}}p^{(\ell)}_{i_\ell s}}_{v^{(k,\alpha)}_{js}}\big] + \sum_{\substack{\alpha\in F\\ k\notin \alpha}}\big[\underbrace{\sum_{I_\alpha} \bs C_{I_\alpha}^\alpha\prod_{\ell \in\alpha}p^{(\ell)}_{i_\ell s}}_{g^\alpha_s}\big]
\end{align*}

One can compute all $g^\alpha_s$'s once in $\mathcal O(r\sum_{\alpha \in F} \prod_{\ell\in\alpha} n_\ell)$ to be used for every $k\notin \alpha$, and the $v^{(k,\alpha)}_{js}$'s in $\mathcal O(r\sum_{\alpha \in F} \sum_{k\in \alpha} \prod_{\ell\in\alpha} n_\ell)$ when $k\in\alpha$. The total complexity to compute each $\nabla_{Q^{(k)}}f$ is thus $\mathcal O (r\sum_{\alpha\in F} (|\alpha|+1)\prod_{\ell\in\alpha}n_\ell)$. Using the graphical structure of the cost tensor reduces the previous $\mathcal O (drn^d)$ to $\mathcal O (drn^{\max_{\alpha\in F} |\alpha|})$.

Note that the $s$-th line of $\mathbf c$ is $u^{(k)}_{.s}\times_k p_{s}^{(k)}=\sum_j p_{js}^{(k)}u_{js}^{(k)}$ for an arbitrary $k$, so that computing $\nabla_g f$ does not change the overall complexity.

In particular, for the class of pairwise costs we consider, we can further use the squared distance structure: $u_{js}^{(k)}  = (d-1)\|x_{i_k}^{(k)}\|^2_{\mathcal H}+\sum_{\ell\ne k} \sum_{i_\ell} p^{(\ell)}_{i_\ell}\|x^{(\ell)}_{i_\ell}\|^2_{\mathcal H} -2\ps{\varphi(x_{i_k}^{(k)})}{\sum_{\ell\ne k} \sum_{i_\ell} p^{(\ell)}_{i_\ell}\varphi(x^{(\ell)}_{i_\ell})}_{\mathcal H} $ reducing the overall cost to $\mathcal O (TDNr)$, where $D$ is the average number of Dykstra iterations, and $T$ the number of mirror descent steps, and $N=\sum_{k=1}^d n_k$.

\paragraph{Convergence.} We follow the convergence criterion from \citep{scetbon_low-rank_2021}. Define $F_\varepsilon(\xi) = f(\xi)-\varepsilon h(\xi)$ the function we aim to minimize. For any $\eta > 0$ and $\xi\in\Xi$, define $$G_\varepsilon(\xi,\eta) = \arg\min_{\zeta\in\Xi}\, \ps{\nabla F_\varepsilon(\xi)}{\zeta} + \frac 1\eta \KL (\zeta,\xi),$$ and $$\Delta_\varepsilon(\xi,\eta) = \frac{1}{\eta^2} \big[\KL(\xi, G_\varepsilon(\xi,\eta)) + \KL(G_\varepsilon(\xi,\eta),\xi)\big].$$ Assume $\bs C$ has finite entries, and that the initial point $\xi_0\in \mathrm{ri}(\Xi)$. Then, for any number of iteration $T\ge 1$, taking the constant step size $\eta = 1/2 L_\varepsilon$, we have $$\min_{0\le t\le T-1} \Delta_\varepsilon(\xi_t,\eta)\le \frac{4L_\varepsilon D_0}{T},$$ where $D_0 = F_\varepsilon(\xi_0)-\min_{\xi\in\Xi} F_\varepsilon(\xi)$ and $L_\varepsilon$ is defined in Eq.~\eqref{eq:Leps}. 

The following proof assumes exact KL and Dykstra projections. In practice, we have tolerance thresholds.

\begin{proof}
    Whenever $\alpha\le 1/r$, $\Xi$ is non empty. Since $\Xi$ is defined by closed constraints, it is a closed and bounded set in a finite dimensional space, and is therefore compact.

    A sufficient condition for the relative smoothness of $f$ with respect to $\varphi = -h$ is $\nabla^2 f (\xi) \preceq L_f \nabla^2 \varphi(\xi)$. Now, we have that $\nabla^2 \varphi(\xi) = \diag(\tfrac{1}{\xi_1},\ldots,\tfrac{1}{\xi_m})$, and $0\le Q_{is}^{(k)}\le 1$, $\alpha\le g_s\le 1$ so that $\nabla^2 \varphi(\xi) \succeq I$. Thus, in $\mathrm{ri}(\Xi)$, if $\sup_{\xi \in \Xi} \| \nabla^2f(\xi)\|_{\mathrm{op}} \le M$, then $\nabla^2f(\xi)\preceq MI \preceq M\nabla^2\varphi(\xi)$. We can extend the resulting inequality to $\Xi$ by continuity.

    Thus, let us check second derivatives of $f$. For $k\ne \ell$, we have $$\frac{\partial^2 f}{\partial Q^{(k)}_{js}Q^{(\ell)}_{is}} = \frac{1}{g_s^{d-1}} \sum_{\substack{i_1,\ldots,i_d\\ i_k=j, i_\ell =i}} \bs C_{i_1\ldots i_d} \prod_{q\ne k,\ell} Q^{(q)}_{i_qs},$$ so $\big|\frac{\partial^2 f}{\partial Q^{(k)}_{js}Q^{(\ell)}_{is}}\big| \le \frac{\|\bs C\|_\infty}{\alpha^{d-1}}\prod_{q=1}^d n_q$. For any $k$, we have $$\frac{\partial^2 f}{\partial Q^{(k)}_{js}\partial g_s} = -\frac{d-1}{g_s^d}\sum_{\substack{i_1,\ldots,i_d\\ i_k=j}} \bs C_{i_1\ldots i_d} \prod_{q\ne k} Q^{(q)}_{i_qs},$$ so $\big|\frac{\partial^2 f}{\partial Q^{(k)}_{js}\partial g_s}\big|\le \frac{(d-1)\|\bs C\|_\infty}{\alpha^d}\prod_{q=1}^d n_q$. Finally, $$\frac{\partial^2 f}{\partial g_s^2} = d(d-1) \frac{1}{g_s^{d+1}}\sum_{\substack{i_1,\ldots,i_d}} \bs C_{i_1\ldots i_d} \prod_{q} Q^{(q)}_{i_qs}, $$ so $\big|\frac{\partial^2 f}{\partial g_s^2}\big| \le \frac{d(d-1)\|\bs C\|_\infty}{\alpha^{d+1}}\prod_{q=1}^d n_q$. The other derivatives vanish.

    Since the Hessian is of size $m\times m$, we use $\|A\|_\mathrm{op} \le m\max_{ij}|A_{ij}|$ to get $$\| \nabla ^2 f(\xi)\|_{\mathrm{op}} \le m \|\bs C\|_\infty\Big(\prod_{q=1}^d n_q\Big)\max \Big(\frac{1}{\alpha^{d-1}}, \frac{d-1}{\alpha^d}, \frac{d(d-1)}{\alpha^{d+1}}\Big) = L_f.$$

    We have $F_\varepsilon = f+\varepsilon \varphi$. Since $\varphi$ is $1$-smooth relatively to itself, we can take \begin{equation}
        L_\varepsilon = L_f + \varepsilon, \label{eq:Leps}
    \end{equation} so that $F_\varepsilon$ is $L_\varepsilon$-smooth relatively to $\varphi$.

    We can now establish the convergence criterion. By relative smoothness, for any $t$, we have $$F_\varepsilon(\xi_{t+1}) \le F_\varepsilon(\xi_t)+\ps{\nabla F_\varepsilon(\xi_t)}{\xi_{t+1}-\xi_t}+L_\varepsilon\KL(\xi_{t+1}||\xi_t).$$ By optimality of the mirror descent update, for any $\xi\in \Xi$, we have $$\ps{\nabla F_\varepsilon(\xi_{t})+\frac{1}{\eta}(\nabla \varphi(\xi_{t+1})-\nabla \varphi(\xi_t))}{\xi-\xi_{t+1}}\ge 0.$$ Taking $\xi = \xi_t$ in this inequality gives $$\ps{\nabla F_\varepsilon(\xi_{t})}{\xi_{t+1}-\xi_t}\le -\frac{1}{\eta} (\KL(\xi_t||\xi_{t+1})+\KL(\xi_{t+1}||\xi_t)).$$ Thus, $$F_\varepsilon(\xi_{t+1}) \le F_\varepsilon(\xi_{t})-(\frac{1}{\eta}-L_\varepsilon)\KL(\xi_{t+1}||\xi_t) - \frac 1\eta \KL(\xi_t || \xi_{t+1}).$$ With $\eta = \frac{1}{2L_\varepsilon}$, $$F_\varepsilon(\xi_{t+1}) \le F_\varepsilon(\xi_{t})-\frac{1}{2\eta}(\KL(\xi_{t+1}||\xi_t) + \KL(\xi_t || \xi_{t+1})).$$ Summing over $t\in\{0,\ldots, T-1\}$, and using $F_\varepsilon(\xi_t) \ge F_\varepsilon^\star$, yields $$\frac{1}{2\eta} \sum_{t=0}^{T-1} (\KL(\xi_{t+1}||\xi_t) + \KL(\xi_t || \xi_{t+1})) \le F_\varepsilon(\xi_0) - F_\varepsilon^\star.$$ Finally, $$\min_{0\le t\le T-1} \Delta_\varepsilon (\xi_t,\eta) \le \frac{2( F_\varepsilon(\xi_0) - F_\varepsilon^\star)}{T\eta} = \frac{4L_\varepsilon(F_\varepsilon(\xi_0) - F_\varepsilon^\star)}{T}.$$

\end{proof}

\section{Alternating minimization for eCPMOT}

\paragraph{Initialization} As the optimal $\varphi(z^\star_s)$ is an arithmetic mean of the source hubs $\mu^{(k)}_s$, we can initialize the hub positions $\varphi(z)$ using the k-means++ initialization method \citep{Sergei2007kmeans} on the sets $\{\varphi(x^{(k)}_{i_k})\}_{i_k=1}^{n_k}$ for each $k$, and then take the mean of the resulting sets of centers. To compute this mean, we need to assign a point from each set to each index $s$. At the cost of breaking the symmetry, we use the following procedure, which requires $d-1$ calls to \textsc{LinearSumAssignment}. For each $k>1$, we get $\tau_k$ by solving \textsc{LinearSumAssignment}$([c(\tilde x_s^{(1)},\tilde x_{s'}^{(k)})]_{s,s'})$ and set $$\varphi(z_s) = \frac 1d \big(\varphi(\tilde x^{(1)}_s + \sum_{k=2}^d \varphi(\tilde x_{\tau_k(s)}^{(k)})\big).$$ 

\paragraph{Initialization sensitivity.} To assess the effect of the initialization of $\varphi(z)$ on the results, Figure~\ref{fig:initialization_cpmmot} show the distribution of the resulting transport cost for random and k-means-z initialization across 500 random seeds on the same problem. We consider $4$ measures, 30 points each sampled from a standard $2$-D Gaussian distribution, and take $r=7$. We observe that k-means-z- initialization substantially reduces the overall runtimes while finding the best local minimum more frequently.

\def\histwidth{0.000136}
\begin{figure}[H]
    \centering
\begin{tikzpicture}
\begin{axis}[
    width=7cm,
    height=4cm,
    xlabel={Transport cost},
    ylabel={Density},
    ylabel style={yshift=-0.5cm},
    ytick=\empty,
    xmin=0.0557,
    xmax=0.0585,
    ymin=0,xtick={0.056,0.057,0.058},
    grid style={
        gray!20
    },
    legend style={
        at={(0.98,0.98)},
        anchor=north east,
        draw=none,
        fill=white,
        fill opacity=0.9,
        text opacity=1
    },
    axis on top=false,
]

\addplot[
    ybar interval,
    fill=blue!60,
    draw=blue!80!black,
    fill opacity=0.25,
    draw opacity=0.5,
]
table[
    x expr=\thisrow{x}-\histwidth/2,
    y=density,
    col sep=space
]{init_mmot/hist_distribution1.dat};

\addplot[
    ybar interval,
    fill=red!60,
    draw=red!80!black,
    fill opacity=0.25,
    draw opacity=0.5,
]
table[
    x expr=\thisrow{x}-\histwidth/2,
    y=density,
    col sep=space
]{init_mmot/hist_distribution2.dat};

\addplot[blue!80!black, dashed] table[row sep = crcr]{0.056519 0 \\ 0.056519 3200 \\};
\addlegendentry{kmeans}
\addplot[red!80!black, dashed] table[row sep = crcr]{0.056776 0 \\ 0.056776 3200 \\};
\addlegendentry{random}
\end{axis}
\end{tikzpicture}
\caption{Distribution of the transport-cost (lower is better) across random seeds for different initialization methods. Average runtimes are 0.37\,s and 0.47\,s for k-means-$z$ and random initialization, respectively, while the corresponding average values of the transport cost are $0.0565$ and $0.0568$.}
    \label{fig:initialization_cpmmot}
\end{figure}

\begin{algorithm}[H]
\caption{k-means $z$ initializer for $d$ sets}
\label{alg:kmeans-general}
\begin{algorithmic}[1]
\Function{d-kmeans-z}{$(x^{(k)})_k$, $r$}
    \For{$1\le k\le d$}
        \State $\varphi(\tilde x^{(k)})
        \gets \textsc{kmeans++}
        ((\varphi(x_{i_k}^{(k)} ))_{i_k},r)$
    \EndFor
    \For{$2\le k\le d$}
        \State $C^{(k)}
        \gets [\|\varphi(\tilde x^{(1)}_s)-\varphi(\tilde x^{(k)}_{s'})\|_{\mathcal H}^2]_{s,s'}$
        \State $\tau_k
        \gets \textsc{LinearSumAssignment}(C^{(k)})$
    \EndFor
        \State $\varphi(z)
        \gets \frac{1}{d}
        \left[
        \varphi(\tilde x^{(1)}_s)
        +\sum_{k=2}^d
        \varphi(\tilde x^{(k)}_{\tau_k(s)})
        \right]_{s}$
    \State \textbf{return} $\varphi(z)$
\EndFunction
\end{algorithmic}
\end{algorithm}

\begin{algorithm}[H]
\caption{Sinkhorn barycenter couplings \cite{benamou2015}}
\label{alg:sk_mmbary}
\begin{algorithmic}[1]
\Function{SinkhornBary}{$(C^{(k)})_k$, $\mathbf{a}$, $\varepsilon$, $r$}
    \State $d \gets \mathrm{Card}(\{C^{(k)}\}_k)$
    \For{$1\le k\le d$}
        \State $K^{(k)} \gets \exp\left(-C^{(k)}/\varepsilon\right)$
        \State $v^{(k)} \gets \mathbf{1}_r$
    \EndFor
    \State $g \gets \frac{1}{r}\mathbf{1}_r$
    \For{$t\ge 0$}
        \For{$1\le k\le d$}
            \State $u^{(k)} \gets \dfrac{a^{(k)}}{K^{(k)}v^{(k)}}$
        \EndFor
        \State $g \gets  \exp\big(\frac{1}{d}\sum_{k=1}^{d} \log(K^{(k)\mathsf T}u^{(k)})\big)$
        \State $g \gets g/\mathbf{1}_r^\mathsf T g$
        \For{$1\le k\le d$}
            \State $v^{(k)} \gets \dfrac{g}{K^{(k)\mathsf T}u^{(k)}}$
        \EndFor
    \EndFor
    \For{$k\ge 0$}
        \State $Q^{(k)} \gets \diag(u^{(k)})K^{(k)}\diag(v^{(k)})$
    \EndFor
    \State \textbf{return} $(\mathbf{Q},g)$
\EndFunction
\end{algorithmic}
\end{algorithm}

\begin{algorithm}[h]
    \caption{Explicit CPMOT}
    %\label{alg:blrmmot}
    \begin{algorithmic}[1] 
        \Function{ECPMOT}{$r$, $(x^{(k)})_k$, $\mathbf{a}$, $\varepsilon$} 
             \State $\varphi(z)\gets$ \textsc{d-kmeans-z}($(\varphi(x^{(k)}))_k,r$)
            \For{$1\le k\le d$}
            \State $C^{(k)} \gets [\|\varphi(x^{(k)}_{i_k})-\varphi(z_s)\|^2_{\mathcal H}]_{i_ks}$
            \EndFor
    \For{$t\ge 0$} \Comment{until convergence}
    
    \State $\mathbf{Q},g\gets$ \textsc{SinkhornBary}($(C^{(k)})_k,\mathbf{a},\varepsilon/d$)
    \State $\varphi(z) \gets [\frac{1}{dg_s}\sum_{k=1}^d Q_{i_ks}^{(k)} \varphi(x_{i_ks}^{(k)})]_{s=1}^r$
    \For{$1\le k\le d$}
            \State $C^{(k)} \gets [\|\varphi(x^{(k)}_{i_k})-\varphi(z_s)\|^2_{\mathcal H}]_{i_ks}$
    \EndFor
    \EndFor
    \State \textbf{return} $\mathbf{Q},g$
    \EndFunction
    \end{algorithmic}
\end{algorithm}

\subsection{Scheme convergence}

As stated in \citep{peyre_optimal_2026} (Section 12.1), the regularization gives strict convexity for the objective in $(\mathbf Q,g)$, but it remains non-convex in $\varphi(z)$. Consequently, the joint problem is non-convex. The sequence of values produced by Algorithm~\ref{alg:blrmmot} is non-increasing because the $(\mathbf Q,g)$ step, as well as the $\varphi(z)$ step, diminish the objective. By definition, the objective is below-bounded, and thus Algorithm~\ref{alg:blrmmot} converges to a local minimum.

\paragraph{Computational cost.} Denote $N=\sum_{k=1}^d n_k$. Each iteration of the inner \textsc{SinkhornBary} loop has a computational cost of $\mathcal O (Nr)$. Initialization, construction of the cost matrices $C^{(k)}$ and updating $\varphi (z)$ cost $\mathcal O (Nr)$ as well. Therefore,  denoting $T$ the average number of iterations of \textsc{SinkhornBary}, and $S$ the number of iterations of the outer loop, the total computational cost is $\mathcal O(TSNr)$. By comparison, the Sinkhorn algorithm~\ref{alg:mmsk} has a computational cost of $\mathcal O(Td\prod_{k=1}^d n_k)$, dominated by the computation of $\mathrm{Marg}_k(\bs T)$. 

\subsection{Explicit Unbalanced CP Multi-Marginal Optimal Transport}
\label{seq:ummot_expl}
In the unbalanced setting \citep{Beier2022ummot}, we allow the transport plan $\bs T$ to not have its $k$-th marginal equal to $a^{(k)}$, but we add a penalization term to the objective to enforce $\mathrm{Marg}_k(\bs T)\approx a^{(k)}$. Denoting the set of tensors of nonnegative CP-rank at most $r$ having positive entries $J_r = \{\bs T \in\bbR_+^{n_1\times \cdots\times n_d}\ |\ \mathrm{rank}_{\text{CP}+}(\bs{T})\le r \}$, the problem reads \begin{equation}
    \mathrm{UCPMOT}_r(\mathbf{a}) = \inf_{\bs T \in J_r} \ps{\bs C}{\bs T} + \tau\sum_{k=1}^d \KL(\mathrm{Marg}_k(\bs T)\|a^{(k)}).\label{eq:ulrmmot}
\end{equation} Using the low-rank factorization of $\bs T$, with $Q^{(k)}\in\bbR^{n_kr}_+$ satisfying $Q^{(k)\mathsf{T}}\1_{n_k} =g$, the problem \eqref{eq:ulrmmot} is equivalent to minimizing \begin{equation*}
    \mathrm{UCPMOT}_r(\mathbf{a}) = \inf_{\mathbf{Q},g} \ps{\mathcal C}{\sum_{s=1}^rg_s \bigcirc_{k=1}^d p^{(k)}_s} + \tau\sum_{k=1}^d \KL(Q^{(k)}\1_r\|a^{(k)}).
\end{equation*} The proof of Theorem~\ref{thm:expl_upperbound} relying on the $Q^{(k)}$'s having $g$ as right marginal, it applies in the unbalanced case. Adding entropic regularization, the upper bound is \begin{align}
    \inf_{\mathbf{Q},g} \ps{\mathcal C}{\sum_{s=1}^rg_s \bigcirc_{k=1}^d p^{(k)}_s} + &\tau\sum_{k=1}^d \KL(Q^{(k)}\1_r\|a^{(k)}) + \varepsilon h(\mathbf{Q}) \le\nonumber \\ & \inf_{\mathbf Q, g,\varphi(z)} d\Big(\sum_{k=1}^d \ps{C^{(k)}}{Q^{(k)}}-\frac{\varepsilon}{d}\mathcal H(Q^{(k)}) + \tau\KL(Q^{(k)}\1_r\|a^{(k)})\Big). \label{eq:ulrmmot_expl}
\end{align}

In order to solve Eq.~\eqref{eq:ulrmmot_expl}, we can apply Algorithm~\ref{alg:blrmmot} replacing \textsc{SinkhornBary} in line 5 by its unbalanced version. Theorem~\ref{thm:bound_mmot} also holds in the case.

\section{Classical bipartite low-rank optimal transport}
\label{seq:lot}

In the bipartite classical case, the optimal transport setup involves two probability measures $a = \sum_{i=1}^n a_i\delta_{x_i}$ and $b= \sum_{j=1}^m b_j\delta_{y_j}\in\mathcal P(\mathcal X)$, and a cost matrix $C\in\bbR^{nm}$, such that $C_{ij} = c(x_i,y_j)$ where $c : \mathcal X \times \mathcal X \to \bbR$ is a cost function over pairs of points in $\mathcal X$. Solving optimal transport is finding a transport plan $T\in\bbR_+^{nm}$, where $T_{ij}$ represents the amout of mass going from $x_i$ to $y_j$, such that $T$ satisfies marginal constraints, i.e. $$T\in \Pi(a,b) = \{T\in\bbR^{nm}_+\ |\ T\1_m =a \text{ and } T\tsp\1_n = b \},$$ and such that $T$ minimizes the total transport cost. The problem reads $$\mathrm{OT}(a,b,C)=\min_{T\in \Pi(a,b)} \ps{C}{T}.$$

The low-rank optimal transport (LOT) problem is defined as \begin{equation}\mathrm{LOT}_r(a,b,C)=\min_{T\in \Pi_r(a,b)} \ps{C}{T}\label{eq:lot}\end{equation} where $\Pi_r(a,b) = \{T\in \Pi(a,b)\ |\ \mathrm{rank}_+(T) \le r\}$. \cite{scetbon_low-rank_2021} introduce a factorization of the low-rank coupling $T$ as a triple $(Q,R,g)$ such that $g\in \Delta_r^*$, $Q \in \Pi(a,g)$, $R\in\Pi(b,g)$, and $$T = Q \diag(1/g)R\tsp,\qquad T_{ij} = \sum_{s=1}^r \frac{Q_{is}R_{js}}{g_s}.$$ The vector $g$ represents weights of an implicit intermediate measure $\sum_{s=1}^r g_s\delta_{z_s}$ where the locations of the hubs $z_s$ do not take part in the formulation of the problem. The transport from $a$ to $b$ can be seen as the transport from $a$ to $g$ and from $g$ to $b$, with transport plans $Q$ and $R\tsp$.

Thus, \cite{scetbon_low-rank_2021} show that solving Eq.~\eqref{eq:lot} amounts to solving \begin{equation*}
\min_{Q,R,g} \ps{C}{Q\diag(1/g) R\tsp}%\label{eq:lot_scetbon}
\end{equation*} where $(Q,R,g)\in \bbR_+^{nr}\times \bbR_+^{mr}\times (\bbR_+^*)^{r}$ such that $Q\1_r = a$, $R\1_r = b$ and $Q\tsp\1_n = R\tsp\1_m = g$.  

This problem can be regularized by solving \begin{equation}
\min_{Q,R,g} \ps{C}{Q\diag(1/g) R\tsp}-\varepsilon h(Q,R,g)\label{eq:lot_scetbon_reg}
\end{equation} where $h(Q,R,g) = \mathcal H(Q)+\mathcal H(R)+\mathcal H(g)$. Eq.~\eqref{eq:lot_scetbon_reg} is solved using a Mirror Descent scheme. 

\paragraph{Riemannian formulation.} \cite{jawanpuria_riemannian_2026} show that the space of low-rank factorizations $\mathcal M = \{(Q,R) \in (\bbR_+^*)^{nr}\times(\bbR_+^*)^{mr}\ |\ Q\1_r=a, R\1_r=b, Q\tsp\1_n=R\tsp\1_m \}$ is a non-empty embedded smooth manifold of $(\bbR_+^*)^{nr}\times(\bbR_+^*)^{mr}$ with dimension $(n+m-1)(r-1)$. Authors derive Riemannian ingredients to use second-order Riemannian algorithms such as Riemannian conjugate gradients and Riemannian trust regions \citep{boumal2023intromanifolds}. This method does not require a regularization parameter $\varepsilon$.

\paragraph{Approximation Error.} \cite{Scetbon2022stats} provide an upper bound on the error made by restricting the transport plan to have rank $r$. We have for any $2\le r\le \min(n,m)$: $$|\mathrm{LOT}_r(a,b,C) - \mathrm{OT}(a,b,C)| \le \|C\|_\infty\ln(\min(n,m)/(r-1)).$$

\paragraph{LOT as clustering.} \cite{Scetbon2022stats} state that LOT can be seen as a generalization of the $k$-means methods. Indeed, if we take $a=b$, then $$\mathrm{LOT}_r(a,a,C)=\min_Q \ps{C}{Q\diag(1/Q\tsp\1_n)Q\tsp}$$ where $Q\1_r = a$. We recover the $k$-means algorithm by taking $r=k$ and $c$ the squared Euclidean distance. A co-clustering algorithm using this formulation is derived in \cite{schmidt2026transport}: authors also propose an approximation of the low-rank optimal transport by a generalized $k$-means and provide the transport clustering (\tc) algorithm, which is specific to the $n=m$ case, and require solving a full-rank optimal transport problem as an initialization.

\paragraph{Computational cost.} Algorithms considered in Appendix~\ref{app:exp_d2} have computational cost \expcp: $\mathcal O(NT(n+m)r)$, \impcp: $\mathcal O (NT(n+m)r)$, \tucker: $\mathcal O(NLr^2(n+m))$, \rcg: $\mathcal O (N(k_{\text{CG}}+L)r(n+m))$, \latot: $\mathcal O(N(8r^3+T(n+m)r))$ where $N$ is the number of iterations of the outer loop, and other constants are related to the number of iterations in inner loops. We refer to related articles for further details.

%% file: appendix_d2_exp.tex
\section{Experiments}
\label{app:exp}
\subsection{More experiments and details}
\label{app:exp_dd}
\paragraph{Time gain and optimality.}

The four densities considered in Table~\ref{tab:mixtures} are $$f_1 = \frac 13 \mathcal N([0,0],\Sigma)+\frac 13 \mathcal N([0, 1],\Sigma)+\frac 13 \mathcal N([1,1],\Sigma)$$
$$f_2 = \frac 12 \mathcal N([0.5,0.5],\Sigma) +\frac 12 \mathcal N([-0.5,0.5],\Sigma)$$
$$f_3 = \frac 14 \mathcal N([0.5,0.5],\Sigma)+\frac 14 \mathcal N([-0.5,0.5],\Sigma)+\frac 14 \mathcal N([0,0.5],\Sigma)+\frac 14 \mathcal N([-0.5,0],\Sigma)$$
$$f_4 = \frac 12 \mathcal N([0.5,-0.5],\Sigma) +\frac 12 \mathcal N([-0.5,-0.5],\Sigma)$$ where $\Sigma = 0.05\times I_2$.

We stopped each method after 3000 outer iterations or after 60\,s in Table~\ref{tab:mmot_time_opt}. 

\begin{table}[H]
\caption{More values of $r$ in the same setting as Table~\ref{tab:mmot_time_opt}.}
\label{tab:more_mmot_time_opt}
\begin{center}
\begin{tabular}{c l c c c c}
\hline
$r$ & Method & Transport cost & Marg. err & Iterations & Time (s)\\
\hline
&\textsc{msk} & $0.186 \pm 0.022$ & $10^{-8}$ & $246 \pm 62$ & $20.3 \pm 6.3$ \\
&\textsc{g-msk} & $0.186 \pm 0.022$ & $10^{-7}$ & $580 \pm 122$ & $8.2 \pm 1.9$ \\
\hline
\multirow{4}{*}{$2$}
&\expcp & $\gr{0.229} \pm \gr{0.022}$ & $10^{-8}$ & $7 \pm 3$ & $\gr{0.2} \pm \gr{0.1}$ \\
&\impcp & $0.258 \pm 0.028$ & $10^{-8}$ & $2007 \pm 690$ & $2.8 \pm 1.9$ \\
&\tucker\ \textsc{g-msk}  & $0.232 \pm 0.022$ & $10^{-17}$ & $532 \pm 415$ & $60.2 \pm 0.1$ \\
&\tucker\ \textsc{msk} & $0.233 \pm 0.022$ & $10^{-17}$ & $492 \pm 377$ & $60.2 \pm 0.3$ \\ 
\hline
\multirow{4}{*}{$4$}
&\expcp & $ \gr{0.202} \pm \gr{0.022}$ & $10^{-8}$ & $20 \pm 9$ & $\gr{0.4} \pm \gr{0.2}$ \\
&\impcp & $0.231 \pm 0.028$ & $10^{-8}$ & $2379 \pm 568$ & $4.4 \pm 2.2$ \\
&\tucker\ \textsc{g-msk} &  $0.231 \pm 0.022$ & $10^{-17}$ & $1434 \pm 786$ & $59.6 \pm 1.8$ \\ 
&\tucker\ \textsc{msk} & $0.231 \pm 0.022$ & $10^{-17}$ & $1677 \pm 970$ & $56.9 \pm 6.6$ \\ \hline
\multirow{4}{*}{$7$}
&\expcp & $\gr{0.193} \pm \gr{0.023}$ & $10^{-8}$ & $35 \pm 18$ & $\gr{0.6} \pm \gr{0.4}$ \\
&\impcp & $0.222 \pm 0.025$ & $10^{-8}$ & $2499 \pm 439$ & $5.1 \pm 1.7$ \\
&\tucker\ \textsc{g-msk}  & $0.232 \pm 0.023$ & $10^{-17}$ & $1127 \pm 370$ & $60.2 \pm 0.3$ \\
&\tucker\ \textsc{msk} & $0.231 \pm 0.022$ & $10^{-17}$ & $1505 \pm 680$ & $59.8 \pm 1.1$ \\
 \hline
\multirow{4}{*}{$10$}
&\expcp &$\gr{0.191} \pm \gr{0.022}$ & $10^{-8}$ & $56 \pm 36$ & $\gr{0.7} \pm \gr{0.4}$ \\
&\impcp & $0.219 \pm 0.023$ & $10^{-8}$ & $2511 \pm 432$ & $5.1 \pm 1.7$ \\
&\tucker\ \textsc{g-msk}  &  $0.235 \pm 0.022$ & $10^{-17}$ & $759 \pm 219$ & $60.3 \pm 0.4$ \\
&\tucker\ \textsc{msk} &  $0.234 \pm 0.022$ & $10^{-17}$ & $875 \pm 295$ & $60.1 \pm 0.1$ \\\hline
\multirow{4}{*}{$15$}
&\expcp & $\gr{0.189} \pm \gr{0.022}$ & $10^{-8}$ & $69 \pm 34$ & $\gr{0.9} \pm \gr{0.5}$ \\
&\impcp & $0.218 \pm 0.025$ & $10^{-8}$ & $2577 \pm 438$ & $5.4 \pm 1.8$ \\
&\tucker\ \textsc{g-msk} & $0.268 \pm 0.031$ & $10^{-17}$ & $278 \pm 74$ & $62.1 \pm 2.0$ \\
&\tucker\ \textsc{msk} & $0.248 \pm 0.022$ & $10^{-17}$ & $389 \pm 110$ & $60.2 \pm 0.1$ \\ \hline
\multirow{4}{*}{$20$}
&\expcp  & $\gr{0.189} \pm \gr{0.022}$ & $10^{-8}$ & $76 \pm 33$ & $\gr{0.8} \pm \gr{0.4}$ \\
&\impcp & $0.217 \pm 0.025$ & $10^{-8}$ & $2508 \pm 452$ & $5.5 \pm 1.9$ \\
&\tucker\ \textsc{g-msk} & $0.295 \pm 0.028$ & $10^{-17}$ & $189 \pm 51$ & $64.3 \pm 4.7$ \\
&\tucker\ \textsc{msk} & $0.282 \pm 0.026$ & $10^{-17}$ & $253 \pm 55$ & $60.3 \pm 0.2$ \\
\end{tabular}
\end{center}
\end{table}

\begin{figure}[H]
    \centering
    \begin{subfigure}[b]{0.32\linewidth}
        \centering
        \includegraphics[width=\linewidth]{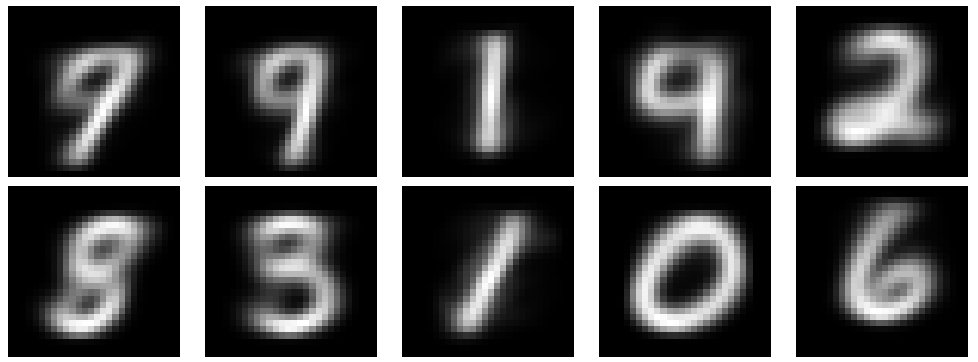}
        \caption{\expcp\ $(d)$}
        \label{fig:mnist_z_ecpd}
    \end{subfigure}
    \hfill
    \begin{subfigure}[b]{0.32\linewidth}
        \centering
        \includegraphics[width=\linewidth]{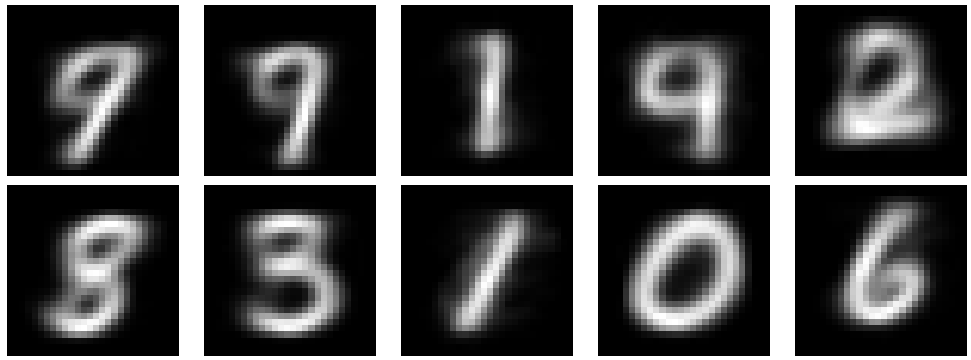}
        \caption{\expcp\ $(2)$}
        \label{fig:mnist_z_ecp2}
    \end{subfigure}
    \hfill
    \begin{subfigure}[b]{0.32\linewidth}
        \centering
        \includegraphics[width=\linewidth]{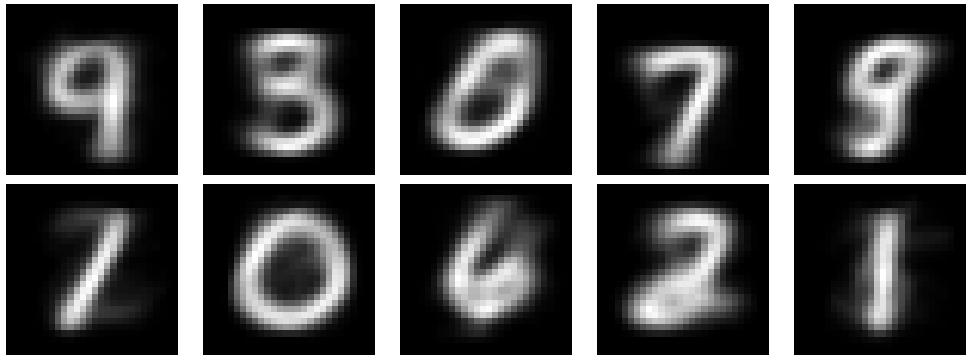}
        \caption{\rcg}
        \label{fig:mnist_z_rcg}
    \end{subfigure}
    
    \caption{Barycenters $z_s\in\bbR^{784}$ displayed as $28\times 28$ images for the three methods.}
    \label{fig:mnist_z}
\end{figure}

\paragraph{Matching MNIST.} We create $d=20$ measures $\mathbf a = (a^{(k)})_{k=1}^{20}$, each supported on $10\times 150$ points which are vectors representing different digits in the MNIST \citep{mnist} dataset. We take the pairwise cost with the squared euclidean distance as cost tensor $\bs C$, and solve $\mathrm{eCPMOT}_{r,\varepsilon}(\mathbf a,\bs C)$ with $r=10$ and $\varepsilon = 0.005$. Note that in this setup, the core $\bs G$ in the Tucker method would have size $10^{20}$, which is impractical.  We then consider only the first two measures $a^{(1)}$ and $a^{(2)}$ and solve the $d=2$ associated problem using $\rcg$, the best competitor in the large scale experiment in Table~\ref{tab:mixtures} and \expcp\ taken only on these 2 measures. In Figure \ref{fig:mnist_z}, we represent the latent barycenter positions $z$ as $28\times 28$ images for each method. In this experiment, we take $\varepsilon=0.001$. \expcp\ ran for 81 iterations in 103\,s and achieved a marginal error of $10^{-9}$.

\begin{figure}[H]
    \centering
    \begin{subfigure}[b]{0.32\linewidth}
        \centering
        \includegraphics[width=\linewidth]{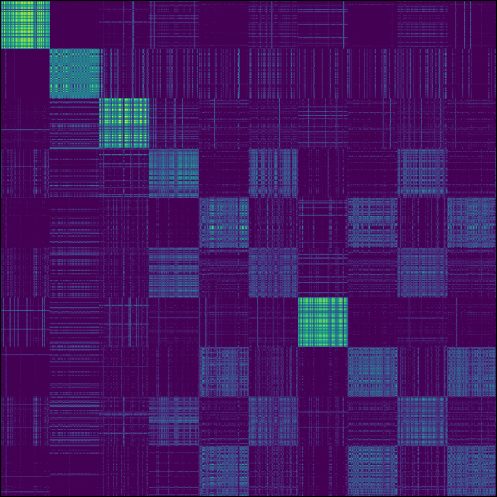}
        \caption{\expcp\ $(d)$}
        \label{fig:mnist_ecpd}
    \end{subfigure}
    \hfill
    \begin{subfigure}[b]{0.32\linewidth}
        \centering
        \includegraphics[width=\linewidth]{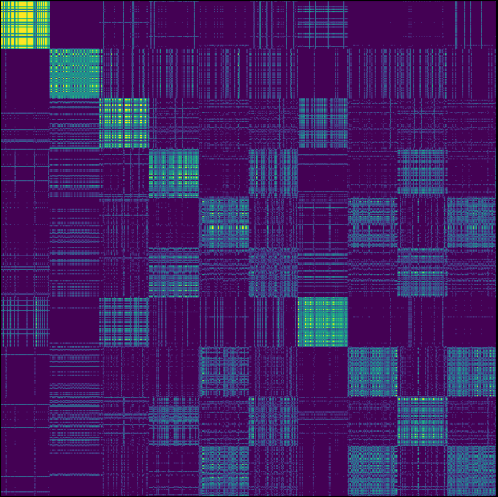}
        \caption{\expcp\ $(2)$}
        \label{fig:mnist_ecp2}
    \end{subfigure}
    \hfill
    \begin{subfigure}[b]{0.32\linewidth}
        \centering
        \includegraphics[width=\linewidth]{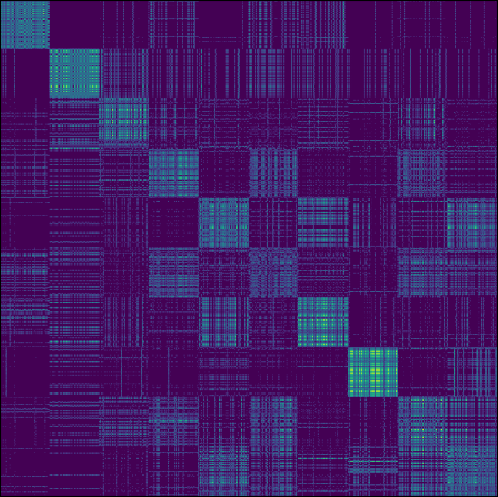}
        \caption{\rcg}
        \label{fig:mnist_rcg}
    \end{subfigure}
    
    \caption{Marginal plan $\mathrm{Marg}_{\{1,2\}}(\bs T)$ and plans from \expcp\ and \rcg\ with $a^{(1)}$ and $a^{(2)}$ in the MNIST experiment.}
    \label{fig:mnist_plans}
\end{figure}

We observe that the multimarginal problem obtains the cleanest barycenters. Some integers are merged together, but we are still able to distinguish all 10 of them. We compare the $d=2$ transport costs with the marginal plan $\mathrm{Marg}_{\{1,2\}}(\bs T)$. Transport costs are \expcp\ $(d)$: 0.3057, \expcp\ $(2)$: 0.3043, \rcg: 0.3312. In particular, taking the $(1,2)$ marginal of $\bs T$ provides a good transport plan from $a^{(1)}$ to $a^{(2)}$. Transport plans are displayed in Figure~\ref{fig:mnist_plans}. Points are ordered according to their corresponding digit, and measures are uniform.

\begin{figure}[h]
    \centering
    \includegraphics[width=0.45\linewidth]{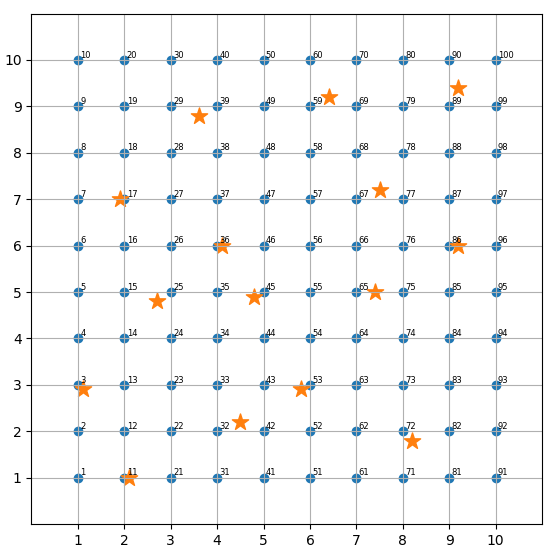}
    \caption{Sensors (orange stars) are spatially distributed over the graph network.}
    \label{fig:sensors}
\end{figure}

\paragraph{Estimating ensemble flows on a hidden Markov chain.} We use \impcp\ in the experiment from Section 6 in \cite{Haasler2021tree} (uncoupled case). We study a population of $N$ agents walking on a $n\times n$ grid according to a lazy random walk on the associated graph during $\tau$ time steps. The reference population evolution is constructed using a Schrödinger bridge, which imposes prescribed initial and final distributions ($\delta_{(1,1)}$ and $\delta_{(n,n)}$). To generate observations, we use $S$ binary sensors distributed according to Figure~\ref{fig:sensors}. Each sensor detects an agent with a probability depending on the euclidean distance between the agent and the sensor.

\begin{figure}[h]
    \centering
    \centering
    \begin{subfigure}[b]{0.245\linewidth}
        \centering
        \includegraphics[width=\linewidth,height=11cm]{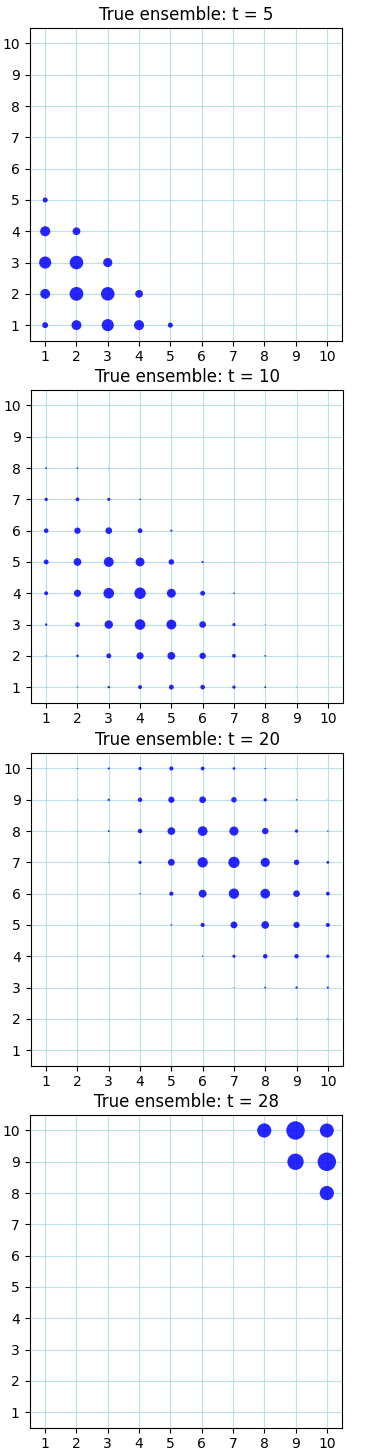}
        \caption{True flow}
        \label{fig:mnist_ecpd}
    \end{subfigure}
    \hfill
    \begin{subfigure}[b]{0.245\linewidth}
        \centering
        \includegraphics[width=\linewidth,height=11cm]{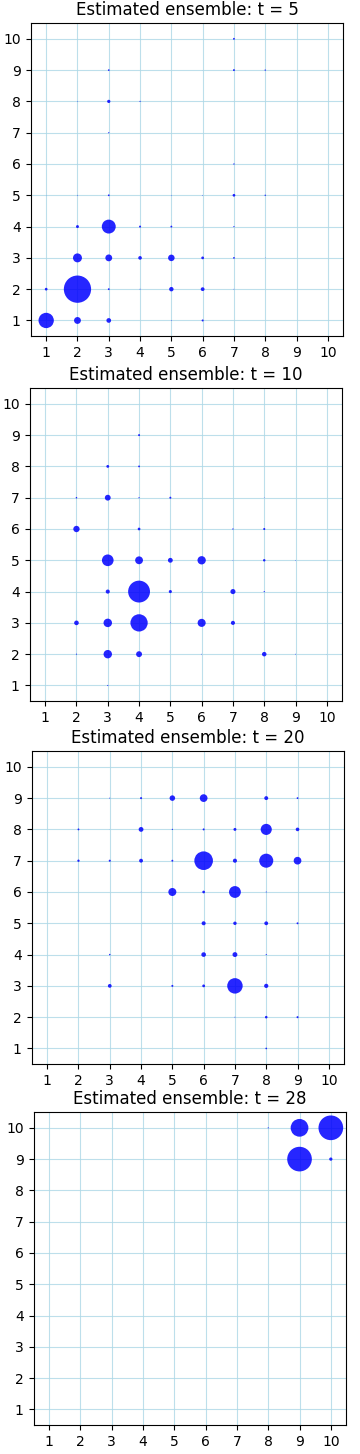}
        \caption{$r=50$}
        \label{fig:mnist_ecp2}
    \end{subfigure}
    \hfill
    \begin{subfigure}[b]{0.245\linewidth}
        \centering
        \includegraphics[width=\linewidth,height=11cm]{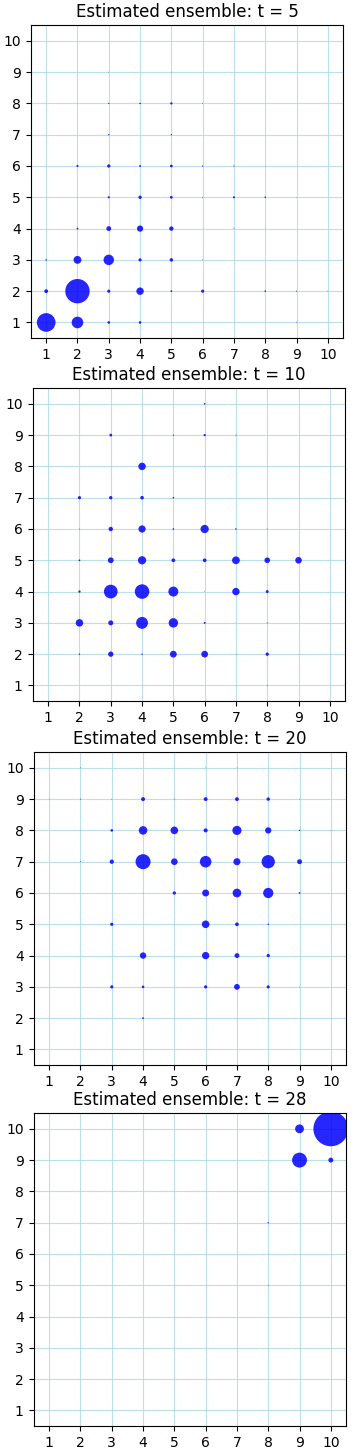}
        \caption{$r=100$}
        \label{fig:mnist_rcg}
    \end{subfigure}
    \hfill
    \begin{subfigure}[b]{0.245\linewidth}
        \centering
        \includegraphics[width=\linewidth,height=11cm]{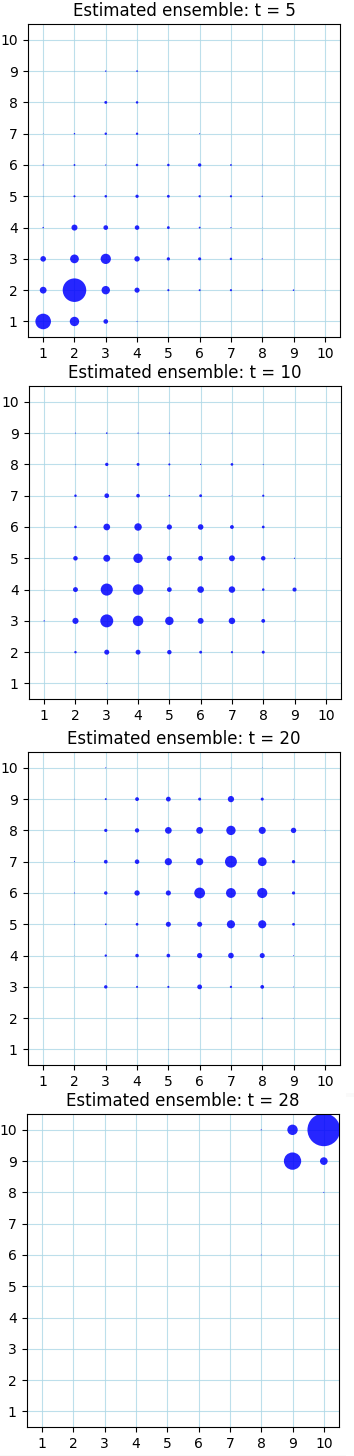}
        \caption{$r=1000$}
        \label{fig:mnist_rcg}
    \end{subfigure}
    \caption{Reconstruction of the ensemble flow using $n=10$, $N=100$, $\tau = 30$, $S=15$ for varying values of $r$ at time steps $t=5,10,20,28$.}
    \label{fig:sensors_result}
\end{figure}

The reconstruction problem is formulated as a multimarginal optimal transport problem with a tree structured cost. The marginals represent the spatial locations of the agents at each time step and the binary observations associated with each sensor. Only some of these marginals are prescribed: we impose the initial and final spatial distributions, as well as the empirical binary observation distributions provided by the sensors at every time step. In contrast, the intermediate spatial marginals are unknown and constitute the quantities to be estimated by solving the MOT problem. The temporal costs encode the random walk prior, while the observation costs are determined by the sensor likelihoods. The dimensions are: $\tau$ marginals of size $n^2$, $\tau S$ marginals of size $2$. The intermediate spatial distributions are then recovered by taking the corresponding marginals of the optimal transport tensor. In Figure~\ref{fig:sensors_result}, we report the recovered distributions of agents over the grid at time steps $t=5,10,20,28$ after solving the problem with \impcp. We observe that \impcp\ correctly tracks the center of mass of the distribution, and that the estimation gets better when the rank $r$ increases.

\paragraph{Approximation error bound.} Since the quantity $R$ in Property~\ref{ppt:approx_error} is difficult to compute, the following corollary provides a looser but more readily computable bound:

\begin{corollary}
\label{cor:approx_error}
    It is known that MOT always has a sparse solution with at most $\sum_{k=1}^d n_k - d +1 $ entries (\cite{anderes2015}, Lemma 3). Thus,  $$0 \le \mathrm{TuckerMOT}_{\mathbf r}(\mathbf{a},\bs C) - \mathrm{MOT}(\mathbf{a},\bs C) \le \|\bs C\|_\infty \ln\Big(\frac{\sum_{k=1}^d n_k - d +1}{r-1}\Big).$$
\end{corollary}

When the cost is normalized, and with $(n_1,n_2,n_3)=(15,15,15)$, $r=5$ this bound is $$\ln\big(\frac{43}{4}\big) \approx 2.37.$$ However, in this setting with Gaussian samples \expcp\ achieves a 0.1039 transport cost, when Sinkhorn gets 0.0830, so the error is of the order of $10^{-2}$, which is far smaller than the bound. Similarly, in the $d=2$ case, the bound $\ln(\min(n_1,n_2)/(r-1))$ provided by \cite{Scetbon2022stats} gives $\ln(200/9)\approx 3.10$ in the setting of Figure~\ref{fig:unit_roots}, when the transport costs achieved by low-rank solvers are $\approx 0.1$.

\subsection{Experiments in the $d=2$ case}
\label{app:exp_d2}
\paragraph{Methods.} In this section, we compare \impcp, \expcp\ and \tucker\ with other state-of-the-art methods, including \latot\ \citep{lin_making_2021} and RiemCG (\rcg) \citep{jawanpuria_riemannian_2026}. In the $d=2$ case, \impcp, \expcp\ and \tucker\ are variants of \lot, \textsc{fc} and \frlc\ from \cite{scetbon_low-rank_2021}, \cite{forrow19fc} and \cite{halmos_low-rank_2024} respectively. For \impcp\ and \tucker, we use code from these works in this section. Note that the implementation of \impcp\ from \cite{scetbon_low-rank_2021} still takes advantage of the structure of the squared Euclidean cost. We also include \tc\ \citep{schmidt2026transport} when applicable, namely, when $n=m$ and $n$ is small enough for its initialization, which involves computing an optimal full-rank Monge map, to be completed within the prescribed time.

%The \frlc\ formulation is more general than Eq.~\eqref{eq:lot_scetbon} as it allows different inner marginals: \cite{halmos_low-rank_2024} take a factorization $T=QXR\tsp$ of a transport plan $T\in\Pi_r(a,b)$ and define the inner marginals $g_Q = Q\tsp\1_n$ and $g_R = R\tsp\1_m$, not necessarily equal. They define the latent coupling factorization of $T$ as $$T = Q\diag(1/g_Q)P\diag(1/g_R)R\tsp$$ where $P\in \Pi(g_Q,g_R)$. Thus, giving a new parametrization of the space $\Pi_r(a,b)$. This equivalent formulation of the LOT problem (\ref{eq:lot}) is also solved by a Mirror Descent scheme.

The \latot\ formulation allows different inner marginals as in the Tucker formulation. \cite{lin_making_2021} optimize over the source and target hub locations $\mu^x$ and $\mu^y$ directly, using the same Sinkhorn barycenter algorithm as in \expcp\ to update couplings. 

\paragraph{Setup.} We adopt the experimental setup from \cite{jawanpuria_riemannian_2026}: each method terminates when $|\ps{C}{T_k} - \ps{C}{T_{k-1}}|/\ps{C}{T_{k-1}} < 10^{-6}$ for five consecutive iterations, after 1000 iterations or after 180 seconds. Apart from this stopping criterion, we use the default parameters for each method. An important parameter of \expcp\ is the maximum number of iteration and tolerance of the inner Sinkhorn-like step, which we set to 1300 and $10^{-7}$ in order to achieve marginal error $< 10^{-7}$. In the tables below, the marginal error denotes the maximum violation among the marginal constraints, i.e. $\max (\|T\1_m-a\|_\infty,\|T\tsp\1_n-b\|_\infty).$ If not specified, the regularization parameter $\varepsilon$ is set to 0.005 and the cost is the squared euclidean cost. We use the normalized cost $C/C_{\max}$ in every experiment, except in the \latot\ method where numerical errors occur otherwise. 

In Table \ref{tab:mixtures}, we reproduce the Gaussian-mixture benchmark from \cite{scetbon_low-rank_2021} (Appendix G). We sample $n=m=10000$ points in $\bbR^2$ from two Gaussian mixtures and compare the transport costs obtained by the different low-rank solvers. We report the ratio $f/f^\star$, where $f^\star$ is the exact transport cost and $f = \ps{C}{T}$ is the transport cost achieved by the concerned method. We also report the \textit{success rate}, where a run fails if the obtained marginal error is above $10^{-4}$.  The experiment is realized 20 times over different sampling seeds, and values are averaged.

\begin{table}[h]
\caption{Comparison of the low-rank transport cost ratio, marginal errors, and computation times of different solvers ($n=m=10000$, $d=2$, Gaussian mixtures). Values are averaged over 20 seeds. The \tc\ method is excluded as its initialization takes more than 180\,s.}
\label{tab:mixtures}
\begin{center}
\begin{tabular}{l c c c c c}
\hline
Method & Ratio $f/f^\star$ & Marg. err & Iterations & Time (s) & Success rate \\
\hline
Sinkhorn & $1.087 \pm 0.043$ & $10^{-9}$ & $135 \pm 6$ & $43.4 \pm 1.8$ & $1.00$ \\
\hline
\gr{Rank} $r=10$\\
\expcp & $1.181 \pm 0.052$ & $10^{-10}$ & $86 \pm 32$ & $\gr{3.7} \pm \gr{1.2}$ & $1.00$\\
\impcp & $1.180 \pm 0.052$ & $10^{-11}$ & $979 \pm 49$ & $153.3 \pm 10.3$ & $1.00$ \\
\tucker & $1.240 \pm 0.058$ & $10^{-8}$ & $101 \pm 2$ & $181.1 \pm 0.6$ & $1.00$ \\
\rcg & $\gr{1.175} \pm \gr{0.051}$ & $10^{-7}$ & $31 \pm 7$ & $188.2 \pm 4.0$ & $1.00$ \\
\latot & $1.839 \pm 0.015$ & $10^{-9}$ & $46 \pm 21$ & $26.2 \pm 11.5$ & $1.00$ \\
\hline
\gr{Rank} $r=50$\\
\expcp & $1.055 \pm 0.045$ & $10^{-10}$ & $171 \pm 76$ & $\gr{16.4} \pm \gr{6.7}$ & $1.00$ \\
\impcp & $1.068 \pm 0.043$ & $10^{-11}$ & $445 \pm 71$ & $180.3 \pm 0.2$ & $1.00$ \\
\tucker & $1.014 \pm 0.045$ & $10^{-8}$ & $70 \pm 2$ & $182.1 \pm 0.8$ & $1.00$ \\
\rcg & $\gr{1.002} \pm \gr{0.045}$ & $10^{-8}$ & $26 \pm 3$ & $199.7 \pm 10.5$ & $1.00$ \\
\latot & $1.832 \pm 0.014$ & $10^{-9}$ & $44 \pm 10$ & $37.3 \pm 8.0$ & $1.00$ \\
\hline
\end{tabular}
\end{center}
\end{table}

%\red{ce qui prend du temps pour sk est le check de la condition pour converger qui oblige à calculer C*T, ce qui est plus rapide en low rank}

It stands out that \expcp\ converges substantially faster than all other method while achieving good $f/f^*$ ratios. We observe that \rcg\ requires fewer but more costly iterations and achieves the best $f/f^*$ values. Such a low $f/f^*$ ratio for \rcg\ can be explained by the fact that low-rank methods can achieve a lower cost than the exact OT algorithm while maintaining a valid transport plan with marginal violation below $10^{-7}$. \expcp\ has the fastest iterations and requires fewer iterations than \impcp\ and \tucker\ to converge. In Figure~\ref{fig:check_time}, we study the evolution of the $f/f^*$ ratio as a function of the elapsed optimization time.

% seed = 1
\begin{figure}[h]
    \centering
\begin{tikzpicture}
\begin{axis}[
    width=6.5cm,
    height=4.5cm,
    xlabel={Time (s)},
    ylabel={$f_k/f^*$},
    ymode=log,
    xmin=0,
    xmax=10,
    ymin=0.8,
    ymax=5,
    legend style={
        at={(1.00,0.98)},
        anchor=north west,
        draw=none,
        fill=none
    },
    tick label style={font=\small},
    label style={font=\small},
]

\addplot[
    blue,
    mark=*,
    mark size=1.5pt,
] table[x=time,y=score] {check_time/check_time_blot.txt};
\addlegendentry{\expcp}

\addplot[
    green!60!black,
    mark=triangle*,
    mark size=1.5pt,
] table[x=time,y=score] {check_time/check_time_lot.txt};
\addlegendentry{\impcp}

\addplot[
    red,
    mark=square*,
    mark size=1.5pt,
] table[x=time,y=score] {check_time/check_time_rcg.txt};
\addlegendentry{\rcg}

\addplot[
    orange,
    mark=diamond*,
    mark size=1.5pt,
] table[x=time,y=score] {check_time/check_time_frlc.txt};
\addlegendentry{\tucker}

%\addplot[
%    gray,
%    mark=+,
%    mark size=1.5pt,
%] table[x=time,y=score] {check_time/%check_time_sk.txt};
%\addlegendentry{\textsc{sk}}

\addplot[
    gray,
    mark=+,
    mark size=1.5pt,
] table[x=time,y=score] {check_time/check_time_latot.txt};
\addlegendentry{\latot}

\addplot[dotted] table[row sep = crcr]{0 1 \\ 10 1 \\};
\end{axis}

\begin{scope}[xshift=7cm]
\begin{axis}[
    width=6.5cm,
    height=4.5cm,
    xlabel={Time (s)},
    ymode=log,
    ytick=\empty,
    xmin=0,
    xmax=10,
    ymin=0.8,
    ymax=5,
    tick label style={font=\small},
    label style={font=\small},
]

\addplot[
    blue,
    mark=*,
    mark size=1.5pt,
] table[x=time,y=score] {check_time2/check_time_blot.txt};

\addplot[
    green!60!black,
    mark=triangle*,
    mark size=1.5pt,
] table[x=time,y=score] {check_time2/check_time_lot.txt};

\addplot[
    red,
    mark=square*,
    mark size=1.5pt,
] table[x=time,y=score] {check_time2/check_time_rcg.txt};

\addplot[
    orange,
    mark=diamond*,
    mark size=1.5pt,
] table[x=time,y=score] {check_time2/check_time_frlc.txt};

%\addplot[
%    gray,
%    mark=+,
%    mark size=1.5pt,
%] table[x=time,y=score] {check_time2/%check_time_sk.txt};

\addplot[
    gray,
    mark=+,
    mark size=1.5pt,
] table[x=time,y=score] {check_time2/check_time_latot.txt};

\addplot[dotted] table[row sep = crcr]{0 1 \\ 10 1 \\};
\end{axis}
\end{scope}
\end{tikzpicture}
    \caption{Evolution of the $f/f^*$ ratio during the optimization. Setting is the same as in Table~\ref{tab:mixtures}. Left $r=10$, right $r=50$.}
    \label{fig:check_time}
\end{figure}

Again, we see that \expcp\ achieves a good time-accuracy tradeoff. In this setting, we see that the optimum is in fact reached in the first iterations. We also observe that \latot\ fails to optimize further than its initialization in this setting with default ultra-parameter values, one could try to select a suitable value of $\varepsilon$ using a bisection method (as advised by \cite{lin_making_2021}) at the cost of increased runtime. 

In Table \ref{tab:rank_evol}, we reproduce the study of \cite{jawanpuria_riemannian_2026}. We sample $n=50$, $m=60$ points from the standard Gaussian distribution in 2D and report the evolution of the low-rank transport cost and the marginal error when the rank $r$ varies.  The experiment is realized 50 times over different sampling seeds, and values are averaged.

\begin{table}[H]
\caption{Transport cost, marginal error, and time for each method as a function of the rank $r$. ($n=50$, $m=60$, $d=2$, standard Gaussians). Values are averaged over 50 runs.}
\label{tab:rank_evol}
\begin{center}
\begin{tabular}{ccccc|cccc}
\hline
$r$ & \multicolumn{4}{c|}{Transport cost ($\cdot 10^2$)}& \multicolumn{4}{c}{Time (s)} \\
 & \impcp & \expcp & \tucker & \rcg &  \impcp & \expcp & \tucker & \rcg \\
\hline
2 & \gr{8.60 $\pm$ 1.19} & 8.70 $\pm$ 1.15 & 9.38 $\pm$ 1.39 & 8.68 $\pm$ 1.21 & 2.6 & \gr{0.1} & 62.9  & 0.8 \\
3 & 5.91 $\pm$ 0.81 & \gr{5.84 $\pm$ 0.82} & 6.82 $\pm$ 1.34 & 5.97 $\pm$ 0.80 & 3.3 & \gr{0.2}  & 93.9  & 1.7 \\
5 & \gr{4.04 $\pm$ 0.55} & 4.08 $\pm$ 0.59 & 5.19 $\pm$ 0.96 & 4.17 $\pm$ 0.57 & 9.4  & \gr{0.3} & 103.8  & 1.4  \\
8 & \gr{2.85 $\pm$ 0.44} & 2.86 $\pm$ 0.46 & 3.70 $\pm$ 0.58 & 2.91 $\pm$ 0.48 & 16.3  & \gr{0.4}  & 110.7  & 1.7  \\
10 & \gr{2.34 $\pm$ 0.40} & 2.35 $\pm$ 0.37 & 3.05 $\pm$ 0.51 & 2.40 $\pm$ 0.45 & 17.5  & \gr{0.4}  & 120.6 & 2.8 \\
15 & 1.95 $\pm$ 0.37 & \gr{1.92 $\pm$ 0.39} & 2.46 $\pm$ 0.39 & 2.02 $\pm$ 0.39 & 14.9  & \gr{0.4} & 120.0 & 1.8\\
20 & 1.72 $\pm$ 0.36 & \gr{1.70 $\pm$ 0.37} & 2.10 $\pm$ 0.35 & 1.77 $\pm$ 0.39 & 13.9  & \gr{0.4} & 123.3  & 2.1\\
30 & \gr{1.55 $\pm$ 0.35} & \gr{1.55 $\pm$ 0.35} & 1.77 $\pm$ 0.36 & \gr{1.55 $\pm$ 0.35} & 13.6 & \gr{0.5}  & 134.5 & 1.8  \\
\hline
\end{tabular}
\end{center}
\end{table}

As expected, we observe that the transport cost decreases as the rank increases. In this small scale setting, we see that \impcp\ and \expcp\ dominate in terms of transport cost optimality, while \expcp\ remains fastest to converge.

In Table~\ref{tab:rankr}, we study the behavior of the algorithms in a setting where the optimal transport plan has rank $r$. To construct such a setting, we create two sets of $r$ hubs, $(\mu^x_i)_i$ and $(\mu_j^Y)_j$, and place $n$  points exactly at the locations $\mu^x_i$ and $m=n$ points at the locations $\mu^y_j$. Measures are taken uniform. We report the average $f/f^*$ value of each method and the rate at which a method is close ($|f/f^*-1| < 10^{-3}$) to the exact optimal transport cost over 50 sampling seeds.

\begin{table}[H]
    \caption{Transport cost ratio and optimality rate for each method when the optimal transport plan has rank $r$. $(n,m,r)=(20,20,5)$. 50 iterations.}
    \label{tab:rankr}
    \begin{center}
    \begin{tabular}{c|ccccccc}
       Method & Sinkhorn & \expcp & \impcp & \tucker & \rcg & \latot & \tc \\ \hline
        Mean $f/f^*$ & 1.003 & \gr{1.001} & 1.032 & 1.245 & 1.035 & 1.802 & 2.440 \\
        Optimality rate & 0.56 & \gr{0.80} & 0.42 & 0.20 & 0.60 & 0.14& 0.00 
    \end{tabular}
    \end{center}
\end{table}

We observed that low-rank methods tended to merge some latent hubs $z_s$ together, resulting in a sub-optimal transport plan. Except for \tucker, \latot\ and \tc, each method finds the optimal transport plan at good rate. In particular, \expcp\ achieves the highest success rate.

In Figure~\ref{fig:unit_roots}, we study the position of the hubs $z_s$ and latent source and target hubs $\mu^x_s$ and $\mu_s^y$ for the 4 best performing methods in Table~\ref{tab:rankr} in the following setting from \cite{halmos_low-rank_2024}: source points are sampled from Gaussians centered at the $5^{\text{th}}$ roots of unity, while target points are sampled from Gaussians centered at the $10^{\text{th}}$ roots of unity. Low-rank transport plan are computed with $r=10$. Note that each "blob" does not have the same mass (number of points). 

% seed = 3, uniform measure
\begin{figure}[H]
    \centering
    \begin{tikzpicture}[scale=0.8]
        \begin{groupplot}[
            group style={
                group size=4 by 1,
                horizontal sep=1cm
            },
            width=4.9cm,
            height=4.9cm,
            axis equal image,
            xlabel={},
            ylabel={},
            legend style={
                font=\scriptsize,
                at={(-0.02,0.98)},
                anchor=north east
            },]

        \nextgroupplot[title={\expcp}]

        \addplot[
            orange,
            opacity=0.2,
            only marks,
            mark=*
        ] table[
            x index=0,
            y index=1
        ] {unit_roots/x.txt};
        \addlegendentry{$x$}

        \addplot[
            green!70!black,
            opacity=0.1,
            only marks,
            mark=*
        ] table[
            x index=0,
            y index=1
        ] {unit_roots/y.txt};
        \addlegendentry{$y$}

        \addplot[
            green!50!black,
            only marks,
            mark=diamond*,
            mark size=3pt
        ] table[
            x index=0,
            y index=1
        ] {unit_roots/mu_x_expl.txt};
        \addlegendentry{$\mu^x$}

        \addplot[
            red!50!black,
            only marks,
            mark=square*,
            mark size=3pt
        ] table[
            x index=0,
            y index=1
        ] {unit_roots/mu_y_expl.txt};
        \addlegendentry{$\mu^y$}

        \addplot[
            black,
            opacity=0.9,
            only marks,
            mark=star
        ] table[
            x index=0,
            y index=1
        ] {unit_roots/z_expl.txt};
        \addlegendentry{$z$}

        \nextgroupplot[title={\impcp}]

        \addplot[
            orange,
            opacity=0.2,
            only marks,
            mark=*
        ] table[
            x index=0,
            y index=1
        ] {unit_roots/x.txt};

        \addplot[
            green!70!black,
            opacity=0.1,
            only marks,
            mark=*
        ] table[
            x index=0,
            y index=1
        ] {unit_roots/y.txt};

        \addplot[
            green!50!black,
            only marks,
            mark=diamond*,
            mark size=3pt
        ] table[
            x index=0,
            y index=1
        ] {unit_roots/mu_x_scetbon.txt};

        \addplot[
            red!50!black,
            only marks,
            mark=square*,
            mark size=3pt
        ] table[
            x index=0,
            y index=1
        ] {unit_roots/mu_y_scetbon.txt};

        \addplot[
            black,
            opacity=0.9,
            only marks,
            mark=star
        ] table[
            x index=0,
            y index=1
        ] {unit_roots/z_scetbon.txt};

        \nextgroupplot[title={\tucker}]

        \addplot[
            orange,
            opacity=0.2,
            only marks,
            mark=*
        ] table[
            x index=0,
            y index=1
        ] {unit_roots/x.txt};

        \addplot[
            green!70!black,
            opacity=0.1,
            only marks,
            mark=*
        ] table[
            x index=0,
            y index=1
        ] {unit_roots/y.txt};

        \addplot[
            green!50!black,
            only marks,
            mark=diamond*,
            mark size=3pt
        ] table[
            x index=0,
            y index=1
        ] {unit_roots/mu_x_frlc.txt};

        \addplot[
            red!50!black,
            only marks,
            mark=square*,
            mark size=3pt
        ] table[
            x index=0,
            y index=1
        ] {unit_roots/mu_y_frlc.txt};

        \nextgroupplot[title={\rcg}]

        \addplot[
            orange,
            opacity=0.2,
            only marks,
            mark=*
        ] table[
            x index=0,
            y index=1
        ] {unit_roots/x.txt};

        \addplot[
            green!70!black,
            opacity=0.1,
            only marks,
            mark=*
        ] table[
            x index=0,
            y index=1
        ] {unit_roots/y.txt};

        \addplot[
            green!50!black,
            only marks,
            mark=diamond*,
            mark size=3pt
        ] table[
            x index=0,
            y index=1
        ] {unit_roots/mu_x_riem.txt};

        \addplot[
            red!50!black,
            only marks,
            mark=square*,
            mark size=3pt
        ] table[
            x index=0,
            y index=1
        ] {unit_roots/mu_y_riem.txt};

        \addplot[
            black,
            opacity=0.9,
            only marks,
            mark=star
        ] table[
            x index=0,
            y index=1
        ] {unit_roots/z_riem.txt};
        \end{groupplot}
    \end{tikzpicture}
    \caption{Position of the hubs $z_s$, the source and target hubs $\mu^x_s$ and $\mu^y_s$ when the $x_i$ are sample from Gaussians centered at the $5^{\text{th}}$ roots of unity ($\times 5$), the $y_j$ are sampled from Gaussians centered at the $10^{\text{th}}$ roots of unity ($\times 10$), and $(n,m,r) = (200,200,10).$ Running times are \expcp: 0.29\,s, \impcp: 25.1\,s, \tucker: 73.3\,s, \rcg: 3.60\,s. Transport cost are \expcp: 0.1059, \impcp: 0.1030, \tucker: 0.1079, \rcg: 0.1039.}
    \label{fig:unit_roots}
\end{figure}